\documentclass{article}

\usepackage{arxiv}

\usepackage[utf8]{inputenc} 
\usepackage[T1]{fontenc}    
\usepackage{hyperref}       
\usepackage{url}            
\usepackage{booktabs}       
\usepackage{amsfonts}       
\usepackage{nicefrac}       
\usepackage{microtype}      
\usepackage{lipsum}		
\usepackage{graphicx}
\usepackage{natbib}
\usepackage{doi}
\usepackage{algorithm}
\usepackage{algorithmic}
\usepackage{amsmath} 
\usepackage{amssymb}
\usepackage{comment}
\usepackage{mathbbol}

\usepackage{newfloat}
\usepackage{listings}
\usepackage{subfigure}

\newtheorem{assumption}{Assumption}
\newtheorem{theorem}{Theorem}[section]
\newtheorem{lemma}[theorem]{Lemma}

\title{Linear Contextual Bandits with Interference}

\date{} 					

\author{ 
Yang Xu
\\
	Department of Statistics\\
	North Carolina State University\\
	\texttt{yxu63@ncsu.edu} \\
	\And
        Wenbin Lu \\
	Department of Statistics\\
	North Carolina State University\\
	\texttt{wlu4@ncsu.edu} \\
 	\And
        Rui Song \\
	Department of Statistics\\
	North Carolina State University\\
	\texttt{songray@gmail.com} \\
}

\renewcommand{\headeright}{}
\renewcommand{\shorttitle}{Linear Contextual Bandits With Interference (LinCBWI)}

\hypersetup{
pdftitle={Linear Contextual Bandits with Interference},
pdfsubject={q-bio.NC, q-bio.QM},
pdfauthor={David S.~Hippocampus, Elias D.~Striatum},
pdfkeywords={First keyword, Second keyword, More},
}

\begin{document}
\maketitle

\begin{abstract}
Interference, a key concept in causal inference, extends the reward modeling process by accounting for the impact of one unit's actions on the rewards of others. In contextual bandit (CB) settings, where multiple units are present in the same round, potential interference can significantly affect the estimation of expected rewards for different arms, thereby influencing the decision-making process. Although some prior work has explored multi-agent and adversarial bandits in interference-aware settings, the effect of interference in CB, as well as the underlying theory, remains significantly underexplored. In this paper, we introduce a systematic framework to address interference in Linear CB (LinCB), bridging the gap between causal inference and online decision-making. We propose a series of algorithms that explicitly quantify the interference effect in the reward modeling process and provide comprehensive theoretical guarantees, including sublinear regret bounds, finite sample upper bounds, and asymptotic properties. The effectiveness of our approach is demonstrated through simulations and a synthetic data generated based on MovieLens data.
\end{abstract}


\section{Introduction}
Interference has increasingly gained attention from researchers in recent years. 
In multi-agent bandit settings, this phenomenon often arises when the outcome for one agent or unit depends on the actions taken by others, which is typically influenced through an underlying neighborhood structure. This structure can introduce dependencies among the agents' actions, which classical bandit model that typically assumes independent operation or simple information sharing often fail to account for. As a result, these traditional models may not adequately address the complexities introduced by interference, leading to significantly biased reward modeling and consequently suboptimal decision-making.

In many real-world scenarios, interference often manifests as a pervasive and challenging factor to address. For instance, during the COVID-19 pandemic, local governments have sought optimal personalized quarantine policies over extended periods. In this context, each local community government functions as an agent/unit. The observed reward, such as the number of infected individuals, reflects the health status of the community after implementing a specific policy. Over time, governments adjust their policies based on the latest health status to mitigate virus spread. This situation naturally aligns with a bandit framework. However, as noted, one community's health outcomes can be influenced by the quarantine policies of neighboring communities due to population movement and the nature of virus spread, illustrating the presence of interference.

Another example is in large advertising campaigns, where advertisers manage multiple ad lines across various campaigns. These campaigns often target potentially overlapping audiences through different channels with unique creatives. The overlap in target consumers and the similarity between ad lines can cause the Return Of Investment (ROI) of one ad line to be affected by the impression status of another. For example, if one ad targets students with Nike shoes and another with Adidas shoes, the overall ROI from showing both ads simultaneously might be lower or higher than showing just one, due to the shared target audience. This illustrates the presence of interference.

The issue of interference has been extensively studied in the causal inference literature, particularly when the classical Stable Unit Treatment Value Assumption (SUTVA) is violated \citep{rosenbaum2007interference}. However, this problem becomes more challenging in the context of online sequential decision-making problems such as contextual bandits for two main reasons listed below.

\textbf{First}, because multiple units' actions affect each other's reward modeling, the action space is often high-dimensional and thus challenging to quantify without a precise understanding of the interference structure. In fact, estimating heterogeneous treatment effects under interference in single-stage settings is already challenging problem to address \citep{viviano2020experimental, leung2022rate}, and extending this to sequential decision-making scenarios like bandit -- where balancing exploration and exploitation complicates the pursuit of larger cumulative rewards -- adds even more difficulty. \textbf{Second}, CB introduce an additional layer of complexity. The dynamic nature of units over time makes it harder to quantify the heterogeneous effects of interference, as contextual information and the evolving patterns of interference complicate the process of online decision making. 

Although several previous work has incorporated interference issue in multi-agent multi-armed bandits \citep{verstraeten2020multi, bargiacchi2018learning, dubey2020kernel,agarwal2024multi} and adversarial bandits \citep{jia2024multi}, the problem of interference in contextual bandits remains largely unexplored. Our contributions are as follows.
\begin{itemize}
    \item We are the very first work to address the interference issue in contextual bandits with multiple units involved in each round, bridging the gap between SUTVA violations in causal inference and online decision making. 
    \item We propose a systematic framework that extends the classical LinCB to interference-aware scenarios, offering comprehensive theoretical guarantees, including finite-sample upper bounds and sublinear regret.
    \item We are also the first work to establish the asymptotic properties of regression coefficients and the optimal value function in an online setting with interference, introducing a probability of exploration and a small clipping rate to ensure estimation consistency. The performance of our estimator is validated through simulation studies and synthetic data based on MovieLens.
\end{itemize} 

\section{Related Work}
\subsection{Interference in Single Stage} 
In single-stage setting, existing literature varies significantly in defining interference, often assuming different structures for simplifying reward modeling. For example, \citet{su2019modelling} considers the reward as a linear function of neighbors’ covariates and treatments, which tends to make stronger but more interpretable assumptions during the interference modeling stage, allowing for flexibility and comprehensive study in both theory and algorithms. 

On the contrary, another body of work focuses on using partial interference and exposure mapping to quantify interference \citep{sobel2006randomized, qu2021efficient, hudgens2008toward, forastiere2021identification, aronow2017estimating, bargagli2020heterogeneous}. While this approach typically requires fewer assumptions during the reward modeling stage, it often relies on additional requirements, such as knowing the form of the exposure mapping function \citep{manski2013identification, aronow2017estimating, bargagli2020heterogeneous} or assuming i.i.d. clusters \citep{qu2021efficient}. These assumptions can be overly restrictive in later stages and may not clearly explain direct and spillover effects. 
Therefore, in our work, we consider linear CB with interference for interpretability, and study both theories and algorithms under this framework.

\subsection{Cooperative Multi-Agent Bandits} 
Multi-agent bandits typically assume a fixed set of $N$ agents making decisions over time. Some existing works, such as \citet{martinez2019decentralized} and \citet{landgren2016distributed}, focus on information sharing between agents in a distributed system. While sharing historical data can enhance the reward learning process among agents, these studies still assume that each agent's reward depends solely on its own actions, excluding the possibility of interference.

Another line of research, including \citet{verstraeten2020multi}, \citet{bargiacchi2018learning}, \citet{dubey2020kernel}, \citet{agarwal2024multi}, and \citet{jia2024multi}, considers more general reward models where the actions of other agents can affect an individual agent's reward. Specifically, \citet{bargiacchi2018learning} and \citet{verstraeten2020multi} extended the Upper Confidence Bound (UCB) algorithm and Thompson Sampling (TS) from the classical MAB setting to multi-agent scenarios. \citet{dubey2020kernel} introduced a kernelized UCB algorithm, where interference is mediated through network contexts. Although \citet{jia2024multi} also considered interference in bandits, their focus was on adversarial bandits with homogeneous actions, which is not as flexible as our approach where heterogeneous actions are allowed for units within the same round.

However, all of the aforementioned literature only considers a multi-armed bandit (MAB) setting with a fixed and finite number of agents interacting at each round, which is different from our setting where agents or units can vary over time with evolving contextual information.

\section{Problem Formulation}
In each round $t\in\{1,\dots, T\}$, we assume there are $N_t$ units with contextual information $\boldsymbol{X}_{ti}\in\mathbb{R}^{d}$ in a network making sequential decisions simultaneously. At each time step $t$, unit $i \in \{1,\dots,N_t\}$ chooses an action $A_{ti} \in \mathcal{A}$ and collects a reward $R_{ti}$. Define $\bar{N}_t = \sum_{s=1}^t N_t$ as the total number of units up to round $t$. Due to interference, the potential outcome of unit $i$ at round $t$ is defined as $R_{ti}(\boldsymbol{a}_t)$, where $\boldsymbol{a}_t = (a_{t1},\dots,a_{tN_t})^T$ is the action assignment vector for all units at round $t$.

To quantify the interference level between any two units in the same round, we suppose there exists a (normalized) adjacency matrix $\boldsymbol{W}_t$ at round $t$, such that $W_{t,ij}$ denotes the causal interference strength from unit $i$ to unit $j$. Note that in our setup, $\boldsymbol{W}_t$ is not required to be symmetric, i.e. the causal interference strength from unit $i$ to unit $j$ can differ from that of $j$ to $i$. By default, we assume $\boldsymbol{W}_{t,ij} \in [-1,1]$, and $\boldsymbol{W}_{t,ii} = 1$ for any $t$ and $1 \leq i, j \leq N_t$.

Defining an interference matrix $\boldsymbol{W}_t$ is both intuitive and flexible enough to model various real-world scenarios. For instance, in the special case where $\boldsymbol{W}_t$ is symmetric and takes values from ${0,1}$, it can represent a neighborhood structure or friend network, where $\boldsymbol{W}_{t,ij} = 1$ indicates that units $i$ and $j$ are connected, and $\boldsymbol{W}_{t,ij} = 0$ means they are not. Since this information is derived from societal interactions, we will assume $W$ is known throughout this paper. We assume the reward of unit $i$ at round $t$ to be generated by
\begin{equation}\label{eq:0}
\begin{aligned}
R_{ti} = \sum_{j=1}^{N_t} W_{t,ij} \cdot f(\boldsymbol{X}_{tj}, A_{tj}) + \epsilon_{ti},
\end{aligned}
\end{equation}
where the reward $R_{ti}$ is a linear combination of some payoff function $f$, $\boldsymbol{W}_t = \{W_{t,ij}\}_{1\leq i,j\leq N_t}$ is an $N_t \times N_t$ matrix quantifying the degree of interference between units. $\epsilon_{ti} \sim \mathcal{N}(0, \sigma^2)$ is a noise term satisfying $\epsilon_{ti}\perp (\boldsymbol{X}_t,\boldsymbol{W}_t) | \boldsymbol{A}_t$\footnote{Here we assume the noise term $\epsilon_{ti}$ is conditionally independent of the contextual information and the interference matrix $\boldsymbol{W}_t$, given the action taken at round $t$. This is a more relaxed assumption compared to i.i.d. random noise.}. Throughout this paper, we assume that $|\mathbb{E}[R_{ti}]| \leq U$, i.e., the expected reward of each unit $i$ at round $t$ can be bounded by a large constant $U$.

Assuming the reward generation process follows Equation \eqref{eq:0} is both intuitive and possesses a very beneficial property. With some simple algebra, we can show that
\begin{equation}\label{eq:2}
\begin{aligned}
&\sum_{i=1}^{N_t}\mathbb{E}[R_{ti}] = \sum_{i=1}^{N_t}\sum_{j=1}^{N_t} W_{t,ij} \cdot f(\boldsymbol{X}_{tj}, A_{tj}) = \sum_{j=1}^{N_t}\sum_{i=1}^{N_t} W_{t,ji}  f(\boldsymbol{X}_{ti}, A_{ti}) = \sum_{i=1}^{N_t} \omega_{ti} f(\boldsymbol{X}_{ti}, A_{ti}),
\end{aligned}
\end{equation}
where we define $\omega_{ti} := \sum_{j=1}^{N_t} W_{t,ji}$ as the \textit{interference weight} of unit $i$ at round $t$. 
The last term further indicates that the optimal action depends solely on the covariates of each individual unit, with interference influencing the direction of optimality through the sign of the weight $\omega_{ti}$. Since the optimal action that maximizes $\sum_{i=1}^{N_t}\mathbb{E}[R_{ti}]$ is determined by $\boldsymbol{X}_{ti}$ and $\omega_{ti}$, we don't need to account for the covariate information and actions of all units to achieve the globally optimal action. This simplifies the decision-making process and makes it more practical for real-world applications.

\subsection{Offline Optimization}
Let's consider a linear payoff function for $f$ with $K=2$ arms to establish the entire framework and theory behind. Suppose there exists a coefficient vector $\boldsymbol{\beta}_a\in \mathbb{R}^{d}$ for each action $a\in \mathcal{A}=\{0,1\}$ to quantify the effect of each covariate in $\boldsymbol{X}_{ti}$. Similar to classical linear CB, we assume
\begin{equation*}
    f(\boldsymbol{X}_{tj},a_{tj}) = \boldsymbol{X}_{tj}'\boldsymbol{\beta}_0\cdot \boldsymbol{1}\{a_{tj}=0\}  + \boldsymbol{X}_{tj}'\boldsymbol{\beta}_1\cdot\boldsymbol{1}\{a_{tj}=1\},
\end{equation*}
which gives us
\begin{equation}\label{eq:linear_r}
\begin{aligned}
    R_{ti} & = \sum_{a\in\mathcal{A}}\sum_{j=1}^{N_t} W_{t,ij}\boldsymbol{X}_{tj}'\boldsymbol{\beta}_a \boldsymbol{1}\{A_{tj}=a\}+\epsilon_{ti}.
\end{aligned}
\end{equation}
Therefore, for each round-unit pair $(t,i)$, the optimal action that maximizes the cumulative reward is given by:
\begin{equation}\label{eq:3}
    A_{ti} = \boldsymbol{1}\left\{\omega_{ti}\boldsymbol{X}_{ti}'(\boldsymbol{\beta}_1-\boldsymbol{\beta}_0) > 0\right\},
\end{equation}
where the sign of $\omega_{ti}$ determines if the interference weight causes a flip in action that maximizes the cumulative reward.

For the simplicity of notation, we denote $\boldsymbol{\beta} = (\boldsymbol{\beta}_0',\boldsymbol{\beta}_1')'\in \mathbb{R}^{2d}$. Define $\boldsymbol{A}_t = (A_{t1},\dots,A_{tN_t})'\in\mathbb{R}^{N_t}$ as the action assignment vector for all units at round $t$, $\boldsymbol{R}_t = (R_{t1},\dots,R_{tN_t})'$ as the vector of rewards collected at round $t$, $\boldsymbol{X}_t = (\boldsymbol{X}_{t1},\dots,\boldsymbol{X}_{tN_t})'\in\mathbb{R}^{N_t\times d}$ as the covariate information matrix for all units at round $t$, and $\boldsymbol{W}_{ti} = \text{diag}(\{W_{t,ij}\}_{1\leq j\leq N_t})$ as an $N_t$ by $N_t$ diagonal matrix.  We then define a $2d$-dimensional transformed covariate vector for each round-unit pair $(t,i)$ as:  
\begin{equation}\label{eq:X_tilde}
    \widetilde{\boldsymbol{X}}_{ti} = (\boldsymbol{1}_{N_t}'\boldsymbol{W}_{ti}\text{diag}(\boldsymbol{1}_{N_t}-\boldsymbol{A}_t)\boldsymbol{X}_t, \boldsymbol{1}_{N_t}'\boldsymbol{W}_{ti}\text{diag}(\boldsymbol{A}_t)\boldsymbol{X}_t)'.
\end{equation}

With some straightforward algebra, the expected reward can be expressed linearly as $\mathbb{E}[R_{ti}] = \widetilde{\boldsymbol{X}}_{ti}'\boldsymbol{\beta}$. Furthermore, we denote $\widetilde{\boldsymbol{X}}_{t} = \big(\widetilde{\boldsymbol{X}}_{t1},\dots,\widetilde{\boldsymbol{X}}_{tN_t}\big)'\in\mathbb{R}^{N_t\times 2d}$ similarly as the transformed covariate information matrix at round $t$, and $\widetilde{\boldsymbol{X}}_{1:t} = \big(\widetilde{\boldsymbol{X}}_{1}',\dots,\widetilde{\boldsymbol{X}}_{t}'\big)'\in \mathbb{R}^{\bar{N}_t\times 2d}$ as the transformed covariate information matrix collected up to round $t$. Similarly, we define $\boldsymbol{R}_{1:t} = (\boldsymbol{R}_{1}',\dots,\boldsymbol{R}_{t}')'\in\mathbb{R}^{\bar{N}_t}$. As such, the ordinary least square (OLS) estimator can be obtained by
\begin{equation}\label{eq:beta_est}
    \widehat{\boldsymbol{\beta}}_t^* = \big(\widetilde{\boldsymbol{X}}_{1:t}'\widetilde{\boldsymbol{X}}_{1:t}\big)^{-1}\widetilde{\boldsymbol{X}}_{1:t}'\boldsymbol{R}_{1:t}\in \mathbb{R}^{2d}.
\end{equation} 
In an offline optimization setting, one can replace the true value of $\boldsymbol{\beta}$ in Equation \eqref{eq:3} with $\widehat{\boldsymbol{\beta}}_t^*$ to obtain an estimate of the optimal individualized treatment rule.

\subsection{Online Algorithms}

In the context of online bandits with interference, 
we extend three algorithms from classical contextual bandits to account for the presence of interference: Linear Epsilon-Greedy With Interference (LinEGWI), Linear Upper Confidence Bound With Interference (LinUCBWI), and Linear Thompson Sampling With Interference (LinTSWI). These algorithms, summarized in Algorithm \ref{algo:LinCBWI}, differ primarily in their approach to exploration.
\begin{algorithm}[th]
\caption{Linear Contextual Bandits with Interference}\label{algo:LinCBWI}
\textbf{Input}: Number of units $N_t$; Burning period $T_0$; Interference structure $\{\boldsymbol{W}_t\}_{1\leq t\leq T}$; Clipping rate $p_t>O(\bar{N}_t^{-1/2})$.\\
\begin{algorithmic}[1] 
\FOR{Time $t=1, \cdots, T_0$}
   {\STATE $a_{ti}$ $\sim$ Bernoulli$(0.5), 1 \leq i \leq N_t$;} 
\ENDFOR
\STATE $A\leftarrow \widetilde{\boldsymbol{X}}_{1:T_0}'\widetilde{\boldsymbol{X}}_{1:T_0}$, $b\leftarrow \boldsymbol{X}_{1:T_0}'\boldsymbol{R}_{1:T_0}$;
\FOR{Time $t=T_0+1, \cdots, T$}
{
\STATE  Observe $N_t$ units with features $\{\boldsymbol{X}_{ti}\}_{1\leq i\leq N_t}$
\STATE Update $\widehat{\boldsymbol{\beta}}_{t-1}\leftarrow A^{-1}b$
\FOR{unit $i=1,2, \cdots, N_t$}
{\STATE Estimate the optimal arm 
\begin{equation*}
\widehat{\pi}_{ti} = \boldsymbol{1}\left\{\omega_{ti}\boldsymbol{X}_{ti}'(\widehat{\boldsymbol{\beta}}_{t-1,1}-\widehat{\boldsymbol{\beta}}_{t-1,0}) > 0\right\};
\end{equation*}
\IF {$\lambda_{\min}\big\{\frac{1}{\bar{N}_{t-1}}\sum_{s=1}^{t-1}\sum_{i=1}^{N_t}\widetilde{\boldsymbol{X}}_{si}\widetilde{\boldsymbol{X}}_{si}'\big\} <  p_{t-1} \cdot  \lambda_{\min}\big\{\frac{1}{\bar{N}_{t-1}}\sum_{s=1}^{t-1}\sum_{i=1}^{N_t}{\boldsymbol{X}}_{si}{\boldsymbol{X}}_{si}'\big\}$}
\STATE Choose $A_{ti}\sim\text{Bernoulli}(0.5)$, $1\leq i\leq N_t$;
\ELSE
\STATE Choose arm $a_{ti}$ by Equation \eqref{eq:EG_act}, \eqref{eq:UCB_act}, or \eqref{eq:TS_act};
\ENDIF
\STATE Receive reward $R_{ti}$;}
\ENDFOR
}
\STATE Update $\widetilde{\boldsymbol{X}}_{ti}$ by Equation \eqref{eq:X_tilde}, $\forall i\in\{1,\dots,N_t\}$
\STATE Update $A \leftarrow A + \widetilde{\boldsymbol{X}}_{t}'\widetilde{\boldsymbol{X}}_{t}$, $b \leftarrow b + \widetilde{\boldsymbol{X}}_{t}'\boldsymbol{R}_t$
\ENDFOR

\end{algorithmic}
\end{algorithm}

\subsubsection{LinEGWI} First, to generalize the classical EG algorithm in the presence of interference, we explore different arms with probability $\epsilon_{ti}$ and select the estimated optimal arm with probability $1-\epsilon_{ti}$. That is, 
\begin{equation}\label{eq:EG_act}
a_{ti}= (1-Z_{ti})\cdot \arg\max_{a} \omega_{ti}\boldsymbol{X}_{ti}'\widehat{\boldsymbol{\beta}}_{ti,a} + Z_{ti}\cdot \text{DU}(1,K),
\end{equation}
where $Z_{ti}\sim \text{Ber}(\epsilon_{ti})$, and $\text{DU}(1,K)$ denotes the discrete uniform distribution s.t. $\mathbb{P}(A=a)=\frac{1}{K}$ for any $a\in[K]$.

\subsubsection{LinUCBWI} We next consider the extension of linear UCB to interference-existing scenarios.
The key idea behind UCB is to use the variance of the parameter estimates, specifically the upper confidence bound, to guide exploration. 
In the presence of interference, this process is equivalent to comparing the UCBs of $\omega_{ti}\boldsymbol{X}_{ti}'\widehat{\boldsymbol{\beta}}_0$ and $\omega_{ti}\boldsymbol{X}_{ti}'\widehat{\boldsymbol{\beta}}_1$. Define $\widetilde{\Sigma}_{t}:= \big(\widetilde{\boldsymbol{X}}_{1:t}'\widetilde{\boldsymbol{X}}_{1:t}\big)^{-1}$. Since $\text{Var}(\widehat{\boldsymbol{\beta}}_t) = \sigma^2\cdot \widetilde{\Sigma}_{t}$, the UCB under $A_{ti}=a \in \{0,1\}$ can be derived as follows:
\begin{equation}
\begin{aligned}
    \text{UCB}_{ti,a} 
    \leftarrow \omega_{ti}\boldsymbol{X}_{ti}'\widehat{\boldsymbol{\beta}}_{t-1,a} + \alpha |\omega_{ti}|\cdot \sqrt{\boldsymbol{X}_{ti}'(\widetilde{\Sigma}_{t-1})^{-1}_{a}\boldsymbol{X}_{ti}},
\end{aligned}
\end{equation}      
where $\alpha$ is a hyperparameter that controls the exploration-exploitation tradeoff,  and $(\widetilde{\Sigma}_{t-1})^{-1}_{a}$ denotes the $d\times d$ submatrix of $(\widetilde{\Sigma}_{t-1})^{-1}$ corresponding to the variance of $\widehat{\boldsymbol{\beta}}_{t-1,a}$. 
Thus, LinUCBWI algorithm selects the arm $a_{ti}$ according to
\begin{equation}\label{eq:UCB_act}
    a_{ti}=\arg\max_{a} \text{UCB}_{ti,a}.
\end{equation}

\subsubsection{LinTSWI} In linear Thompson sampling, the prior of $\boldsymbol{\beta}$ is often pre-specified. At each round, units with transformed covariate matrix $\widetilde{\boldsymbol{X}}_{ti}$ is used to update the posterior of $\boldsymbol{\beta}$ after collecting the reward $R_{ti}$. Here, we adapt a normal prior for $\boldsymbol{\beta}$, which follows
\begin{equation}
    \begin{aligned}
    \text{(Prior)}&\quad \boldsymbol{\beta}\sim \pi(\boldsymbol{\beta}):=\mathcal{N}(\boldsymbol{\mu}_0,\boldsymbol{\Sigma}_0)\\
    \text{(Update)}&\quad R_{ti} = \widetilde{\boldsymbol{X}}_{ti}'\boldsymbol{\beta} + \epsilon_{ti}, \quad \forall i\in\{1,\dots,N_t\}
    \end{aligned}
\end{equation}

The posterior distribution of $\boldsymbol{\beta}$ given $\{\boldsymbol{X}_{1:t},\boldsymbol{A}_{1:t},\boldsymbol{R}_{1:t}\}$ can be derived as
\begin{equation*}
    f\big(\boldsymbol{\beta}|\{\boldsymbol{X}_{1:t},\boldsymbol{A}_{1:t},\boldsymbol{R}_{1:t}\}\big) \propto f(\boldsymbol{R}_{1:t}|\boldsymbol{\beta},\widetilde{\boldsymbol{X}}_{1:t})\cdot \pi(\boldsymbol{\beta}),
\end{equation*}
After simple calculations, the posterior mean and variance for $\boldsymbol{\beta}$ can be obtained by
\begin{equation*}
\begin{aligned} \boldsymbol{\Sigma}^{-1}_{t,\text{post}} &\leftarrow \boldsymbol{\Sigma}^{-1}_0 + \sum_{i,t}\widetilde{\boldsymbol{X}}_{ti}\widetilde{\boldsymbol{X}}_{ti}'/\sigma^2,\\
\boldsymbol{\beta}_{t,\text{post}} &\leftarrow \boldsymbol{\Sigma}_{t,\text{post}}\Big\{\boldsymbol{\Sigma}_0^{-1}\boldsymbol{\mu}_0+\sum_{i,t} R_{ti}\widetilde{\boldsymbol{X}}_{ti}/\sigma^2\Big\}.
\end{aligned}
\end{equation*}

Suppose $v$ is a hyper-parameter deciding the level of exploration in TS. For unit $i$ at round $t$, LinTSWI will sample $\widetilde{\boldsymbol{\beta}}_{ti}\sim \mathcal{N}(\boldsymbol{\beta}_{t,\text{post}},v^2\boldsymbol{\Sigma}^{-1}_{t,\text{post}})$ and then choose arm $a_{ti}$ such that 
\begin{equation}\label{eq:TS_act}
    a_{ti}=\arg\max_{a} \omega_{ti}\boldsymbol{X}_{ti}'\widetilde{\boldsymbol{\beta}}_{ti,a}.
\end{equation}

There are two main differences between Algorithm \ref{algo:LinCBWI} and classical linear contextual bandit algorithms. First, due to the presence of interference, $\boldsymbol{\beta}$ is estimated using the transformed covariate information $\widetilde{\boldsymbol{X}}_{ti}$. This transformation depends on the covariates, interference matrix, and actions involving all units in round $t$, as shown in Line 17 of Algorithm \ref{algo:LinCBWI}. Second, we incorporate an additional clipping step in Line 10 to ensure that the probability of exploration does not decay faster than $O(\bar{N}_t^{-1/2})$ [see Assumption \ref{assump:2}]. This clipping step is crucial for maintaining sufficient exploration, which is necessary for estimation consistency and valid inference of $\boldsymbol{\beta}$, as will be detailed in the theory section. Note that when $\boldsymbol{W}_t\equiv I$ for all $t$, our method downgrades to the classical linear contextual bandit algorithms, aside from adding a step for clipping to ensure valid statistical inference.

\section{Theory}
In this section, we provide theoretical guarantees for the probability of exploration, tail bounds, and the asymptotic distributions of the online ordinary least squares (OLS) estimator and the optimal value function. Before proceeding, we outline the key assumptions needed for the following theory.

\begin{assumption}\label{assump:1}
    (Boundedness) 
    \begin{enumerate}
    \item[a.] Define $\Sigma := \mathbb{E}\left[\boldsymbol{x}\boldsymbol{x}'\right]$ as the covariance of contextual information. There exists a constant $\lambda>0$, such that $\lambda_{\min}(\Sigma) > \lambda$.
    \item[b.] $\forall \boldsymbol{X}_{ti}\in\mathcal{X}$, there exists a constant $L_x$, such that $\|\boldsymbol{X}_{ti}\|_{2}\leq L_x$ for any $t\in[T]$ and $i\in [N_t]$.
    \item[c.] $\forall \boldsymbol{W}_t\in \mathcal{W}$, there exists a constant $L_w$, such that $\sum_j|W_{t,ij}|\leq L_w$ and $\sum_j|W_{t,ji}|\leq L_w$ for any $t\in[T]$ and $i\in [N_t]$.
\end{enumerate}
\end{assumption}

\begin{assumption}\label{assump:2}
    (Clipping) For any action $a$ and round $t$, there exists a positive and non-increasing sequence $p_t$, such that $\lambda_{\min}\big\{\frac{1}{\bar{N}_{t-1}}\sum_{s=1}^{t-1}\sum_{i=1}^{N_s}\widetilde{\boldsymbol{X}}_{si}\widetilde{\boldsymbol{X}}_{si}'\big\} >  p_{t-1} \cdot \lambda_{\min}(\Sigma)$.
\end{assumption}

\begin{assumption} \label{assump:5t}
(Margin Condition) For any $\epsilon>0$, there exists a positive constant $\gamma>0$, such that $\mathbb{P}(0<|f(\boldsymbol{X},1)-f(\boldsymbol{X},0)|<\epsilon) = O(\epsilon^{\gamma})$.
\end{assumption}

Assumption \ref{assump:1} includes several bounded conditions. Assumption \ref{assump:1}.a ensures that there is no strong collinearity between different features, which is necessary for a stable OLS estimator. This condition is commonly assumed in bandit-related inference papers \citep{zhang2020inference, chen2021statisticala, ye2023doubly}. Assumptions \ref{assump:1}.b and \ref{assump:1}.c ensure that the contextual information and the interference level for each individual unit are bounded. Assumption \ref{assump:2} is a technical requirement that guarantees the bandit algorithm explores all actions sufficiently at a rate of $p_t$, enabling consistent estimation of the OLS estimator. This exploration procedure is widely assumed in bandits inference literature \citep{deshpande2018accurate, hadad2021confidence,ye2023doubly}, which is enforced via the clipping step in Line 10 of algorithm \ref{algo:LinCBWI}. Assumption \ref{assump:5t}, known as the margin condition, is a common assumption in the contextual bandits literature \citep{audibert2007fast,luedtke2016statistical}. It ensures that the rewards obtained from pulling different arms are not too close to each other.

\subsection{Tail bound of the online OLS estimator}

\begin{theorem}\label{thm:UB}(Tail Bound of the Online OLS Estimator)
    Suppose Assumptions \ref{assump:1}-\ref{assump:2} hold. In either LinUCBWI, LinTSWI or LinEGWI, for any $h>0$, we have
    \begin{equation}\label{eq:tailbound}
        \mathbb{P}\left(\|\widehat{\boldsymbol{\beta}}_t -\boldsymbol{\beta} \|_1 >h\right)\leq  4d \exp\bigg\{-\frac{h^2\bar{N}_t p_{t}^2}{16d^3 \sigma^2 L_w^2L_x^2}\bigg\},
    \end{equation}
    where $L_w$ and $L_x$ are some constants for boundedness, and $p_t$ controls the clipping rate in Algorithm \ref{algo:LinCBWI}.
\end{theorem}

\noindent\textbf{Remark.} Given that  $d$, $\sigma$, $L_w$ and $L_x$ are positive constants, the tail bound for the online OLS estimator simplifies to $\mathbb{P}\left(\|\widehat{\boldsymbol{\beta}}_t -\boldsymbol{\beta} \|_1 >h\right)\lesssim  \exp(-h\bar{N}_t p_{t-1}^2)$. As detailed in Assumption \ref{assump:2}, $p_t$ is a non-increasing sequence. As long as $\bar{N}_t p_{t}^2\rightarrow \infty$, $\widehat{\boldsymbol{\beta}}_t$ will converge in probability to $\boldsymbol{\beta} $. Therefore, in Algorithm \ref{algo:LinCBWI}, we set the clipping rate at round $t$ to $p_t>O(\bar{N}_t^{-1/2})$ to ensure sufficient exploration and thus the convergence of the online OLS estimator.

\subsection{The probability of exploration}
Define $\kappa_{ti}(\omega_{ti},\boldsymbol{X}_{ti}) = \mathbb{P}(a_{ti}\neq \widehat{\pi}_{t-1}(\boldsymbol{X}_{ti}))$, where the probability operator $\mathbb{P}$ is taken with respect to $a_{ti}$ and all historical data collected before round $t$. The term $\kappa_{ti}(\omega_{ti},\boldsymbol{X}_{ti})$ represents the probability of exploration for unit $i$ at round $t$, depending on its contextual information and overall interference level. We define the limit of $\kappa_{ti}(\omega_{ti},\boldsymbol{X}_{ti})$ as  $\kappa_{\infty}(\omega,\boldsymbol{x}) =\lim_{\bar{N}_t\rightarrow \infty} \mathbb{P}(a_{ti}\neq \pi^*(\boldsymbol{x}))$. Since $\kappa_{ti}$ is nonnegative by definition, it follows immediately from the Sandwich Theorem that $\kappa_{\infty}$ exists for both UCB and TS. For EG, $\kappa_{\infty}(\omega,\boldsymbol{X}) = \lim_{\bar{N}_t\rightarrow \infty}\kappa_{ti}(\omega,\boldsymbol{X})=\lim_{\bar{N}_t\rightarrow \infty}\epsilon_{ti}/2$. In the following theorem, we establish the exploration upper bounds for LinUCBWI, LinTSWI, and LinEGWI, which are crucial for understanding the consistency conditions of the online OLS estimator and the necessity of clipping.

\begin{theorem} \label{thm:kappa_bound}
    Suppose Assumptions \ref{assump:1}-\ref{assump:5t} hold. In either LinUCBWI, LinTSWI or LinEGWI, for any $0<\xi<|\zeta_{ti}|/2$ with $\zeta_{ti} =\omega_{ti}\boldsymbol{X}_{ti}'({\boldsymbol{\beta}}_{1}-{\boldsymbol{\beta}}_{0})$, we have 
    \begin{enumerate}
        \item[(1)] In UCB, there exists a constant $C>0$, such that
\begin{equation}\label{eq:kappaof_UCB}
\begin{aligned}
    \kappa_{ti}(\omega_{ti},\boldsymbol{X}_{ti})  &\leq C\left(\frac{2\alpha L_wL_x}{\sqrt{\bar{N}_{t-1}p_{t-1} \lambda}}+ \xi\right)^\gamma + 8d \exp\bigg\{-\frac{\xi^2\bar{N}_{t-1} p_{t-1}^2}{64d^3 \sigma^2 L_w^4L_x^4}\bigg\}.
\end{aligned}
\end{equation}
\item[(2)] In TS, 
\begin{equation}\label{eq:kappaof_TS}
\begin{aligned}
    \kappa_{ti}(\omega_{ti},\boldsymbol{X}_{ti}) &\leq \exp\bigg\{-\frac{\bar{N}_{t-1}p_{t-1} \lambda(|{\zeta}_{ti}| - \xi)^2}{4v^2 L_w^2L_x^2}\bigg\} + 8d \exp\bigg\{-\frac{\xi^2\bar{N}_{t-1} p_{t-1}^2}{64d^3 \sigma^2 L_w^4L_x^4}\bigg\}.
\end{aligned}
\end{equation}
\item[(3)] In EG, we have $\kappa_{ti}(\omega_{ti},\boldsymbol{X}_{ti}) = \epsilon_{ti}/2$.
    \end{enumerate}
\end{theorem}

\noindent\textbf{Remark.} This theorem extends the results from \citet{ye2023doubly} to scenarios with interference. As $\bar{N}_t p_{t} \rightarrow \infty$, the exploration probability in both UCB and TS will converge to $0$ as $\bar{N}_t\rightarrow \infty$. Specifically, in UCB, the exploration upper bound consists of two components: the first term arises from the margin condition, and the second term from the tail bound of $\widehat{\boldsymbol{\beta}}_t$. When $\bar{N}_t$ is large, the second term, which decays at a rate of $O(\exp\{-\bar{N}_{t-1} p_{t-1}^2\})$, will be dominated by the first term, which decays at a rate of $O((\bar{N}_{t-1} p_{t-1})^{-\gamma/2})$ if we set $\xi = O((\bar{N}_{t-1} p_{t-1}^2)^{-1/2})$. In TS, the upper bound is dominated by the second term, which converges to $0$ at a rate $O(\exp\{-\bar{N}_{t-1} p_{t-1}^2\})$ as $\bar{N}_{t-1} p_{t-1}^2\rightarrow \infty$. Note that $L_w$ serves as an upper bound that controls the overall level of interference in Assumption \ref{assump:1}.c.  
A larger $L_w$ would increase the upper bound of exploration for both UCB and TS. However, $L_w$ has no effect on EG where the probability of exploration is often pre-specified.

\subsection{Statistical Inference on \texorpdfstring{$\boldsymbol{\beta}$}{beta}}

\begin{theorem}\label{thm:1}
Suppose Assumptions \ref{assump:1}-\ref{assump:5t} hold, and $\bar{N}_t p_{t} \rightarrow \infty$ as $\bar{N}_t \rightarrow \infty$. We have    
\begin{equation}\label{eq:thm1}
    \sqrt{\bar{N}_t}(\widehat{\boldsymbol{\beta}}_t -\boldsymbol{\beta})\xrightarrow{\mathcal{D}} \mathcal{N}(\boldsymbol{0}_{2d},\sigma^4G^{-1}), 
\end{equation}
where $G$ is specified in Equation \eqref{eq:9} in Appendix. 
\end{theorem}

\noindent\textbf{Remark.} Theorem \ref{thm:1} establishes the asymptotic normality of the online OLS estimator, providing an explicit form for its asymptotic variance. This result holds for the EG, UCB, and TS algorithms used for exploration. Despite the presence of interference, the asymptotic normality of the estimator only requires the total number of units $\bar{N}_t$ to approach infinity. In other words, bidirectional asymptotic normality is achieved as long as either $t\rightarrow \infty$ or the number of units at some stage $N_t\rightarrow \infty$.

\subsection{Statistical Inference on \texorpdfstring{$\widehat{\boldsymbol{V}}^{\pi^*}$}{V}}\label{sec:inference_V}
Suppose that the contextual information $\boldsymbol{X}_{ti}\sim \mathcal{P}_{\mathcal{X}}$ and the interference weight $\omega_{ti}\sim \mathcal{W}$. Define the oracle policy as $\pi^*(\boldsymbol{X}_{ti}) = \boldsymbol{1}\left\{\omega_{ti}\boldsymbol{X}_{ti}'(\boldsymbol{\beta}_1-\boldsymbol{\beta}_0) > 0\right\}$, and the optimal value function as $V^{\pi^*}$, which represents the expected reward under the oracle policy $\pi^*$. Specifically,
\begin{equation}
    V^{\pi^*} = \mathbb{E}\left[ \pi^*(\boldsymbol{X}) \omega\boldsymbol{\beta}_1'\boldsymbol{X}  +(1-\pi^*(\boldsymbol{X}))\omega \boldsymbol{\beta}_0'\boldsymbol{X} \right],
\end{equation}
where the expectation is taken w.r.t. $\boldsymbol{X}$ and $\omega$.


The first estimator we propose to estimate $V^{\pi^*}$ is the Inverse Probability Weighting (IPW) estimator, also known as the Importance Sampling (IS) estimator in reinforcement learning. The core idea is to use the propensity ratio, $\frac{\boldsymbol{1}\{a_{ti} = {\pi}^*(\boldsymbol{X}_{ti})\}}{\mathbb{P}\{a_{ti} = {\pi}^*(\boldsymbol{X}_{ti})\}}$, to adjust for distribution shifts caused by exploration. However, since the true values of $\pi^*$ and $\mathbb{P}\{a_{ti} = {\pi}^*(\boldsymbol{X}_{ti})\}$ are unknown, we replace them with their corresponding sample estimates. Therefore, 
\begin{equation*}
    \widehat{V}^{\text{IPW}}_t = \frac{1}{\bar{N}_t}\sum_{s=1}^t \sum_{i=1}^{N_s} \frac{\boldsymbol{1}\{a_{si} = \widehat{\pi}_{s-1}(\boldsymbol{X}_{si})\}}{1-\hat{\kappa}_{s-1} (\omega_{si}, \boldsymbol{X}_{si})} \cdot r_{si},
\end{equation*}
where $\hat{\kappa}_{t-1}(\omega_{ti}, \boldsymbol{X}_{ti}) =  \sum_{s\leq t-1,i\in[N_s]}\boldsymbol{1}\{A_{si}\neq \widehat{\pi}(\boldsymbol{X}_{si})\}/\bar{N}_{t-1}$, and $\widehat{\pi}_{s-1}(\boldsymbol{X}_{si})$ is obtained from Line 9 of Algorithm \ref{algo:LinCBWI}.

The second estimator we propose is the Direct Method (DM). The concept is straightforward: we substitute the unknown true values, such as $\pi^*$ and $\boldsymbol{\beta}$, with their sample estimates in the optimal value function $V^{\pi^*}$ to directly estimate the optimal reward. Thus,
\begin{equation*}
\begin{aligned}
    \widehat{V}^{\text{DM}}_t = \frac{1}{\bar{N}_t}\sum_{s=1}^t \sum_{i=1}^{N_s} \omega_{si}\Big\{\boldsymbol{X}_{si}'\widehat{\boldsymbol{\beta}}_{s-1,1}\widehat{\pi}_{s-1}(\boldsymbol{X}_{si})+\boldsymbol{X}_{si}'\widehat{\boldsymbol{\beta}}_{s-1,0}(1-\widehat{\pi}_{s-1}(\boldsymbol{X}_{si}))\Big\}.
\end{aligned}
\end{equation*}
Following the same logic used to derive Equation \eqref{eq:2}, the above estimator can be rewritten as
\begin{equation*}
\begin{aligned}
    \widehat{V}^{\text{DM}}_t &= \frac{1}{\bar{N}_t}\sum_{s=1}^t \sum_{i=1}^{N_s} \widehat{\mu}_{s-1}^{(i,s)}(\boldsymbol{X}_s,\widehat{\pi}_{s-1}(\boldsymbol{X}_s)),
\end{aligned}
\end{equation*}
where $\widehat{\mu}_{t-1}^{(t,i)}(\boldsymbol{X}_t,\boldsymbol{A}_t) = \widetilde{\boldsymbol{X}}_{ti}\widehat{\boldsymbol{\beta}}_{t-1}$ denotes the expected reward that unit $i$ can obtain given the covariate information and actions of all units at round $t$.

By combining the two estimators mentioned above, we derive the doubly robust (DR) estimator, where
\begin{equation*}
\begin{aligned}
    \widehat{V}^{\text{DR}}_t = \frac{1}{\bar{N}_t}\sum_{s=1}^t \sum_{i=1}^{N_s} \left[\frac{\boldsymbol{1}\{a_{si} = \widehat{\pi}_{s-1}(\boldsymbol{X}_{si})\}}{1-\hat{\kappa}_{s-1}(\omega_{ti}, \boldsymbol{X}_{si})} \cdot \Big\{r_{si} -\widehat{\mu}_{s-1}^{(i,s)}(\boldsymbol{X}_s,\widehat{\pi}_{s-1}(\boldsymbol{X}_s)) \Big\} + \widehat{\mu}_{s-1}^{(i,s)}(\boldsymbol{X}_s,\widehat{\pi}_{s-1}(\boldsymbol{X}_s))\right].
\end{aligned}
\end{equation*}
In $\widehat{V}^{\text{DR}}_t$, the second term, i.e. $\widehat{\mu}_{s-1}^{(i,s)}(\boldsymbol{X}_s,\widehat{\pi}_{s-1}(\boldsymbol{X}_s))$, corresponds to the direct estimator. The first term involving the propensity ratio is an augmentation term derived from the IPW estimator, which provides additional protection against model misspecifications, thereby ensuring double robustness. Specifically, as long as either the propensity score model $\hat{\kappa}_{t-1}$ or the outcome regression model $\widehat{\mu}_{t-1}^{(t,i)}$ is correctly specified, $\widehat{V}^{\text{DR}}_t$ becomes a consistent estimator of the optimal value function $V^{\pi^*}$. 

\begin{assumption} \label{assump:4}
(Rate Double Robustness) Define the $L_2$ norm as $\|z_t\|_{2,N_T} =\sqrt{\frac{1}{\bar{N}_T}\sum_{t=1}^T \sum_{i=1}^{N_t} z_t^2}$. We assume that the convergence rate of propensity score model $\|\hat{\kappa}_{t-1}(\omega_{ti}, \boldsymbol{X}_{ti}) - {\kappa}_{t-1}(\omega_{ti}, \boldsymbol{X}_{ti})\|_{2,N_T} = O_p(\bar{N}_T^{-\alpha_1})$, the convergence of outcome regression model $\|\widehat{\mu}_{t-1}^{(t,i)}(\boldsymbol{X}_t,\widehat{\pi}_{t-1}(\boldsymbol{X}_{t})) - {\mu}^{(t,i)}(\boldsymbol{X}_t,\widehat{\pi}_{t-1}(\boldsymbol{X}_{t}))\|_{2,N_T} = O
_p(\bar{N}_T^{-\alpha_2})$, with $\alpha_1+\alpha_2>1/2$. 
\end{assumption}

Assumption \ref{assump:4} requires that the convergence rates of the conditional mean function and the estimated probability of exploration satisfy certain conditions. This is a standard assumption in causal inference literature, as noted in \citet{luedtke2016statistical,kennedy2022semiparametric}. In our setting, this assumption is almost always satisfied, given that  $\|\widehat{\mu}_{t-1}^{(t,i)}(\boldsymbol{X}_t,\widehat{\pi}_{t-1}(\boldsymbol{X}_{t})) - {\mu}^{(t,i)}(\boldsymbol{X}_t,\widehat{\pi}_{t-1}(\boldsymbol{X}_{t}))\|_{2,N_T} = O
_p(\bar{N}_T^{-1/2})$ follows directly from Theorem \ref{thm:1}. Therefore, it suffices to ensure that $\|\hat{\kappa}_{t-1}(\omega_{ti}, \boldsymbol{X}_{ti}) - {\kappa}_{t-1}(\omega_{ti}, \boldsymbol{X}_{ti})\|_{2,N_T} = o_p(1)$ for Assumption \ref{assump:5t} to hold. This can be easily achieved by using a sample-based exploration estimand. In practice, as $\hat{\kappa}_{t-1}(\omega_{ti}, \boldsymbol{X}_{ti})$ tends to be small as $t$ increases, we set $\hat{\kappa}_{t-1}(\omega_{ti}, \boldsymbol{X}_{ti}) =  \sum_{s\leq t-1,i\in[N_s]}\boldsymbol{1}\{A_{si}\neq \widehat{\pi}(\boldsymbol{X}_{si})\}/\bar{N}_{t-1}$, which proves to be sufficient in simulation and real data analysis.

Under Assumption \ref{assump:4}, the DR estimator achieves asymptotic normality as the total number of units, $\bar{N}_t$, approaches infinity. Details are summarized in Theorem \ref{thm:2}.

\begin{theorem}\label{thm:2}
Suppose Assumptions \ref{assump:1}-\ref{assump:4} hold. We have    
\begin{equation}\label{eq:thm2}
    \sqrt{\bar{N}_t}(\widehat{V}^{\text{DR}}_t -V^{\pi^*})\xrightarrow{\mathcal{D}} \mathcal{N}(\boldsymbol{0}_{2d},\sigma^2_V), 
\end{equation}
where $\sigma^2_V$ is given by
\begin{equation}\label{eq:sigma_V}
\begin{aligned}
    \sigma^2_V &= \mathbb{E}\bigg[\frac{\sigma^2}{1-\kappa_\infty (\omega, \boldsymbol{x})}\bigg] +  \text{Var}\big[{\pi}^*(\boldsymbol{x})\cdot\omega \boldsymbol{x}'\boldsymbol{\beta}_1 +\{1-{\pi}^*(\boldsymbol{x})\}\cdot \omega\boldsymbol{x}'\boldsymbol{\beta}_0\big].
\end{aligned}
\end{equation}
\end{theorem}
\noindent \textbf{Remark.} The asymptotic variance of the optimal value function comprises two components. The first term arises from the IPW estimator and accounts for the variance of the random noise $\epsilon_{ti}$ The second term originates from the DM estimator and captures the variance due to uncertainty in the context $\boldsymbol{x}$ and the interference weight $\omega$. Notably, our theorem extends the results of \citet{ye2023doubly} by establishing the asymptotic properties of the estimated optimal value function under interference. In the special case where $\omega \equiv 1$ in Equation \eqref{eq:sigma_V}, our results reduce to theirs.

\subsection{Regret Bound}
Now we establish the regret bound for Algorithm \ref{algo:LinCBWI}. We define the regret at the end of round $T$ as 
\begin{equation*}
\begin{aligned}
    R_T &=\sum_{t=1}^T\sum_{i=1}^{N_t}\mathbb{E}\big[\mu^{(t,i)}(\boldsymbol{X}_{t},\pi^*(\boldsymbol{X}_{t})) - \mu^{(t,i)}(\boldsymbol{X}_{t},\boldsymbol{A}_{t})\big]=\sum_{t=1}^T\sum_{i=1}^{N_t}\mathbb{E}\big[\omega_{ti}\boldsymbol{X}_{ti}\boldsymbol{\beta}_{\pi^*(\boldsymbol{X}_{ti})} -\omega_{ti}\boldsymbol{X}_{ti}\boldsymbol{\beta}_{a_{ti}}\big].
\end{aligned}
\end{equation*}

\begin{theorem}\label{thm:regret_bound}
    For LinEGWI, LinUCBWI and LinTSWI in Algorithm \ref{algo:LinCBWI}, the general regret bound under interference can be derived as 
    $$R_T = \sum_{t=1}^T\sum_{i=1}^{N_t}\mathbb{E}[R^*_{ti} - R_{ti}] 
 = O(\bar{N}_T^{1/2}\log \bar{N}_T),$$ 
 which is sublinear in $\bar{N}_T$.
\end{theorem}

\noindent \textbf{Remark.} The regret upper bound can be decomposed into two components: (1) the regret due to estimation accuracy (exploitation), denoted by  $R^{(1)}_T$, and (2) the regret due to exploration, denoted by $R^{(2)}_T$. For the EG, UCB, and TS algorithms, the regret from exploitation is proven to be $o(\bar{N}_T^{-1/2})$ and is thus negligible. However, the regret due to exploration varies across algorithms. Specifically, in UCB and TS, $R^{(2)}_T$ also depends on the interference level  $L_w$, which increases as $L_w$ becomes larger. In contrast, for EG, the probability of exploration is user-specified and independent of the interference weight $\omega$. As a result, $R^{(2)}_T$ is $O(\bar{N}_T^{1/2}\log \bar{N}_T)$ by setting $\epsilon_{ti}$ properly. For detailed expressions of the upper bounds for each algorithm and the order for hyperparameters, please refer to Appendix \ref{appendix:regret_bound}.

\section{Simulation}

In this section, we first establish the asymptotic normality of $\widehat{\boldsymbol{\beta}}_t$ and $\widehat{V}^{\text{DR}}_t$ via coverage probability analysis, and then demonstrate the performance of our proposed algorithms by comparing them with baseline approaches. 

\subsection{Coverage Probability}\label{sec:cov_prob}

To demonstrate the asymptotic normality of $\boldsymbol{\beta}$ and $V^{\pi^*}$ in Theorem  \ref{thm:1}-\ref{thm:2}, we estimate the asymptotic variance and verify whether the true value of $\boldsymbol{\beta}$ and $V^{\pi^*}$ falls within the estimated confidence interval with a high probability of coverage under $B=1000$ times of replicates. By Equation \eqref{eq:thm1} and \eqref{eq:thm2}, $\boldsymbol{\beta}$ falls into the confidence region if and only if
$
    \frac{\bar{N}_t}{\sigma^4}(\widehat{\boldsymbol{\beta}} - \boldsymbol{\beta})^T G (\widehat{\boldsymbol{\beta}} - \boldsymbol{\beta})\leq \chi_{\alpha}^2(2d),
$
where $df=2d$ is the degree of freedom of the chi-square distribution. Similarly, $V^{\pi^*}$ falls into the confidence interval if and only if
$
    \sqrt{\bar{N}_t}|\widehat{V}^{\text{DR}}_t -V^{\pi^*}|\leq z_{\alpha/2}{\sigma_V}.
$
Detailed simulation setup is summarized in Appendix \ref{appendix:sim_setup1}.

\begin{figure*}[ht!]
    \centering
    \subfigure
    {
        \includegraphics[width=0.42\linewidth]{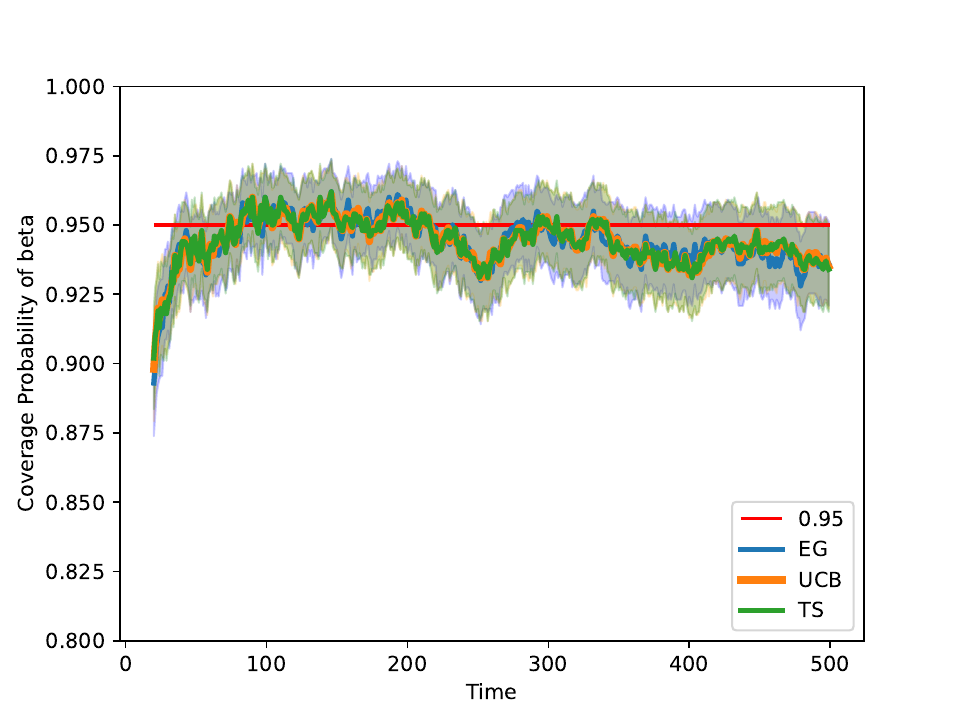}
    }
    \hspace{0.08\textwidth}
    \subfigure
    {
        \includegraphics[width=0.42\linewidth]{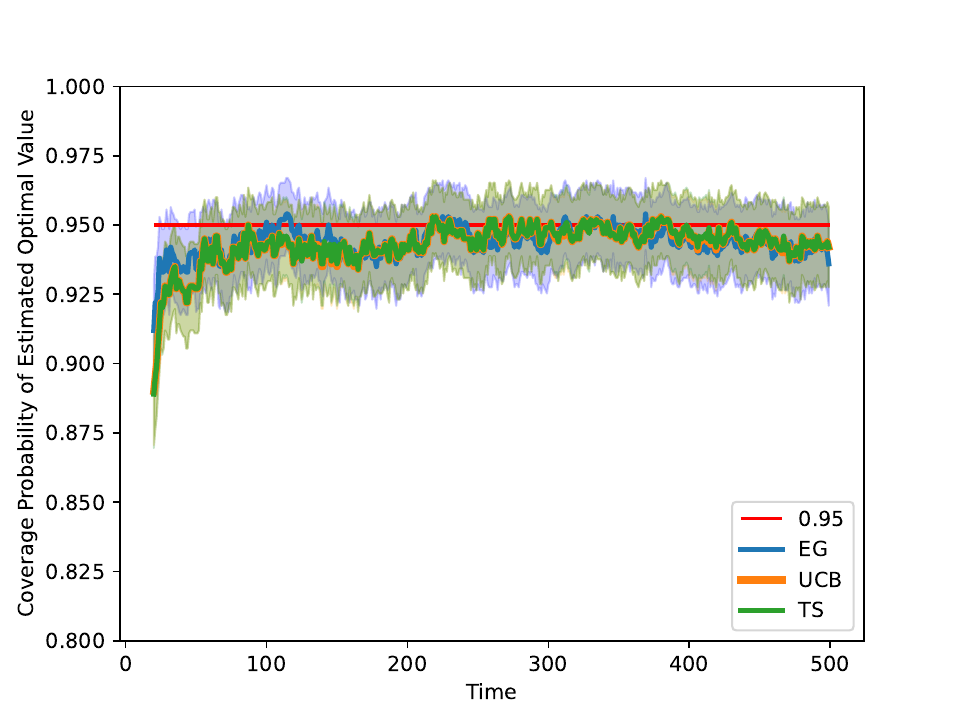}
    }
    \caption{The coverage plot of $\boldsymbol{\beta}$ (a. left) and $V^{\pi^*}$ (b. right)}\label{fig:CP_all}
\end{figure*}

Coverage probabilities of the OLS estimator $\widehat{\boldsymbol{\beta}}$ and optimal value function $V^{\pi^*}$ under three exploration algorithms (LinEGWI, LinUCBWI, and LinTSWI) are shown in Figure \ref{fig:CP_all}. As we can see, the coverage consistently hovers around $95\%$, with the estimated confidence band almost always covering the red line. This result supports the validity of the statistical inference presented in Theorems \ref{thm:1} and \ref{thm:2}.

\subsection{Comparison with Baseline Approaches}\label{sec:comp}

\begin{figure}[h!]
    \centering
    \includegraphics[width=0.55\linewidth]{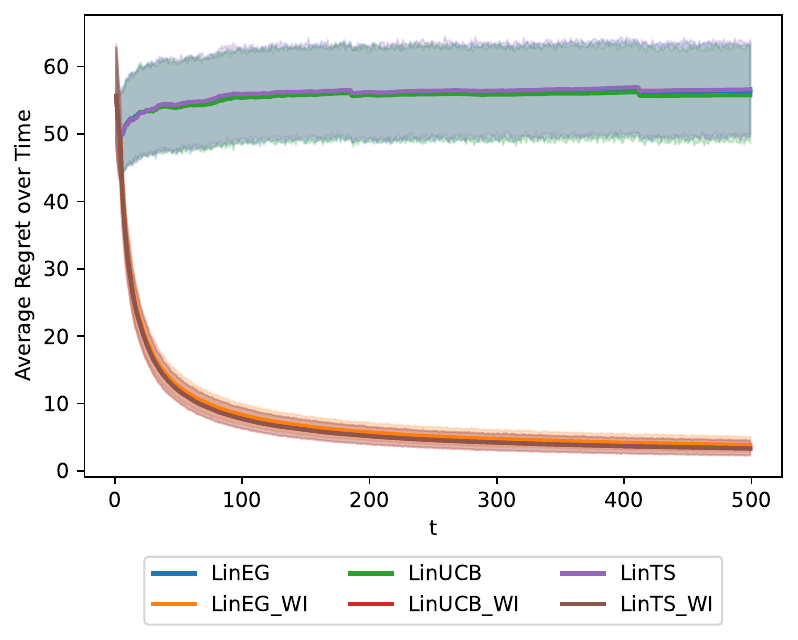}
    \caption{Comparison of average regret in the presence of interference}
    \label{fig:comp1}
\end{figure}

First, we compare our proposed method with the classical linear contextual bandit algorithms to illustrate the importance of taking interference into consideration. The results are shown in Figure \ref{fig:comp1} based on $B=100$ times of replication.   
As we can see, our approaches -- LinEGWI, LinUCBWI, and LinTSWI -- yield significantly smaller average regrets at a fast rate than classical linear contextual bandits algorithms. This demonstrates the validity of our algorithms in handling interference. When there is no interference, our algorithm reduces to the classical LinCB approach, delivering comparable results. The simulation setup and additional comparison results in the absence of interference are detailed in Appendix \ref{appendix:sim_setup2}.

\section{Synthetic Data based on MovieLens}
The MovieLens 1M dataset is a publicly available resource containing over $1$ million anonymous ratings of movies by 6k users, which aids in recommending movies to users based on their historical ratings. For each round $t$, when a user $i$ with contextual information $\boldsymbol{X}_{ti}$ visits the MovieLens website, the system recommends a movie genre ($a_{ti}$), and the user provides a rating ($R_{ti}$) for that genre. Here, we define $A = 1$ as recommending the ``Comedy'' genre and $A = 0$ as recommending the ``Drama'' genre. In this dataset, there are two types of interference that can affect the reward modeling of $R_{ti}$, which is overlooked in classical bandits settings:
\begin{enumerate}
    \item During a short time interval (which we define as a round), users often rate multiple movies they watched. This indicates that a recommendation made to a user might influence their ratings for all movies rated in the same round.
    \item Across different users in each round, there might be potential connections based on contextual information, such as occupation, ZIP code, and age. As a result, a recommendation made to one user could influence the ratings and reactions of other users in the same round.
\end{enumerate}

Based on the timestamps of each rating and the relative user density, we divided the dataset into $T=200$ rounds. For each round-unit pair $(t,i)$, $\boldsymbol{X}_{ti}\in\mathbb{R}^{d}$ is a $d=7$ dimensional vector that includes an intercept term, age, gender, and $4$ dummy variables representing the top $4$ most popular occupation types. We construct an interference matrix $\boldsymbol{W}_t$ based on the contextual information of users in the same round using normalized Jaccard similarity. Note that if a user provides multiple ratings in the same round (which is highly likely according to our observations), we treat them as ``different'' users with the same contextual information, thus the corresponding element in $\boldsymbol{W}_t$ is set to $1$. We proceed with two different pseudo-true reward generating processes. 
\begin{enumerate}
    \item[I:] For each user $j$, we calculate $\bar{R}_{j}(a)$ as the average rating of user $j$ under movie type $A=a$. Then the true reward of user $i$ at round $t$ is given by $R_{ti} = \sum_{j=1}^{N_t} W_{t,ij}\bar{R}_{j}(a)$.
    \item[II:] We fit a linear regression model to $R_{ti}$ to estimate $\boldsymbol{\beta}_0$, $\boldsymbol{\beta}_1$, and $\sigma$ as specified in Equation \eqref{eq:linear_r}. We then use these estimated values to regenerate $\widetilde{R}_{ti}$, which we assume represents the true reward of user $i$ at round $t$.
\end{enumerate}

\begin{figure*}[tbh]
    \centering
    \subfigure
    {
        \includegraphics[width=0.42\textwidth]{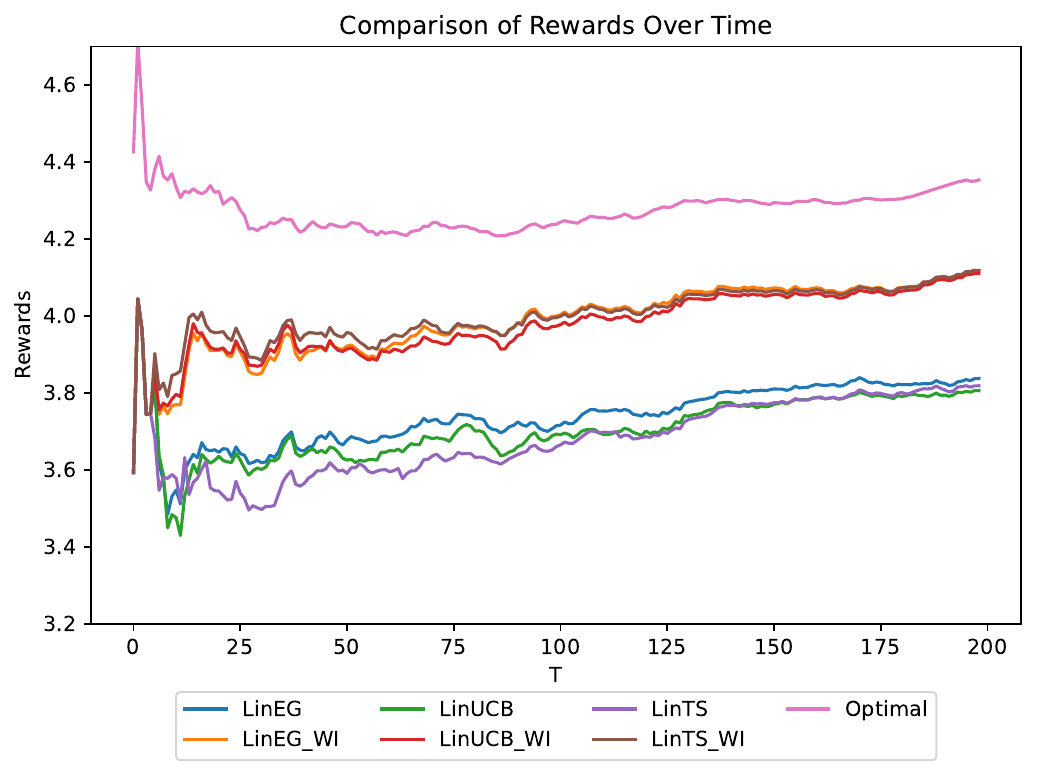}
    }
    \hspace{0.07\textwidth}
    \subfigure
    {
        \includegraphics[width=0.42\textwidth]{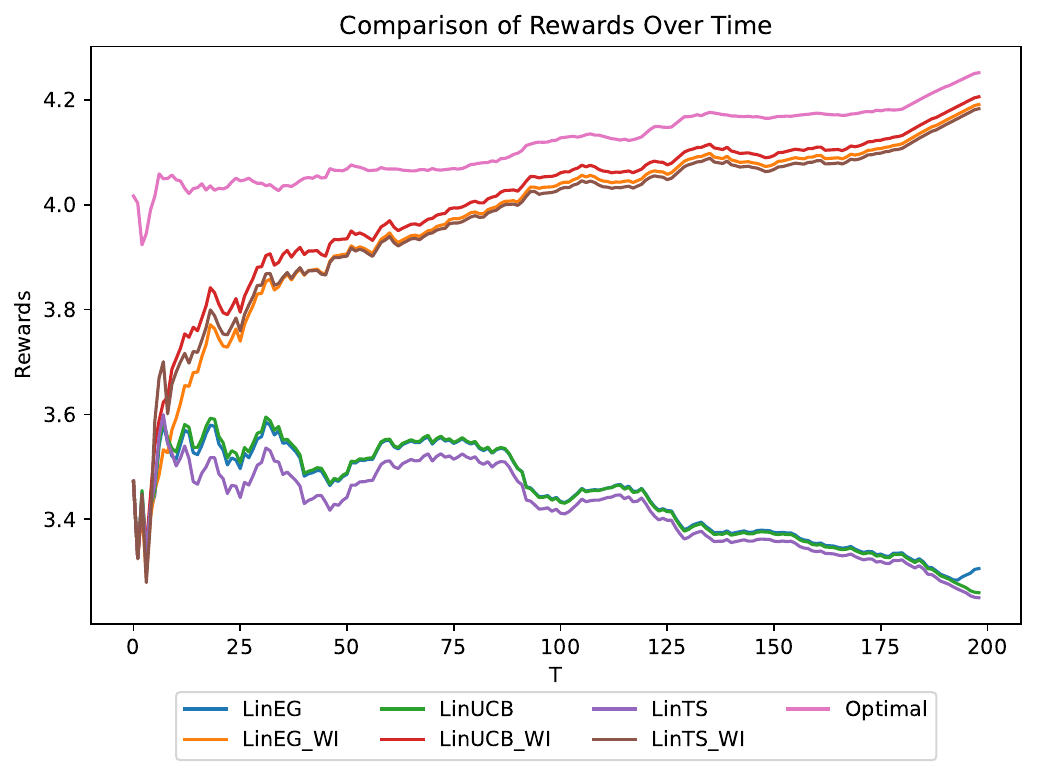}
    }
    \caption{Average rating comparison under reward generating model I (left) and II (right)}\label{fig:movielens}
\end{figure*}
The comparison results for each case are shown in Figure \ref{fig:movielens}. In both figures, our algorithms that account for interference consistently outperform classical contextual bandit approaches. Notably, in the case of reward-generating Model I, there is a gap of average reward between our algorithms and the oracle model, which disappears in reward-generating Model II. This gap likely arises because the true reward model may not be linear. Despite so, the effectiveness of our approach can be validated through both scenarios as it consistently outperforms baselines due to its ability to account for the potential interference structure.

\section{Extension to \texorpdfstring{$K>2$}{Kgeq2} Arms}
The extension to $K>2$ arms is generally straightforward from an algorithmic perspective. Specifically, when $\mathcal{A}=[K]$, one can still follow Equation \eqref{eq:linear_r} and easily extend the algorithm by modifying Line 9 of Algorithm \ref{algo:LinCBWI} to $\widehat{\pi}_{ti} = \arg\max_{a}\big\{\omega_{ti}\boldsymbol{X}_{ti}'\widehat{\boldsymbol{\beta}}_{t-1,a}\big\}$, thereby making the entire system applicable to multi-armed scenarios. 

From a theoretical perspective, extending to $K>2$  is mathematically straightforward but becomes tedious due to the nature of multi-arm comparisons. Specifically, the tail bound results would include $K$ in the denominator of the exponential term in Equation \eqref{eq:tailbound}, which does not affect the consistency result we established for the $K=2$ case. The bidirectional asymptotic normality still holds by following the martingale difference sequence we constructed, which flattens the units across different rounds into a single sequence, as detailed in Appendix \ref{appendix:proof_thm1}. The detailed derivation is beyond the scope of this paper and will be addressed in future work. 
\section{Summary}
In this paper, we pioneer a series of algorithms to address interference in linear CB, accompanied by comprehensive regret analysis, upper bound analysis, and the asymptotic properties of the online OLS estimator and the optimal value function estimator. Future work could explore several directions, including a detailed theoretical extension for $K>2$ arms, addressing scenarios where interference matrices might be unknown, and handling model misspecification using methods like weighted least squares. 

\bibliographystyle{unsrtnat}
\bibliography{Reference} 

\begin{thebibliography}{30}
\providecommand{\natexlab}[1]{#1}
\providecommand{\url}[1]{\texttt{#1}}
\expandafter\ifx\csname urlstyle\endcsname\relax
  \providecommand{\doi}[1]{doi: #1}\else
  \providecommand{\doi}{doi: \begingroup \urlstyle{rm}\Url}\fi

\bibitem[Rosenbaum(2007)]{rosenbaum2007interference}
Paul~R Rosenbaum.
\newblock Interference between units in randomized experiments.
\newblock \emph{Journal of the american statistical association}, 102\penalty0 (477):\penalty0 191--200, 2007.

\bibitem[Viviano(2020)]{viviano2020experimental}
Davide Viviano.
\newblock Experimental design under network interference.
\newblock \emph{arXiv preprint arXiv:2003.08421}, 2020.

\bibitem[Leung(2022)]{leung2022rate}
Michael~P Leung.
\newblock Rate-optimal cluster-randomized designs for spatial interference.
\newblock \emph{The Annals of Statistics}, 50\penalty0 (5):\penalty0 3064--3087, 2022.

\bibitem[Verstraeten et~al.(2020)Verstraeten, Bargiacchi, Libin, Helsen, Roijers, and Now{\'e}]{verstraeten2020multi}
Timothy Verstraeten, Eugenio Bargiacchi, Pieter~JK Libin, Jan Helsen, Diederik~M Roijers, and Ann Now{\'e}.
\newblock Multi-agent thompson sampling for bandit applications with sparse neighbourhood structures.
\newblock \emph{Scientific reports}, 10\penalty0 (1):\penalty0 6728, 2020.

\bibitem[Bargiacchi et~al.(2018)Bargiacchi, Verstraeten, Roijers, Now{\'e}, and Hasselt]{bargiacchi2018learning}
Eugenio Bargiacchi, Timothy Verstraeten, Diederik Roijers, Ann Now{\'e}, and Hado Hasselt.
\newblock Learning to coordinate with coordination graphs in repeated single-stage multi-agent decision problems.
\newblock In \emph{International conference on machine learning}, pages 482--490. PMLR, 2018.

\bibitem[Dubey et~al.(2020)]{dubey2020kernel}
Abhimanyu Dubey et~al.
\newblock Kernel methods for cooperative multi-agent contextual bandits.
\newblock In \emph{International Conference on Machine Learning}, pages 2740--2750. PMLR, 2020.

\bibitem[Agarwal et~al.(2024)Agarwal, Agarwal, Masoero, and Whitehouse]{agarwal2024multi}
Abhineet Agarwal, Anish Agarwal, Lorenzo Masoero, and Justin Whitehouse.
\newblock Multi-armed bandits with network interference.
\newblock \emph{arXiv preprint arXiv:2405.18621}, 2024.

\bibitem[Jia et~al.(2024)Jia, Frazier, and Kallus]{jia2024multi}
Su~Jia, Peter Frazier, and Nathan Kallus.
\newblock Multi-armed bandits with interference.
\newblock \emph{arXiv preprint arXiv:2402.01845}, 2024.

\bibitem[Su et~al.(2019)Su, Lu, and Song]{su2019modelling}
Lin Su, Wenbin Lu, and Rui Song.
\newblock Modelling and estimation for optimal treatment decision with interference.
\newblock \emph{Stat}, 8\penalty0 (1):\penalty0 e219, 2019.

\bibitem[Sobel(2006)]{sobel2006randomized}
Michael~E Sobel.
\newblock What do randomized studies of housing mobility demonstrate? causal inference in the face of interference.
\newblock \emph{Journal of the American Statistical Association}, 101\penalty0 (476):\penalty0 1398--1407, 2006.

\bibitem[Qu et~al.(2021)Qu, Xiong, Liu, and Imbens]{qu2021efficient}
Zhaonan Qu, Ruoxuan Xiong, Jizhou Liu, and Guido Imbens.
\newblock Efficient treatment effect estimation in observational studies under heterogeneous partial interference.
\newblock \emph{arXiv preprint arXiv:2107.12420}, 2021.

\bibitem[Hudgens and Halloran(2008)]{hudgens2008toward}
Michael~G Hudgens and M~Elizabeth Halloran.
\newblock Toward causal inference with interference.
\newblock \emph{Journal of the American Statistical Association}, 103\penalty0 (482):\penalty0 832--842, 2008.

\bibitem[Forastiere et~al.(2021)Forastiere, Airoldi, and Mealli]{forastiere2021identification}
Laura Forastiere, Edoardo~M Airoldi, and Fabrizia Mealli.
\newblock Identification and estimation of treatment and interference effects in observational studies on networks.
\newblock \emph{Journal of the American Statistical Association}, 116\penalty0 (534):\penalty0 901--918, 2021.

\bibitem[Aronow and Samii(2017)]{aronow2017estimating}
Peter~M Aronow and Cyrus Samii.
\newblock Estimating average causal effects under general interference, with application to a social network experiment.
\newblock 2017.

\bibitem[Bargagli-Stoffi et~al.(2020)Bargagli-Stoffi, Tort{\`u}, and Forastiere]{bargagli2020heterogeneous}
Falco~J Bargagli-Stoffi, Costanza Tort{\`u}, and Laura Forastiere.
\newblock Heterogeneous treatment and spillover effects under clustered network interference.
\newblock \emph{arXiv preprint arXiv:2008.00707}, 2020.

\bibitem[Manski(2013)]{manski2013identification}
Charles~F Manski.
\newblock Identification of treatment response with social interactions.
\newblock \emph{The Econometrics Journal}, 16\penalty0 (1):\penalty0 S1--S23, 2013.

\bibitem[Mart{\'\i}nez-Rubio et~al.(2019)Mart{\'\i}nez-Rubio, Kanade, and Rebeschini]{martinez2019decentralized}
David Mart{\'\i}nez-Rubio, Varun Kanade, and Patrick Rebeschini.
\newblock Decentralized cooperative stochastic bandits.
\newblock \emph{Advances in Neural Information Processing Systems}, 32, 2019.

\bibitem[Landgren et~al.(2016)Landgren, Srivastava, and Leonard]{landgren2016distributed}
Peter Landgren, Vaibhav Srivastava, and Naomi~Ehrich Leonard.
\newblock On distributed cooperative decision-making in multiarmed bandits.
\newblock In \emph{2016 European Control Conference (ECC)}, pages 243--248. IEEE, 2016.

\bibitem[Zhang et~al.(2020)Zhang, Janson, and Murphy]{zhang2020inference}
Kelly Zhang, Lucas Janson, and Susan Murphy.
\newblock Inference for batched bandits.
\newblock \emph{Advances in neural information processing systems}, 33:\penalty0 9818--9829, 2020.

\bibitem[Chen et~al.(2021)Chen, Lu, and Song]{chen2021statisticala}
Haoyu Chen, Wenbin Lu, and Rui Song.
\newblock Statistical inference for online decision making: In a contextual bandit setting.
\newblock \emph{Journal of the American Statistical Association}, 116\penalty0 (533):\penalty0 240--255, 2021.

\bibitem[Ye et~al.(2023)Ye, Cai, and Song]{ye2023doubly}
Shen Ye, Hengrui Cai, and Rui Song.
\newblock Doubly robust interval estimation for optimal policy evaluation in online learning.
\newblock \emph{Journal of the American Statistical Association}, \penalty0 (just-accepted):\penalty0 1--20, 2023.

\bibitem[Deshpande et~al.(2018)Deshpande, Mackey, Syrgkanis, and Taddy]{deshpande2018accurate}
Yash Deshpande, Lester Mackey, Vasilis Syrgkanis, and Matt Taddy.
\newblock Accurate inference for adaptive linear models.
\newblock In \emph{International Conference on Machine Learning}, pages 1194--1203. PMLR, 2018.

\bibitem[Hadad et~al.(2021)Hadad, Hirshberg, Zhan, Wager, and Athey]{hadad2021confidence}
Vitor Hadad, David~A Hirshberg, Ruohan Zhan, Stefan Wager, and Susan Athey.
\newblock Confidence intervals for policy evaluation in adaptive experiments.
\newblock \emph{Proceedings of the national academy of sciences}, 118\penalty0 (15):\penalty0 e2014602118, 2021.

\bibitem[Audibert and Tsybakov(2007)]{audibert2007fast}
Jean-Yves Audibert and Alexandre~B Tsybakov.
\newblock Fast learning rates for plug-in classifiers.
\newblock 2007.

\bibitem[Luedtke and Van Der~Laan(2016)]{luedtke2016statistical}
Alexander~R Luedtke and Mark~J Van Der~Laan.
\newblock Statistical inference for the mean outcome under a possibly non-unique optimal treatment strategy.
\newblock \emph{Annals of statistics}, 44\penalty0 (2):\penalty0 713, 2016.

\bibitem[Kennedy(2022)]{kennedy2022semiparametric}
Edward~H Kennedy.
\newblock Semiparametric doubly robust targeted double machine learning: a review.
\newblock \emph{arXiv preprint arXiv:2203.06469}, 2022.

\bibitem[Bastani and Bayati(2020)]{bastani2020online}
Hamsa Bastani and Mohsen Bayati.
\newblock Online decision making with high-dimensional covariates.
\newblock \emph{Operations Research}, 68\penalty0 (1):\penalty0 276--294, 2020.

\bibitem[Feller(1991)]{feller1991introduction}
William Feller.
\newblock \emph{An introduction to probability theory and its applications, Volume 2}, volume~81.
\newblock John Wiley \& Sons, 1991.

\bibitem[Hall and Heyde(2014)]{hall2014martingale}
Peter Hall and Christopher~C Heyde.
\newblock \emph{Martingale limit theory and its application}.
\newblock Academic press, 2014.

\bibitem[Dedecker and Louhichi(2002)]{dedecker2002maximal}
J{\'e}r{\^o}me Dedecker and Sana Louhichi.
\newblock Maximal inequalities and empirical central limit theorems.
\newblock In \emph{Empirical process techniques for dependent data}, pages 137--159. Springer, 2002.

\end{thebibliography}

\newpage
\appendix
\begin{center}
    \textbf{\LARGE Appendix}
\end{center}
\section{Simulation Setup and a Supplementary Plot Without Interference}

\subsection{Simulation Setup in Section \ref{sec:cov_prob}}\label{appendix:sim_setup1}
The simulation setup of testing coverage probability  is as follows. In the estimation of $\boldsymbol{\beta}$, the entire process is replicated for $B=1000$ times to calculate the empirical coverage. For each replication, we assume there are a total of $T=500$ rounds, and we randomly sample the true $\boldsymbol{\beta}$ from $\boldsymbol{\beta}_0=(2,-3,1)'$ and $\boldsymbol{\beta}_1=(1,1,3)'$. 

In the estimation of $\boldsymbol{\beta}$, we assume $N_t\sim \text{Poisson}(5)$ units are interacting with the environment. $\boldsymbol{X}_{ti} = (X_{ti,1},\dots,X_{ti,3})\in\mathbb{R}^3$ denotes the feature information of unit $i$ at round $t$, where $X_{ti,1} \equiv 1$, $X_{ti,2}\sim \mathcal{N}(4,1)$, $X_{ti,3}\sim \text{Unif}(0,3)$, and all of the samples are i.i.d. over $(t,i)$. At each round, we generate the interference matrix $\boldsymbol{W}_t\in \mathbb{R}^{N_t\times N_t}$ as follows. Suppose the diagonal elements $W_{ii} = 1$. For each $i>j$, we generate $W_{ij}\sim \alpha \cdot \text{Unif}(-0.6,-0.3)+ (1-\alpha) \cdot \text{Unif}(0.1,0.4)$, where $\alpha\sim \text{Bernoulli}(0.5)$. 

In the estimation of $V^{\pi*}$, for a more balanced variance composition in Equation \eqref{eq:sigma_V}, we set up the data generating process for $\boldsymbol{W}_t$ and $\boldsymbol{X}_{ti}$ as follows. For each $i\neq j$, we generate $W_{t,ij}\sim \alpha \cdot \text{Unif}(-0.2,-0.1)+ (1-\alpha) \cdot \text{Unif}(0.05,0.2)$, where $\alpha\sim \text{Bernoulli}(0.5)$.  For contextual information, we generate $\boldsymbol{X}_{ti} = (X_{ti,1},\dots,X_{ti,3})\in\mathbb{R}^3$ for unit $i$ at round $t$, where $X_{ti,1} \equiv 0.2$, $X_{ti,2}\sim \mathcal{N}(0.8,0.04)$, $X_{ti,3}\sim \text{Unif}(0,0.6)$, and all of the samples are i.i.d. over $(t,i)$.

\subsection{Simulation Setup in Section \ref{sec:comp}}\label{appendix:sim_setup2}
In reward comparison, we set a total of $T=100$ rounds, and for each round $t$, a total of $N_t\sim\text{Poisson}(\lambda)$ units will interact with the environment. We generate the interference matrix $\boldsymbol{W}_t\in \mathbb{R}^{N_t\times N_t}$ as follows: Suppose the diagonal elements $W_{ii} = 1$. For each $i>j$, we generate $W_{ij}\sim \alpha \cdot \text{Unif}(-0.9,-0.6)+ (1-\alpha) \cdot \text{Unif}(0.1,0.4)$, where $\alpha\sim \text{Bernoulli}(0.5)$. The lower triangular elements are set equivalent to the upper triangular.

Define $\boldsymbol{X}_{ti} = (X_{ti,1},\dots,X_{ti,p})\in\mathbb{R}^p$ as the feature information of unit $i$ at round $t$. Here, we let $p=5$, where the first column is intercept ${1}$, $(X_{ti,2},X_{ti,3})\sim MVN(\boldsymbol{\mu},\Sigma)$, and $(X_{ti,4},X_{ti,5})$ follows some uniform distribution. 

Following the reward generating process described in Equation \eqref{eq:linear_r}, we uniformly sample $\boldsymbol{\beta}_0\sim \text{Unif}(1,3)$ and $\boldsymbol{\beta}_1\sim \text{Unif}(-2,5)$, and replicate this process for $S=100$ times to test the robustness of different approaches w.r.t. the change of environment. All experiments were conducted on a local computer with 16 GB of memory.

\subsection{Results Comparison Without Interference}
\begin{figure}[ht!]
    \centering
    \includegraphics[width=0.55\linewidth]{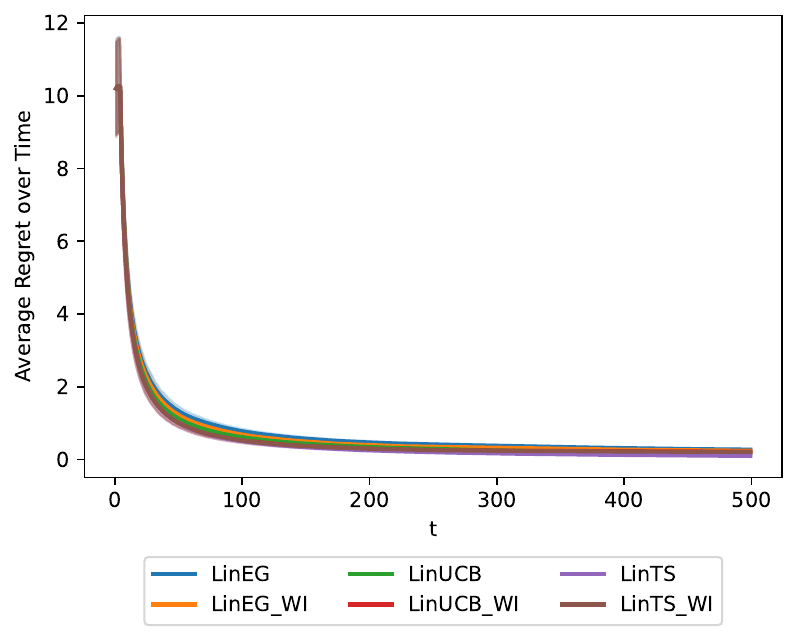}
    \caption{Comparison of average regret in the absence of interference}
    \label{fig:comp2}
\end{figure}
Using the same simulation setup as in Section \ref{sec:comp}, but with the interference matrix $\boldsymbol{W}_t$ replaced by an identity matrix, we compare the results of the classical linear CB algorithm with our proposed methods. The results, shown in Figure \ref{fig:comp2}, indicate that all methods yield comparable performance over time, with average regrets converging to zero at a rapid rate.

\section{Proof of Theorem \ref{thm:UB}: the Upper Bound of the Online OLS Estimator}

The proof of this theorem was originally presented in \citet{bastani2020online}. Here, we provide a slightly modified version to fit our specific context with interference. Define $\widetilde{\Sigma}_{t} = \frac{1}{\bar{N}_t}\sum_{s=1}^{t}\sum_{i=1}^{N_s}\widetilde{\boldsymbol{X}}_{si}\widetilde{\boldsymbol{X}}_{si}'$. According to the definition of $\widehat{\boldsymbol{\beta}}_t$,
\begin{equation*}
\begin{aligned}
    \|\widehat{\boldsymbol{\beta}}_t -\boldsymbol{\beta} \|_2& = \left\|\widetilde{\Sigma}_{t}^{-1}\cdot\bigg\{\frac{1}{\bar{N}_t}\sum_{s=1}^{t}\sum_{i=1}^{N_s}\widetilde{\boldsymbol{X}}_{si} \epsilon_{si}\bigg\}\right\|_2  \leq \left\|\widetilde{\Sigma}_{t}^{-1}\right\|_2 \cdot \left\|\bigg\{\frac{1}{\bar{N}_t}\sum_{s=1}^{t}\sum_{i=1}^{N_s}\widetilde{\boldsymbol{X}}_{si} \epsilon_{si}\bigg\}\right\|_2.
\end{aligned}
\end{equation*}
Since $\widetilde{\Sigma}_{t} $ is a symmetric positive semi-definite matrix, we have
\begin{equation*}
    \left\|\widetilde{\Sigma}_{t}^{-1}\right\|_2  = \lambda_{\max}\left(\widetilde{\Sigma}_{t}^{-1}\right) = \left\{\lambda_{\min}(\widetilde{\Sigma}_{t})\right\}^{-1},
\end{equation*}
where the right hand side of the above equation, by Assumption \ref{assump:1}-\ref{assump:2}, is lower bounded by $p_{t-1}\lambda$. Therefore,
\begin{equation*}
    \|\widehat{\boldsymbol{\beta}}_t -\boldsymbol{\beta} \|_2\leq \left\|\widetilde{\Sigma}_{t}^{-1}\right\|_2 \cdot \left\|\frac{1}{\bar{N}_t}\sum_{s=1}^{t}\sum_{i=1}^{N_s}\widetilde{\boldsymbol{X}}_{si} \epsilon_{si}\right\|_2= \left\{\lambda_{\min}(\widetilde{\Sigma}_{t})\right\}^{-1}\cdot \left\|\frac{1}{\bar{N}_t}\sum_{s=1}^{t}\sum_{i=1}^{N_s}\widetilde{\boldsymbol{X}}_{si} \epsilon_{si}\right\|_2\leq \frac{1}{\bar{N}_tp_{t-1}\lambda}\left\|\sum_{s=1}^{t}\sum_{i=1}^{N_s}\widetilde{\boldsymbol{X}}_{si} \epsilon_{si}\right\|_2.
\end{equation*}

Denote the $l$th element of $\widetilde{\boldsymbol{X}}_{ti}$ as $\widetilde{{X}}_{ti,l}$, where $l=1,\dots,2d$. For any $h>0$,
\begin{equation}\label{eq:11}
\begin{aligned}
 \mathbb{P}\left(\|\widehat{\boldsymbol{\beta}}_t -\boldsymbol{\beta} \|_2 \leq h\right)&\geq  \mathbb{P}\left(\bigg\|\sum_{s=1}^{t}\sum_{i=1}^{N_s}\widetilde{\boldsymbol{X}}_{si} \epsilon_{si}\bigg\|_2 \leq h\bar{N}_t p_{t-1}\lambda\right)\\
 & \geq \mathbb{P}\left(\bigg|\sum_{s=1}^{t}\sum_{i=1}^{N_s}\widetilde{{X}}_{si,1} \epsilon_{si}\bigg| \leq \frac{h\bar{N}_t p_{t-1}}{\sqrt{2d}},\dots, \bigg|\sum_{s=1}^{t}\sum_{i=1}^{N_s}\widetilde{{X}}_{si,2d} \epsilon_{si}\bigg| \leq \frac{h\bar{N}_t p_{t-1}}{\sqrt{2d}}\right)\\
 & = 1-\mathbb{P}\left(\bigcup_{l=1}^{2d}\left\{\bigg|\sum_{s=1}^{t}\sum_{i=1}^{N_s}\widetilde{{X}}_{si,l} \epsilon_{si}\bigg| > \frac{h\bar{N}_t p_{t-1}}{\sqrt{2d}}\right\}\right)\\
 & \geq 1- \sum_{l=1}^{2d} \mathbb{P}\left(\bigg|\sum_{s=1}^{t}\sum_{i=1}^{N_s}\widetilde{{X}}_{si,l} \epsilon_{si}\bigg| > \frac{h\bar{N}_t p_{t-1}}{\sqrt{2d}}\right).
\end{aligned}
\end{equation}
To proceed with deriving the lower bound of the above equation, we will utilize Lemma 1 from \citet{chen2021statisticala}. As this lemma is directly applicable to our context, we will state it here and refer readers to the original paper for the proof.

\begin{lemma}\label{lemma:1}
Suppose $\{\mathcal{F}_{q}:q=1,\dots,\bar{N}_T\}$ is an increasing filtration of $\sigma-$fields. Let $\{Z_{q}: q=1,\dots,\bar{N}_T\}$ be a sequence of random variables such that $Z_q$ is $\mathcal{F}_{q-1}-$measurable and $|Z_q|\leq L$. Let $\epsilon_{q}: q=1,\dots,\bar{N}_T$ be independent $\sigma-$gaussian, and $\epsilon_{q}\perp \mathcal{F}_{q-1}$ for all $q$. Let $\mathcal{S}= \{s_1,\dots,s_{|\mathcal{S}|}\}\subseteq \{1,\dots, \bar{N}_T\}$ be an index set where $|\mathcal{S}|$ is the number of elements in $\mathcal{S}$. Then for any $h>0$, 
\begin{equation} \mathbb{P}\left(\sum_{s\in\mathcal{S}}Z_s\epsilon_s \geq h\right) \leq \exp\bigg\{-\frac{h^2}{2|\mathcal{S}| \sigma^2 L^2}\bigg\}.
\end{equation}
\end{lemma}
In our context, we flatten the unit for $\{t,i\}_{1\leq t\leq T, 1\leq i \leq N_t}$ to an unit queue $Q(t,i) = \sum_{s=1}^{t-1} {N_s} + i$, such that all of the units are measured in a chronological order. As such, we also use $\widetilde{{X}}_{q,l}$ to denote the $l$th element of $\widetilde{\boldsymbol{X}}_{ti}$. To use Lemma \ref{lemma:1}, we define a filtration $\mathcal{F}_{q}$ as 
$$
\mathcal{F}_{q} = \sigma (\widetilde{{X}}_{1,l}\epsilon_{1},\dots, \widetilde{{X}}_{q,l}\epsilon_{q}),
$$
which satisfies $\epsilon_{q}\perp \mathcal{F}_{q-1}$ for any $q\in \{1,\dots, \bar{N}_T\}$. Let $Z_q = \widetilde{{X}}_{q,j}$. Then by Assumption \ref{assump:1}.b-c, 
\begin{equation*}
    |\widetilde{{X}}_{q,l}|^2 \leq \|\widetilde{\boldsymbol{X}}_{q}\|_2^2 \leq 2\Big\|\sum_{j=1}^{N_t} W_{t,ij}\boldsymbol{X}_{tj}\Big\|_2^2\leq 2L_x^2 d \cdot \Big(\sum_j |W_{t,ij}|\Big)^2 \leq 2dL_w^2 L_x^2.
\end{equation*}
Define $L:= \sqrt{2d}L_wL_x$, so that $|\widetilde{{X}}_{q,l}|\leq L$. According to the conclusion of Lemma \ref{lemma:1}, we have 
\begin{equation}\label{eq:12}
    \mathbb{P}\left(\bigg|\sum_{s=1}^{t}\sum_{i=1}^{N_s}\widetilde{{X}}_{si,l} \epsilon_{si}\bigg| \geq  \frac{h\bar{N}_t p_{t}}{\sqrt{2d}}\right) \leq \exp\bigg\{-\frac{h^2\bar{N}_t p_{t}^2}{8d^2 \sigma^2 L_w^2L_x^2}\bigg\}.
\end{equation}

Combining the result of Equation \eqref{eq:11} and \eqref{eq:12}, we have 
\begin{equation*}
\begin{aligned}
\mathbb{P}\left(\|\widehat{\boldsymbol{\beta}}_t -\boldsymbol{\beta} \|_2 \leq h\right)
 & \geq 1- \sum_{l=1}^{2d} \mathbb{P}\left(\bigg|\sum_{s=1}^{t}\sum_{i=1}^{N_s}\widetilde{{X}}_{si,l} \epsilon_{si}\bigg| > \frac{h\bar{N}_t p_{t}}{\sqrt{2d}}\right)\geq 1- \sum_{l=1}^{2d} 2\cdot \exp\bigg\{-\frac{h^2\bar{N}_t p_{t}^2}{8d^2 \sigma^2 L_w^2L_x^2}\bigg\} \\
 &= 1- 4d \exp\bigg\{-\frac{h^2\bar{N}_t p_{t}^2}{8d^2 \sigma^2 L_w^2L_x^2}\bigg\}.
\end{aligned}
\end{equation*}
Lastly, since $\|\boldsymbol{v}\|_1\leq \sqrt{2d}\|\boldsymbol{v}\|_2$ for any $\boldsymbol{v}\in \mathbb{R}^{2d}$, we have 
\begin{equation*}
\begin{aligned}
\mathbb{P}\left(\|\widehat{\boldsymbol{\beta}}_t -\boldsymbol{\beta} \|_1 \leq h\right)
 & \geq \mathbb{P}\left(\|\widehat{\boldsymbol{\beta}}_t -\boldsymbol{\beta} \|_2 \leq \frac{h}{\sqrt{2d}}\right) \geq  1- 4d \exp\bigg\{-\frac{h^2\bar{N}_t p_{t}^2}{16d^3 \sigma^2 L_w^2L_x^2}\bigg\}.
\end{aligned}
\end{equation*}
The proof of Theorem \ref{thm:UB} is thus complete.

\section{Proof of Theorem \ref{thm:kappa_bound}: the Upper Bound of Exploration}

\subsection{Probability of exploration for UCB}\label{proof:explore_UCB}
In this subsection, we aim to prove that the upper bound of $\kappa_{ti}$ for UCB algorithm satisfies
\begin{equation}
\begin{aligned}
    \kappa_{ti}(\omega_{ti},\boldsymbol{X}_{ti})  &= C\left(\frac{2\alpha L_wL_x}{\sqrt{\bar{N}_{t-1}p_{t-1} \lambda}} + \xi\right)^\gamma + 8d \exp\bigg\{-\frac{\xi^2\bar{N}_{t-1} p_{t-1}^2}{64d^3 \sigma^2 L_w^4L_x^4}\bigg\}.
    \end{aligned}
\end{equation}

We split the proof into three steps.

\noindent\textbf{Step 1:}  Rewrite $\kappa_{ti}(\omega_{ti},\boldsymbol{X}_{ti}) = \mathbb{P}\left(|\omega_{ti}\boldsymbol{X}_{ti}'(\widehat{\boldsymbol{\beta}}_{t-1,1}-\widehat{\boldsymbol{\beta}}_{t-1,0})|<
        \alpha \{\widehat{\sigma}_{t0}(\boldsymbol{X}_{ti}) - \widehat{\sigma}_{t1}(\boldsymbol{X}_{ti})\}\right)$.

\begin{equation*}
\begin{aligned}
\kappa_{ti}(\omega_{ti},\boldsymbol{X}_{ti}) &= \mathbb{P}(A_{ti}\neq \widehat{\pi}_{t-1}(\boldsymbol{X}_{ti})) \\
&= \underbrace{\mathbb{P}(A_{ti}\neq \widehat{\pi}_{t-1}(\boldsymbol{X}_{ti}) ,\widehat{\pi}_{t-1}(\boldsymbol{X}_{ti})=1)}_{\delta_1} + \underbrace{\mathbb{P}(A_{ti}\neq \widehat{\pi}_{t-1}(\boldsymbol{X}_{ti}) ,\widehat{\pi}_{t-1}(\boldsymbol{X}_{ti})=0)}_{\delta_0}.
\end{aligned}
\end{equation*}

We first consider $\delta_1$. Based on the exploration of UCB algorithm that
\begin{equation*}
    A_{ti} = \boldsymbol{1}\left\{\omega_{ti}\boldsymbol{X}_{ti}'\widehat{\boldsymbol{\beta}}_{t-1,1}
        +\alpha \widehat{\sigma}_{t-1,1}(\boldsymbol{X}_{ti}) > \omega_{ti}\boldsymbol{X}_{ti}'\widehat{\boldsymbol{\beta}}_{t-1,0}
        +\alpha \widehat{\sigma}_{t-1,0}(\boldsymbol{X}_{ti})\right\},
\end{equation*}
where $\widehat{\sigma}_{t-1,1}(\boldsymbol{X}_{ti}) = |\omega_{ti}|\sqrt{\boldsymbol{X}_{ti}'\left(\widetilde{\boldsymbol{X}}_{1:(t-1)}'\widetilde{\boldsymbol{X}}_{1:(t-1)}\right)^{-1}_{11}\boldsymbol{X}_{ti}}$, and $\widehat{\sigma}_{t-1,0}(\boldsymbol{X}_{ti}) = |\omega_{ti}|\sqrt{\boldsymbol{X}_{ti}'\left(\widetilde{\boldsymbol{X}}_{1:(t-1)}'\widetilde{\boldsymbol{X}}_{1:(t-1)}\right)^{-1}_{00}\boldsymbol{X}_{ti}}$. Thus, given that $\widehat{\pi}_{t-1}(\boldsymbol{X}_{ti})=1$, i.e. $\omega_{ti}\boldsymbol{X}_{ti}'(\widehat{\boldsymbol{\beta}}_{t-1,1}-\widehat{\boldsymbol{\beta}}_{t-1,0}) > 0$, we have
\begin{equation}\label{eq:ddelta_1}
\begin{aligned}
    \delta_1 =&\mathbb{P}(A_{ti}\neq \widehat{\pi}_{t-1}(\boldsymbol{X}_{ti}) ,\widehat{\pi}_{t-1}(\boldsymbol{X}_{ti})=1) \\
    =& \mathbb{P}(\omega_{ti}\boldsymbol{X}_{ti}'\widehat{\boldsymbol{\beta}}_{t-1,1}
        +\alpha \widehat{\sigma}_{t-1,1}(\boldsymbol{X}_{ti}) < \omega_{ti}\boldsymbol{X}_{ti}'\widehat{\boldsymbol{\beta}}_{t-1,0}
        +\alpha \widehat{\sigma}_{t-1,0}(\boldsymbol{X}_{ti}), \omega_{ti}\boldsymbol{X}_{ti}'\widehat{\boldsymbol{\beta}}_{t-1,1} > \omega_{ti}\boldsymbol{X}_{ti}'\widehat{\boldsymbol{\beta}}_{t-1,0})\\
        =&  \mathbb{P}\left(0<\omega_{ti}\boldsymbol{X}_{ti}'(\widehat{\boldsymbol{\beta}}_{t-1,1}-\widehat{\boldsymbol{\beta}}_{t-1,0})<
        \alpha \{\widehat{\sigma}_{t-1,0}(\boldsymbol{X}_{ti}) - \widehat{\sigma}_{t-1,1}(\boldsymbol{X}_{ti})\}\right).
\end{aligned}
\end{equation}
Similarly, we have 
\begin{equation}\label{eq:ddelta_0}
\begin{aligned}
    \delta_0 =&\mathbb{P}(A_{ti}\neq \widehat{\pi}_{t-1}(\boldsymbol{X}_{ti}) ,\widehat{\pi}_{t-1}(\boldsymbol{X}_{ti})=0) \\
    =& \mathbb{P}(\omega_{ti}\boldsymbol{X}_{ti}'\widehat{\boldsymbol{\beta}}_{t-1,1}
        +\alpha \widehat{\sigma}_{t-1,1}(\boldsymbol{X}_{ti}) > \omega_{ti}\boldsymbol{X}_{ti}'\widehat{\boldsymbol{\beta}}_{t-1,0}
        +\alpha \widehat{\sigma}_{t-1,0}(\boldsymbol{X}_{ti}), \omega_{ti}\boldsymbol{X}_{ti}'\widehat{\boldsymbol{\beta}}_{t-1,1} < \omega_{ti}\boldsymbol{X}_{ti}'\widehat{\boldsymbol{\beta}}_{t-1,0})\\
        =&  \mathbb{P}\left(\alpha \{\widehat{\sigma}_{t-1,0}(\boldsymbol{X}_{ti}) - \widehat{\sigma}_{t-1,1}(\boldsymbol{X}_{ti})\}<\omega_{ti}\boldsymbol{X}_{ti}'(\widehat{\boldsymbol{\beta}}_{t-1,1}-\widehat{\boldsymbol{\beta}}_{t-1,0})<
        0\right).
\end{aligned}
\end{equation}
Combining the result of Equation \eqref{eq:ddelta_1} and \eqref{eq:ddelta_0}, we have 
\begin{equation}\label{eq:thm1_step1}
    \kappa_{ti}(\omega_{ti},\boldsymbol{X}_{ti})  = \delta_1 + \delta_0 =  \mathbb{P}\left(|\omega_{ti}\boldsymbol{X}_{ti}'(\widehat{\boldsymbol{\beta}}_{t-1,1}-\widehat{\boldsymbol{\beta}}_{t-1,0})|<
        \alpha \{\widehat{\sigma}_{t-1,0}(\boldsymbol{X}_{ti}) - \widehat{\sigma}_{t-1,1}(\boldsymbol{X}_{ti})\}\right).
\end{equation}

\noindent\textbf{Step 2:}  Bound the variance terms $\widehat{\sigma}_{t-1,0}(\boldsymbol{X}_{ti})$ and $\widehat{\sigma}_{t-1,1}(\boldsymbol{X}_{ti})$.

We first consider $\widehat{\sigma}_{t-1,0}(\boldsymbol{X}_{ti})$.  Notice that
\begin{equation*}
\begin{aligned}
\widehat{\sigma}_{t-1,0}(\boldsymbol{X}_{ti})^2& =\omega_{ti}^2\boldsymbol{X}_{ti}'\left(\widetilde{\boldsymbol{X}}_{1:(t-1)}'\widetilde{\boldsymbol{X}}_{1:(t-1)}\right)^{-1}_{0}\boldsymbol{X}_{ti} \leq  \omega_{ti}^2 \|\boldsymbol{X}_{ti}\|_2^2 \cdot \max_{\|\boldsymbol{v}\|_2 = 1}\boldsymbol{v}^T\left(\widetilde{\boldsymbol{X}}_{1:(t-1)}'\widetilde{\boldsymbol{X}}_{1:(t-1)}\right)^{-1}_{0} \boldsymbol{v}\\
&\leq \omega_{ti}^2 \|\boldsymbol{X}_{ti}\|_2^2 \cdot \lambda_{\max}\left\{\left(\widetilde{\boldsymbol{X}}_{1:(t-1)}'\widetilde{\boldsymbol{X}}_{1:(t-1)}\right)^{-1}_{0}\right\}\leq L_w^2L_x^2 \cdot \lambda_{\max}\left\{\left(\widetilde{\boldsymbol{X}}_{1:(t-1)}'\widetilde{\boldsymbol{X}}_{1:(t-1)}\right)^{-1}\right\}\\
&= \frac{L_w^2L_x^2 }{\lambda_{\min}\left\{\big(\widetilde{\boldsymbol{X}}_{1:(t-1)}'\widetilde{\boldsymbol{X}}_{1:(t-1)}\big)\right\}},
\end{aligned}
\end{equation*}
where the first inequality holds by the definition of eigenvalues, and the last inequality holds by Assumption \ref{assump:1}.b-c.

Furthermore, by Assumption \ref{assump:1} and \ref{assump:2},
\begin{equation*}
    \lambda_{\min}\left\{\big(\widetilde{\boldsymbol{X}}_{1:(t-1)}'\widetilde{\boldsymbol{X}}_{1:(t-1)}\big)\right\}=\bar{N}_{t-1}\cdot \lambda_{\min}\bigg\{\frac{1}{\bar{N}_{t-1}}\sum_{s=1}^{t-1}\sum_{i=1}^{N_t}\widetilde{\boldsymbol{X}}_{si}\widetilde{\boldsymbol{X}}_{si}'\bigg\} > \bar{N}_{t-1}\cdot p_{t-1} \cdot \lambda_{\min}(\Sigma)\geq\bar{N}_{t-1}\cdot p_{t-1}  \lambda .
\end{equation*}
Combining the result above to the expression of $\widehat{\sigma}_{t-1,0}(\boldsymbol{X}_{ti})$, we can further derive
\begin{equation}\label{eq:sigma_bound}
    \widehat{\sigma}_{t-1,0}(\boldsymbol{X}_{ti}) \leq \frac{L_wL_x }{\sqrt{\lambda_{\min}\Big\{\big(\widetilde{\boldsymbol{X}}_{1:(t-1)}'\widetilde{\boldsymbol{X}}_{1:(t-1)}\big)\Big\}}} \leq \frac{L_wL_x}{\sqrt{\bar{N}_{t-1}p_{t-1} \lambda}}.
\end{equation}

Similarly, one can also show that $    0<\widehat{\sigma}_{t-1,1}(\boldsymbol{X}_{ti}) \leq \frac{L_wL_x}{\sqrt{\bar{N}_{t-1}p_{t-1} \lambda}}$. Therefore, 
\begin{equation*}
    \alpha \big|\widehat{\sigma}_{t-1,0}(\boldsymbol{X}_{ti}) - \widehat{\sigma}_{t-1,1}(\boldsymbol{X}_{ti})\big|\leq \alpha \Big\{\big|\widehat{\sigma}_{t-1,0}(\boldsymbol{X}_{ti}) \big|+ \big|\widehat{\sigma}_{t-1,1}(\boldsymbol{X}_{ti})\big|\Big\} \leq \frac{2\alpha L_wL_x}{\sqrt{\bar{N}_{t-1}p_{t-1} \lambda}}.
\end{equation*}
Combining the result above and Equation \eqref{eq:thm1_step1}, we have 
\begin{equation}\label{eq:thm1_step2}
\begin{aligned}
    \kappa_{ti}(\omega_{ti},\boldsymbol{X}_{ti})  &\leq   \mathbb{P}\left(|\omega_{ti}\boldsymbol{X}_{ti}'(\widehat{\boldsymbol{\beta}}_{t-1,1}-\widehat{\boldsymbol{\beta}}_{t-1,0})|<
    \alpha \big|\widehat{\sigma}_{t-1,0}(\boldsymbol{X}_{ti}) - \widehat{\sigma}_{t-1,1}(\boldsymbol{X}_{ti})\big|\right)\\
    & \leq \mathbb{P}\left(|\omega_{ti}\boldsymbol{X}_{ti}'(\widehat{\boldsymbol{\beta}}_{t-1,1}-\widehat{\boldsymbol{\beta}}_{t-1,0})|<
    \frac{2\alpha L_wL_x}{\sqrt{\bar{N}_{t-1}p_{t-1} \lambda}}\right)
\end{aligned}
\end{equation}

\noindent\textbf{Step 3:}  Further bound the RHS of Equation \eqref{eq:thm1_step2}. 

For the brevity of notation, we denote $\widehat{\zeta}_{ti} = \omega_{ti}\boldsymbol{X}_{ti}'(\widehat{\boldsymbol{\beta}}_{t-1,1}-\widehat{\boldsymbol{\beta}}_{t-1,0})$, and ${\zeta}_{ti} = \omega_{ti}\boldsymbol{X}_{ti}'({\boldsymbol{\beta}}_{1}-{\boldsymbol{\beta}}_{0})$.

For any $\xi >0$,  define a event $E:=\big\{|\widehat{\zeta}_{ti}-{\zeta}_{ti}| \leq \xi\big\}$. By Holder's inequality and $\|\boldsymbol{v}\|_\infty\leq \|\boldsymbol{v}\|_2$, we have 

\begin{equation*}
    |\omega_{ti}\boldsymbol{X}_{ti}'\widehat{\boldsymbol{\beta}}_{t-1,a} - \omega_{ti}\boldsymbol{X}_{ti}'{\boldsymbol{\beta}}_{a}| \leq \big\|\omega_{ti}\boldsymbol{X}_{ti}\big\|_{\infty} \left\|\widehat{\boldsymbol{\beta}}_{t-1,a} - {\boldsymbol{\beta}}_{a}\right\|_1\leq L_w\|\boldsymbol{X}_{ti}\|_2\left\|\widehat{\boldsymbol{\beta}}_{t-1,a} - {\boldsymbol{\beta}}_{a}\right\|_1\leq L_wL_x \left\|\widehat{\boldsymbol{\beta}}_{t-1,a} - {\boldsymbol{\beta}}_{a}\right\|_1.
\end{equation*}
According to Theorem \ref{thm:UB}, 
\begin{equation*}
\begin{aligned}
    \mathbb{P}\big\{|\omega_{ti}\boldsymbol{X}_{ti}'\widehat{\boldsymbol{\beta}}_{t-1,a} - \omega_{ti}\boldsymbol{X}_{ti}'{\boldsymbol{\beta}}_{a}|>\xi\big\} &\leq \mathbb{P}\Big\{L_wL_x \big\|\widehat{\boldsymbol{\beta}}_{t-1,a} - {\boldsymbol{\beta}}_{a}\big\|_1>\xi\Big\} =  \mathbb{P}\Big\{ \big\|\widehat{\boldsymbol{\beta}}_{t-1,a} - {\boldsymbol{\beta}}_{a}\big\|_1>\frac{\xi}{L_wL_x}\Big\}\\
    & \leq \mathbb{P}\left(\|\widehat{\boldsymbol{\beta}}_{t-1} -\boldsymbol{\beta} \|_1 > \frac{\xi}{L_wL_x}\right) \leq 4d \exp\bigg\{-\frac{\xi^2\bar{N}_{t-1} p_{t-1}^2}{16d^3 \sigma^2 L_w^4L_x^4}\bigg\}.
\end{aligned}
\end{equation*}
By the triangle inequality,
\begin{equation*}
\begin{aligned}
    |\widehat{\zeta}_{ti} - {\zeta}_{ti}| = &\big|\{\omega_{ti}\boldsymbol{X}_{ti}'\widehat{\boldsymbol{\beta}}_{t-1,1} - \omega_{ti}\boldsymbol{X}_{ti}'{\boldsymbol{\beta}}_{1}\} - \{\omega_{ti}\boldsymbol{X}_{ti}'\widehat{\boldsymbol{\beta}}_{t-1,0} - \omega_{ti}\boldsymbol{X}_{ti}'{\boldsymbol{\beta}}_{0}\} \big| \\
    &
    \leq \big|\omega_{ti}\boldsymbol{X}_{ti}'\widehat{\boldsymbol{\beta}}_{t-1,1} - \omega_{ti}\boldsymbol{X}_{ti}'{\boldsymbol{\beta}}_{1} \big| + \big|\omega_{ti}\boldsymbol{X}_{ti}'\widehat{\boldsymbol{\beta}}_{t-1,0} - \omega_{ti}\boldsymbol{X}_{ti}'{\boldsymbol{\beta}}_{0} \big|.
\end{aligned}
\end{equation*}
Thus, for $|\widehat{\zeta}_{ti} - {\zeta}_{ti}|$, we have 
\begin{equation*}
\begin{aligned}
    \mathbb{P}(|\widehat{\zeta}_{ti} - {\zeta}_{ti}|>\xi) &\leq \mathbb{P}\Big(\big|\omega_{ti}\boldsymbol{X}_{ti}'\widehat{\boldsymbol{\beta}}_{t-1,1} - \omega_{ti}\boldsymbol{X}_{ti}'{\boldsymbol{\beta}}_{1} \big| + \big|\omega_{ti}\boldsymbol{X}_{ti}'\widehat{\boldsymbol{\beta}}_{t-1,0} - \omega_{ti}\boldsymbol{X}_{ti}'{\boldsymbol{\beta}}_{0} \big|>\xi\Big)\\
    & \leq \mathbb{P}\Big(\big|\omega_{ti}\boldsymbol{X}_{ti}'\widehat{\boldsymbol{\beta}}_{t-1,1} - \omega_{ti}\boldsymbol{X}_{ti}'{\boldsymbol{\beta}}_{1} \big| >\xi/2\Big) + \mathbb{P}\Big(\big|\omega_{ti}\boldsymbol{X}_{ti}'\widehat{\boldsymbol{\beta}}_{t-1,0} - \omega_{ti}\boldsymbol{X}_{ti}'{\boldsymbol{\beta}}_{0} \big| >\xi/2\Big)\\
    & \leq 4d \exp\bigg\{-\frac{\xi^2\bar{N}_{t-1} p_{t-1}^2}{64d^3 \sigma^2 L_w^4L_x^4}\bigg\} + 4d \exp\bigg\{-\frac{\xi^2\bar{N}_{t-1} p_{t-1}^2}{64d^3 \sigma^2 L_w^4L_x^4}\bigg\}\\
    & = 8d \exp\bigg\{-\frac{\xi^2\bar{N}_{t-1} p_{t-1}^2}{64d^3 \sigma^2 L_w^4L_x^4}\bigg\}.
\end{aligned}
\end{equation*}

Therefore, event $E$ satisfies 
\begin{equation}\label{eq:E_bound}
    \mathbb{P}(E)\geq 1-8d \exp\bigg\{-\frac{\xi^2\bar{N}_{t-1} p_{t-1}^2}{64d^3 \sigma^2 L_w^4L_x^4}\bigg\}.
\end{equation}

On event $E$, we have $
|\widehat{\zeta}_{ti}| \geq |{\zeta}_{ti}| - |\widehat{\zeta}_{ti}-{\zeta}_{ti}| \geq |{\zeta}_{ti}| - \xi$. Then
going back to Equation \eqref{eq:thm1_step2}, we further have
\begin{equation}\label{eq:kappa_UCB}
\begin{aligned}
    \kappa_{ti}(\omega_{ti},\boldsymbol{X}_{ti})  &\leq \mathbb{P}\left(|\widehat{\zeta}_{ti}|<
    \frac{2\alpha L_wL_x}{\sqrt{\bar{N}_{t-1}p_{t-1} \lambda}}\right) \leq \mathbb{P}\left\{|\widehat{\zeta}_{ti}|<
    \frac{2\alpha L_wL_x}{\sqrt{\bar{N}_{t-1}p_{t-1} \lambda}}\text{ }\big| \text{ }E\right\} + \mathbb{P}(E^c)\\
    & \leq \mathbb{P}\left\{|{\zeta}_{ti}| - \xi<
    \frac{2\alpha L_wL_x}{\sqrt{\bar{N}_{t-1}p_{t-1} \lambda}}\right\} + 8d \exp\bigg\{-\frac{\xi^2\bar{N}_{t-1} p_{t-1}^2}{64d^3 \sigma^2 L_w^4L_x^4}\bigg\}\\
    & \leq \mathbb{P}\left\{|{\zeta}_{ti}| <
    \frac{2\alpha L_wL_x}{\sqrt{\bar{N}_{t-1}p_{t-1} \lambda}} + \xi\right\} + 8d \exp\bigg\{-\frac{\xi^2\bar{N}_{t-1} p_{t-1}^2}{64d^3 \sigma^2 L_w^4L_x^4}\bigg\}.
    \end{aligned}
\end{equation}
By definition, $|{\zeta}_{ti}| = |\omega_{ti}\boldsymbol{X}_{ti}'(\boldsymbol{\beta}_1 - \boldsymbol{\beta}_1)| = |\omega_{ti}|\cdot \big|f(\boldsymbol{X}_{ti},1)-f(\boldsymbol{X}_{ti},0)\big|$. Since $W_{t,ii}=1$, we always have $|\omega_{ti}|\geq 1$ for any round-unit pair $(t,i)$. Therefore, $|{\zeta}_{ti}|\geq \big|f(\boldsymbol{X}_{ti},1)-f(\boldsymbol{X}_{ti},0)\big|$.
According to Assumption \ref{assump:5t}, there exists some constant $\gamma$ such that $\mathbb{P}\Big\{|{\zeta}_{ti}| <
    \frac{2\alpha L_wL_x}{\sqrt{\bar{N}_{t-1}p_{t-1} \lambda}} + \xi\Big\}\leq \mathbb{P}\Big\{\big|f(\boldsymbol{X}_{ti},1)-f(\boldsymbol{X}_{ti},0)\big| <
    \frac{2\alpha L_wL_x}{\sqrt{\bar{N}_{t-1}p_{t-1} \lambda}} + \xi\Big\} \leq O\Big\{\big(\frac{2\alpha L_wL_x}{\sqrt{\bar{N}_{t-1}p_{t-1} \lambda}} + \xi\big)^\gamma\Big\}$.
    
By taking this result back to Equation \eqref{eq:kappa_UCB}, we are able to show that there exists a constant $C$, such that
\begin{equation}
\begin{aligned}
    \kappa_{ti}(\omega_{ti},\boldsymbol{X}_{ti})  & \leq \mathbb{P}\left\{|{\zeta}_{ti}| <
    \frac{2\alpha L_wL_x}{\sqrt{\bar{N}_{t-1}p_{t-1} \lambda}} + \xi\right\} + 8d \exp\bigg\{-\frac{\xi^2\bar{N}_{t-1} p_{t-1}^2}{64d^3 \sigma^2 L_w^4L_x^4}\bigg\} \\
    &= C\left(\frac{2\alpha L_wL_x}{\sqrt{\bar{N}_{t-1}p_{t-1} \lambda}} + \xi\right)^\gamma + 8d \exp\bigg\{-\frac{\xi^2\bar{N}_{t-1} p_{t-1}^2}{64d^3 \sigma^2 L_w^4L_x^4}\bigg\}.
    \end{aligned}
\end{equation}
The proof is thus complete.

\subsection{Probability of exploration for TS}

In this subsection, we aim to prove that the upper bound of $\kappa_{ti}$ for TS algorithm satisfies
\begin{equation}
    \kappa_{ti}(\omega_{ti},\boldsymbol{X}_{ti}) \leq \exp\bigg(-\frac{\bar{N}_{t-1}p_{t-1} \lambda(|{\zeta}_{ti}| - \xi)^2}{4v^2 L_w^2L_x^2}\bigg) + 8d \exp\bigg\{-\frac{\xi^2\bar{N}_{t-1} p_{t-1}^2}{64d^3 \sigma^2 L_w^4L_x^4}\bigg\}.
\end{equation}
 The proof can be split into three steps.

\noindent\textbf{Step 1:}  Decompose $\kappa_{ti}(\omega_{ti},\boldsymbol{X}_{ti})$ and bound it by $\mathbb{P}(E)$.

Similar to Step 3 of Section \ref{proof:explore_UCB}, we define an event $E:=\big\{|\widehat{\zeta}_{ti}-{\zeta}_{ti}| \leq \xi\big\}$ for any $\xi\in (0,|{\zeta}_{ti}|/2)$, where $\widehat{\zeta}_{ti} = \omega_{ti}\boldsymbol{X}_{ti}'(\widehat{\boldsymbol{\beta}}_{t1}-\widehat{\boldsymbol{\beta}}_{t0})$, and ${\zeta}_{ti} = \omega_{ti}\boldsymbol{X}_{ti}'({\boldsymbol{\beta}}_{1}-{\boldsymbol{\beta}}_{0})$. According to the result of Equation \eqref{eq:E_bound}, we have 
\begin{equation*}
    \mathbb{P}(E)\geq 1-8d \exp\bigg\{-\frac{\xi^2\bar{N}_{t-1} p_{t-1}^2}{64d^3 \sigma^2 L_w^4L_x^4}\bigg\}.
\end{equation*}
Then 
\begin{equation}\label{eq:TS_explore_S1}
\begin{aligned}
    \kappa_{ti}(\omega_{ti},\boldsymbol{X}_{ti}) &= \mathbb{P}(A_{ti}\neq \widehat{\pi}_{t-1}(\boldsymbol{X}_{ti})) \leq \mathbb{P}(A_{ti}\neq \widehat{\pi}_{t-1}(\boldsymbol{X}_{ti})|E) + \mathbb{P}(E^c)\\
    & \leq \mathbb{P}(A_{ti}\neq \widehat{\pi}_{t-1}(\boldsymbol{X}_{ti})|E) + 8d \exp\bigg\{-\frac{\xi^2\bar{N}_{t-1} p_{t-1}^2}{64d^3 \sigma^2 L_w^4L_x^4}\bigg\}.
\end{aligned}
\end{equation}
Next, we focus on bounding the first term $\mathbb{P}(A_{ti}\neq \widehat{\pi}_{t-1}(\boldsymbol{X}_{ti})|E)$. Without the loss of generality, we suppose ${\zeta}_{ti} > 0$. Then on event $E$, $0<{\zeta}_{ti} - \xi\leq \widehat{\zeta}_{ti} \leq {\zeta}_{ti} + \xi$, which implies $\widehat{\pi}_{t-1}(\boldsymbol{X}_{ti}) = 1$.
Therefore, 
\begin{equation*}
    \mathbb{P}(A_{ti}\neq \widehat{\pi}_{t-1}(\boldsymbol{X}_{ti})|E) = \mathbb{P}(A_{ti}\neq 1|E) = \mathbb{E}\Big\{\mathbb{E}\big[\boldsymbol{1}\{A_{ti} = 0\}|\widehat{\zeta}_{ti} \big]\text{ }\big|\text{ }E\Big\}.
\end{equation*}

\noindent\textbf{Step 2:}  Bound the probability of $\mathbb{E}\big[\boldsymbol{1}\{A_{ti} = 0\}|\widehat{\zeta}_{ti} \big]$ on event $E$.

Recall that in TS, we have $A_{ti} = \boldsymbol{1}\{\omega_{ti}\boldsymbol{X}_{ti}'\widetilde{\boldsymbol{\beta}}_{t-1,1} >\omega_{ti}\boldsymbol{X}_{ti}'\widetilde{\boldsymbol{\beta}}_{t-1,0}\}$, where $\widetilde{\boldsymbol{\beta}}_{t-1}\sim \mathcal{N}(\widehat{\boldsymbol{\beta}}_{t-1},v^2A_{t-1}^{-1})$ with $\widehat{\boldsymbol{\beta}}_{t} = \left(\widetilde{\boldsymbol{X}}_{1:t}'\widetilde{\boldsymbol{X}}_{1:t}\right)^{-1}\widetilde{\boldsymbol{X}}_{1:t}'\boldsymbol{R}_{1:t}$ and
$A_t =\widetilde{\boldsymbol{X}}_{1:t}'\widetilde{\boldsymbol{X}}_{1:t}$. After simple transformations, $\omega_{ti}\boldsymbol{X}_{ti}'\widetilde{\boldsymbol{\beta}}_{t-1,1} -\omega_{ti}\boldsymbol{X}_{ti}'\widetilde{\boldsymbol{\beta}}_{t-1,0}$ also follows a normal distribution with 
\begin{equation}\label{eq:dist_betatilde}
    \omega_{ti}\boldsymbol{X}_{ti}'\widetilde{\boldsymbol{\beta}}_{t-1,1} -\omega_{ti}\boldsymbol{X}_{ti}'\widetilde{\boldsymbol{\beta}}_{t-1,0} \sim \mathcal{N}\left( \omega_{ti}\boldsymbol{X}_{ti}'(\widehat{\boldsymbol{\beta}}_{t-1,1}-\widehat{\boldsymbol{\beta}}_{t-1,0}), v^2\omega_{ti}^2\boldsymbol{X}_{ti}'\mathcal{D}_{t-1}\boldsymbol{X}_{ti}\right),
\end{equation}
where $\mathcal{D}_{t-1}= \text{Var}(\widehat{\boldsymbol{\beta}}_{t-1,1} - \widehat{\boldsymbol{\beta}}_{t-1,0}) = \left(A_t^{-1}\right)_{11} + \left(A_t^{-1}\right)_{00}-2 \left(A_t^{-1}\right)_{01}$, with $\left(A_t^{-1}\right)_{11}$ denoting the upper left $d\times d$ dimensional block matrix, $\left(A_t^{-1}\right)_{00}$ denoting the bottom right $d\times d$ dimensional block matrix, and $\left(A_t^{-1}\right)_{01}$ denoting the upper right or bottom left covariance sub-matrix. For the simplicity of implementation, we exclude the interaction term $\left(A_t^{-1}\right)_{01}$ in Algorithm \ref{algo:LinCBWI}, i.e. assuming independence between $\widehat{\boldsymbol{\beta}}_{1,t}$ and $\widehat{\boldsymbol{\beta}}_{0,t}$ while making decisions. 

On event $E= \big\{|\widehat{\zeta}_{ti}-{\zeta}_{ti}| \leq \xi\big\}$, recall that ${\zeta}_{ti} > 0$ would result in $\widehat{\zeta}_{ti} > 0$ as well. According to the distribution we derived in Equation \eqref{eq:dist_betatilde}, we have
\begin{equation*}
    \begin{aligned}\mathbb{E}\big[\boldsymbol{1}\{A_{ti} = 0\}|\widehat{\zeta}_{ti} \big] &= \mathbb{P}\big\{\omega_{ti}\boldsymbol{X}_{ti}\widetilde{\boldsymbol{\beta}}_{t-1,1} <\omega_{ti}\boldsymbol{X}_{ti}\widetilde{\boldsymbol{\beta}}_{t-1,0}\mid \omega_{ti}\boldsymbol{X}_{ti}'(\widehat{\boldsymbol{\beta}}_{t-1,1}-\widehat{\boldsymbol{\beta}}_{t-1,0}) \big\}\\
    & = \Phi\Big[ -\omega_{ti}\boldsymbol{X}_{ti}'(\widehat{\boldsymbol{\beta}}_{t-1,1}-\widehat{\boldsymbol{\beta}}_{t-1,0}) \big/  \sqrt{v^2\omega_{ti}^2\boldsymbol{X}_{ti}'\mathcal{D}_{t-1}\boldsymbol{X}_{ti}} \text{ }\Big]\\
    & = 1- \Phi\Big[ \omega_{ti}\boldsymbol{X}_{ti}'(\widehat{\boldsymbol{\beta}}_{t-1,1}-\widehat{\boldsymbol{\beta}}_{t-1,0}) \big/  \sqrt{v^2\omega_{ti}^2\boldsymbol{X}_{ti}'\mathcal{D}_{t-1}\boldsymbol{X}_{ti}} \text{ }\Big],
    \end{aligned}
\end{equation*}
where $\Phi(\cdot)$ is the cumulatice distribution function of $\mathcal{N}(0,1)$. 

Denote $\widehat{z}_{ti} = \omega_{ti}\boldsymbol{X}_{ti}'(\widehat{\boldsymbol{\beta}}_{t-1,1}-\widehat{\boldsymbol{\beta}}_{t-1,0}) \big/  \sqrt{v^2\omega_{ti}^2\boldsymbol{X}_{ti}'\mathcal{D}_{t-1}\boldsymbol{X}_{ti}}$. According to the tail bound established for standard normal distribution in Section 7.1 of \citet{feller1991introduction}, we have 
\begin{equation}\label{eq:TS_explore_S2}
    \mathbb{E}\big[\boldsymbol{1}\{A_{ti} = 0\}|\widehat{\zeta}_{ti} \big] \leq \mathbb{E}\big\{\exp(-\widehat{z}_{ti}^2/2)\big\} = \mathbb{E}\bigg\{\exp\bigg(-\frac{\omega_{ti}^2\big\{\boldsymbol{X}_{ti}'(\widehat{\boldsymbol{\beta}}_{t-1,1}-\widehat{\boldsymbol{\beta}}_{t-1,0})\big\}^2}{2v^2\omega_{ti}^2\boldsymbol{X}_{ti}'\mathcal{D}_{t-1}\boldsymbol{X}_{ti}}\bigg)\bigg\}
\end{equation}

Define $\widehat{\sigma}_{t1}(\boldsymbol{X}_{ti}) = |\omega_{ti}|\sqrt{\boldsymbol{X}_{ti}'\left(\widetilde{\boldsymbol{X}}_{1:(t-1)}'\widetilde{\boldsymbol{X}}_{1:(t-1)}\right)^{-1}_{1}\boldsymbol{X}_{ti}}$, and $\widehat{\sigma}_{t0}(\boldsymbol{X}_{ti}) = |\omega_{ti}|\sqrt{\boldsymbol{X}_{ti}'\left(\widetilde{\boldsymbol{X}}_{1:(t-1)}'\widetilde{\boldsymbol{X}}_{1:(t-1)}\right)^{-1}_{0}\boldsymbol{X}_{ti}}$. According to the upper bound derived in Equation \eqref{eq:sigma_bound}, we have
\begin{equation*}
    \omega_{ti}^2\boldsymbol{X}_{ti}'\mathcal{D}_{t-1}\boldsymbol{X}_{ti} =\widehat{\sigma}_{t1}(\boldsymbol{X}_{ti})^2 + \widehat{\sigma}_{t0}(\boldsymbol{X}_{ti})^2  \leq \frac{2L_w^2L_x^2}{\bar{N}_{t-1}p_{t-1} \lambda}.
\end{equation*}
when we exclude the covariance term for simplicity, as done in Equation \eqref{eq:TS_act}.

Combining the result above to Equation \eqref{eq:TS_explore_S2}, we can further derive
\begin{equation*}
    \mathbb{E}\big[\boldsymbol{1}\{A_{ti} = 0\}|\widehat{\zeta}_{ti} \big]  \leq \mathbb{E}\bigg\{\exp\bigg(-\frac{\omega_{ti}^2\big\{\boldsymbol{X}_{ti}'(\widehat{\boldsymbol{\beta}}_{t-1,1}-\widehat{\boldsymbol{\beta}}_{t-1,0})\big\}^2}{2v^2\omega_{ti}^2\boldsymbol{X}_{ti}'\mathcal{D}_{t-1}\boldsymbol{X}_{ti}}\bigg)\bigg\} \leq \mathbb{E}\bigg\{\exp\bigg(-\frac{\bar{N}_{t-1}p_{t-1} \lambda\omega_{ti}^2\big\{\boldsymbol{X}_{ti}'(\widehat{\boldsymbol{\beta}}_{t-1,1}-\widehat{\boldsymbol{\beta}}_{t-1,0})\big\}^2}{4v^2 L_w^2L_x^2}\bigg)\bigg\}.
\end{equation*}

Note that on event $E$, $\widehat{\zeta}_{ti}^2 = \omega_{ti}^2\big\{\boldsymbol{X}_{ti}'(\widehat{\boldsymbol{\beta}}_{t-1,1}-\widehat{\boldsymbol{\beta}}_{t-1,0})\big\}^2 \geq  (|{\zeta}_{ti}| - \xi)^2$. Therefore,
\begin{equation}\label{eq:TS_explore_S3}
    \mathbb{E}\big[\boldsymbol{1}\{A_{ti} = 0\}|\widehat{\zeta}_{ti} \big]   \leq \mathbb{E}\bigg\{\exp\bigg(-\frac{\bar{N}_{t-1}p_{t-1} \lambda(|{\zeta}_{ti}| - \xi)^2}{4v^2 L_w^2L_x^2}\bigg)\bigg\} \leq \exp\bigg(-\frac{\bar{N}_{t-1}p_{t-1} \lambda(|{\zeta}_{ti}| - \xi)^2}{4v^2 L_w^2L_x^2}\bigg).
\end{equation}
\noindent\textbf{Step 3:}  Summary.

Combining the results of Equation \eqref{eq:TS_explore_S1} and \eqref{eq:TS_explore_S3}, we finally have
\begin{equation*}
\begin{aligned}
    \kappa_{ti}(\omega_{ti},\boldsymbol{X}_{ti}) &\leq \exp\bigg(-\frac{\bar{N}_{t-1}p_{t-1} \lambda(|{\zeta}_{ti}| - \xi)^2}{4v^2 L_w^2L_x^2}\bigg) + 8d \exp\bigg\{-\frac{\xi^2\bar{N}_{t-1} p_{t-1}^2}{64d^3 \sigma^2 L_w^4L_x^4}\bigg\}.
\end{aligned}
\end{equation*}
The proof is thus complete.
\section{Proof of Theorem \ref{thm:1}: the Asymptotic Normality of \texorpdfstring{$\widehat{\boldsymbol{\beta}}_t$}{betat}}\label{appendix:proof_thm1}

Recall that $\mathbb{E}[R_{ti}] = \widetilde{\boldsymbol{X}}_{ti}'\boldsymbol{\beta}$, where $
\widetilde{\boldsymbol{X}}_{ti} = (\boldsymbol{1}_{N_t}'\boldsymbol{W}_{ti}\text{diag}(\boldsymbol{1}_{N_t}-\boldsymbol{A}_t)\boldsymbol{X}_t, \boldsymbol{1}_{N_t}'\boldsymbol{W}_{ti}\text{diag}(\boldsymbol{A}_t)\boldsymbol{X}_t)'\in\mathbb{R}^{2d}$. Define the number of samples collected till the end of round t as $\bar{N}_t= \sum_{s=1}^{t}N_t$ and further define $N := \bar{N}_T$. Therefore, we estimate $\widehat{\boldsymbol{\beta}}_t$ at round $t$ by
\begin{equation}
    \widehat{\boldsymbol{\beta}}_t = \left(\widetilde{\boldsymbol{X}}_{1:t}'\widetilde{\boldsymbol{X}}_{1:t}\right)^{-1}\widetilde{\boldsymbol{X}}_{1:t}'\boldsymbol{R}_{1:t} = \left\{\frac{1}{\bar{N}_t}\sum_{s=1}^{t}\sum_{i=1}^{N_s}\widetilde{\boldsymbol{X}}_{si}\widetilde{\boldsymbol{X}}_{si}' \right\}^{-1}\left\{\frac{1}{\bar{N}_t}\sum_{s=1}^{t}\sum_{i=1}^{N_s}\widetilde{\boldsymbol{X}}_{si} R_{si}\right\}.
\end{equation} 
Since $R_{si} = \widetilde{\boldsymbol{X}}_{ti}'\boldsymbol{\beta} + \epsilon_{si}$, we can write
\begin{equation}
    \sqrt{\bar{N}_t}(\widehat{\boldsymbol{\beta}}_t - {\boldsymbol{\beta}}) = \underbrace{\left\{\frac{1}{\bar{N}_t}\sum_{s=1}^{t}\sum_{i=1}^{N_s}\widetilde{\boldsymbol{X}}_{si}\widetilde{\boldsymbol{X}}_{si}' \right\}^{-1}}_{\eta_2}\underbrace{\left\{\frac{1}{\sqrt{\bar{N}_t}}\sum_{s=1}^{t}\sum_{i=1}^{N_s}\widetilde{\boldsymbol{X}}_{si} \epsilon_{si}\right\}}_{\eta_1}
\end{equation}

\noindent\textbf{Step 1:} Show that $\eta_1 = \frac{1}{\sqrt{\bar{N}_t}}\sum_{s=1}^{t}\sum_{i=1}^{N_s}\widetilde{\boldsymbol{X}}_{si} \epsilon_{si}\xrightarrow{\mathcal{D}}\mathcal{N}(\boldsymbol{0}_{2d}, G)$.

According to Cramer-Wold device, it suffices to show that for any $\boldsymbol{v}\in\mathbb{R}^{2d}$,
\begin{equation}
    \eta_1(\boldsymbol{v}) = \frac{1}{\sqrt{\bar{N}_t}}\sum_{s=1}^{t}\sum_{i=1}^{N_s}\boldsymbol{v}'\widetilde{\boldsymbol{X}}_{si} \epsilon_{si}\xrightarrow{\mathcal{D}}\mathcal{N}(\boldsymbol{0}_{2d}, \boldsymbol{v}'G\boldsymbol{v}).
\end{equation}

Before proceeding, let's flatten the round-unit pairs $\{(t,i)\}_{1\leq t\leq T, 1\leq i \leq N_t}$ to an unit queue $Q(t,i) = \sum_{s=1}^{t-1} {N_s} + i$, such that all of the units are measured in a chronological order. Notice that the order of units in the same round doesn't matter, since the action decisions for all units in round $t$ are made at the end of that round. For any ``flattened'' unit index $q_0 =Q(i_0,t_0) \subset \{1,\dots,N\}$, 
we define $\mathcal{H}_{q_0}$ as the $\sigma$-algebra containing the information up to unit $q_0$. That is, 
$$
\mathcal{H}_{q_0} = \sigma (\boldsymbol{v}'\widetilde{\boldsymbol{X}}_{1}\epsilon_{1},\dots, \boldsymbol{v}'\widetilde{\boldsymbol{X}}_{q_0}\epsilon_{q_0}).
$$
For different indices $q$, there is a jump in information gathering for $\mathcal{H}_{q}$ whenever $q=Q(i=1,t)$ for some $t$. Since $\widetilde{\boldsymbol{X}}_{q}\in \mathcal{H}_{q}$, all of the action assignment information collected at round $t$, i.e. $\boldsymbol{A}_t$, is contained in $\mathcal{H}_{q}$ at the beginning of this round. With a slight abuse of notation, in the following proof, we will also use $\mathcal{H}_{t}$ to denote all historical data collected up to round $t$.

The tricky part of establishing asymptotic porperties for $\widehat{\boldsymbol{\beta}}_t$ lies in the data dependence. Specifically, the transformed covariate vector $\widetilde{\boldsymbol{X}}_{si}$ is a function of $(\boldsymbol{W}_t,\boldsymbol{A}_s,{\boldsymbol{X}}_{s})$, thus depending on all of the actions and original covaraites information collected at round $t$. As such, $\widetilde{\boldsymbol{X}}_{si}\epsilon_{si}\not\perp \widetilde{\boldsymbol{X}}_{i's'}\epsilon_{i's'}$ for any $(s,s')$, since (1) if $s=s'$, units in the same round $s$ are correlated by $\boldsymbol{W}_s$; (2) if $s<s'$, unit are dependent since the later decisions made on $\boldsymbol{A}_{s'}$ will depend on $(\boldsymbol{W}_s, \boldsymbol{X}_{s},\boldsymbol{A}_{s})$.

Now let's use Martingale CLT to establish the asymptotic properties. We will prove shortly that $\{\boldsymbol{v}'\widetilde{\boldsymbol{X}}_{si}\epsilon_{si}\}$, or equivalently $\{\boldsymbol{v}'\widetilde{\boldsymbol{X}}_{q}\epsilon_{q}\}$ after flattening, is a Martingale difference sequence. That is, we would like to show
\begin{equation}
    \mathbb{E}[\boldsymbol{v}'\widetilde{\boldsymbol{X}}_{q}\epsilon_{q}|\mathcal{H}_{q-1}] = 0, \quad \forall q\in\{1,\dots,N\}.
\end{equation}
Suppose $q=  Q(t,i)$ for some $(t,i)$ pair. According to our assumption on the noise term in the main paper, $\epsilon_{ti}\perp ({\boldsymbol{X}}_t,\boldsymbol{W}_t) | \boldsymbol{A}_t \text{ }\Rightarrow\text{ } \epsilon_{ti}\perp \widetilde{\boldsymbol{X}}_t | \boldsymbol{A}_t$ as $\widetilde{\boldsymbol{X}}_t$ is a function of $(\boldsymbol{W}_t,{\boldsymbol{X}}_t,{\boldsymbol{A}}_t)$.
$$
\mathbb{E}[\boldsymbol{v}'\widetilde{\boldsymbol{X}}_{q}\epsilon_{q}|\mathcal{H}_{q-1}] = \mathbb{E}\left[\mathbb{E}[\boldsymbol{v}'\widetilde{\boldsymbol{X}}_{q}\epsilon_{q}|\mathcal{H}_{q-1},\boldsymbol{A}_{t}]|\mathcal{H}_{q-1}\right]\stackrel{(\text{A1})}{=}
\mathbb{E}\bigg[\mathbb{E}[\boldsymbol{v}'\widetilde{\boldsymbol{X}}_{q}|\mathcal{H}_{q-1},\boldsymbol{A}_{t}]\cdot \underbrace{\mathbb{E}[\epsilon_{q}|\mathcal{H}_{q-1},\boldsymbol{A}_{t}]}_{=0}|\mathcal{H}_{q-1}\bigg],
$$

Now it suffice to (1) check the Lindeberg condition, and (2) derive the limit of conditional variance.

\textbf{(1) We first check the Lindeberg condition.}\\
For any $\delta>0$, we define
\begin{equation}
    \psi = \sum_{q=1}^{\bar{N}_q} \mathbb{E}\left[\frac{1}{\bar{N}_q}(\boldsymbol{v}'\widetilde{\boldsymbol{X}}_{q})^2 \epsilon_{q}^2\cdot \boldsymbol{1}\bigg\{\Big|\frac{1}{\sqrt{\bar{N}_q}}\boldsymbol{v}'\widetilde{\boldsymbol{X}}_{q} \epsilon_{q}\Big|>\delta\bigg\}\text{ }\big|\text{ } \mathcal{H}_{q-1}\right]
\end{equation}

According to Assumption \ref{assump:1}.b-c,
\begin{equation}\label{eq:7}
    (\boldsymbol{v}'\widetilde{\boldsymbol{X}}_{q})^2\leq  \|\boldsymbol{v}\|_2^2\cdot \|\widetilde{\boldsymbol{X}}_{q}\|_2^2\leq 2 \|\boldsymbol{v}\|_2^2\cdot \Big\|\sum_{j=1}^{N_t}W_{t,ij}\boldsymbol{X}_{tj}\Big\|^2_2\leq 2\|\boldsymbol{v}\|_2^2\cdot L_x^2d\cdot \Big(\sum_j|W_{t,ij}|\Big)^2 \leq 2dL_w^2L_x^2\|\boldsymbol{v}\|_2^2.
\end{equation}
Then
\begin{equation*}
    \boldsymbol{1}\bigg\{\Big|\frac{1}{\sqrt{\bar{N}_q}}\boldsymbol{v}'\widetilde{\boldsymbol{X}}_{q} \epsilon_{q}\Big|>\delta\bigg\}
    \leq \boldsymbol{1}\bigg\{2dL_w^2 L_x^2\|\boldsymbol{v}\|_2^2 \epsilon_{q}^2>\bar{N}_q\delta^2\bigg\}
    = \boldsymbol{1}\bigg\{\epsilon_{q}^2>\frac{\bar{N}_q\delta^2}{2dL_w^2 L_x^2\|\boldsymbol{v}\|_2^2 }\bigg\}.
\end{equation*}
Thus, 
\begin{equation}\label{eq:13}
    \psi \leq \frac{2dL_w^2 L_x^2 \|\boldsymbol{v}\|_2^2}{\bar{N}_q}\sum_{q=1}^{\bar{N}_q} \mathbb{E}\left[\epsilon_{q}^2\cdot \boldsymbol{1}\bigg\{\epsilon_{q}^2>\frac{\bar{N}_q\delta^2}{2dL_w^2 L_x^2\|\boldsymbol{v}\|_2^2 }\bigg\}\text{ }\big|\text{ } \mathcal{H}_{q-1}\right].
\end{equation}

Define $f_{\bar{N}_q}=\frac{2dL_w^2 L_x^2 \|\boldsymbol{v}\|_2^2}{\bar{N}_q}\sum_{q=1}^{\bar{N}_q} \epsilon_{q}^2\cdot \boldsymbol{1}\bigg\{\epsilon_{q}^2>\frac{\bar{N}_q\delta^2}{2dL_w^2 L_x^2\|\boldsymbol{v}\|_2^2 }\bigg\}$, and $g_{\bar{N}_q}=\frac{2dL_w^2 L_x^2 \|\boldsymbol{v}\|_2^2}{\bar{N}_q}\sum_{q=1}^{\bar{N}_q} \epsilon_{q}^2$. It is obvious that $|f_{\bar{N}_q}|\leq g_{\bar{N}_q}$ a.s. and for all $q$. Since 
$$
\mathbb{E}[\epsilon_q^2|\mathcal{H}_{q-1}] = \mathbb{E}\left[\mathbb{E}[\epsilon_q^2|\boldsymbol{A}_t,\mathcal{H}_{q-1}]|\mathcal{H}_{q-1}\right] =\sigma^2 < \infty,
$$
we have 
$$
\mathbb{E}[g_{\bar{N}_q}|\mathcal{H}_{q-1}] \frac{2dL_w^2 L_x^2 \|\boldsymbol{v}\|_2^2}{\bar{N}_q}\sum_{q=1}^{\bar{N}_q} \sigma^2\leq 2dL_w^2 L_x^2 \|\boldsymbol{v}\|_2^2\sigma^2<\infty,
$$
thus $g_{\bar{N}_q}$ is integrable for all $q$. 

For each realization of random variable sequence $\{\epsilon_q\}_{ q=1}^{\infty}$, $\lim_{\bar{N}_q\rightarrow\infty}f_{\bar{N}_q}=0$ as $\boldsymbol{1}\bigg\{\epsilon_{q}^2>\frac{\bar{N}_q\delta^2}{2dL_w^2 L_x^2\|\boldsymbol{v}\|_2^2 }\bigg\}=0$ when $\bar{N}_q$ is large enough.

Therefore, by Generalized Dominated Convergence Theorem (GDCT), it follows from Equation \eqref{eq:13} that $\psi \leq \mathbb{E}[f_{\bar{N}_q}|\mathcal{H}_{q-1}]\rightarrow 0$ as $q\rightarrow\infty$. The Lindeberg condition is thus verified.

\textbf{(2) We next derive the limit of conditional variance.}\\
\begin{equation}
\begin{aligned}
    \frac{1}{\bar{N}_q} \sum_{s=1}^t \sum_{i=1}^{N_s} \mathbb{E}\big[(\boldsymbol{v}'\widetilde{\boldsymbol{X}}_{q})^2 \epsilon_{q}^2|\mathcal{H}_{q-1}\big] &= \frac{1}{\bar{N}_q} \sum_{s=1}^t \sum_{i=1}^{N_s} \mathbb{E}\big[(\boldsymbol{v}'\widetilde{\boldsymbol{X}}_{q})^2 \mathbb{E}[\epsilon_{q}^2|\boldsymbol{A}_i]|\mathcal{H}_{q-1}\big]\\
    & = \frac{1}{\bar{N}_q} \sum_{s=1}^t \sum_{i=1}^{N_s} \sigma^2\mathbb{E}\big[(\boldsymbol{v}'\widetilde{\boldsymbol{X}}_{q})^2 |\mathcal{H}_{q-1}\big]
\end{aligned}
\end{equation}

where the last equality holds since $\epsilon_{q}^2|\boldsymbol{A}_i$ i.i.d. follows $\mathcal{N}(0,\sigma^2)$. 

Recall that for any unit index $q=Q(t,i)$, 
$$
\widetilde{\boldsymbol{X}}_{q} =\widetilde{\boldsymbol{X}}_{ti}= (\boldsymbol{1}_{N_t}'\boldsymbol{W}_{ti}\text{diag}(\boldsymbol{1}\{{A}_{ti} = 0\}_{1\leq i\leq N_t})\boldsymbol{X}_t, \boldsymbol{1}_{N_t}'\boldsymbol{W}_{ti}\text{diag}(\boldsymbol{1}\{{A}_{ti} = 1\}_{1\leq i\leq N_t})\boldsymbol{X}_t)'.
$$
After some manipulations, we have
\begin{equation}
    \begin{aligned}
        &\widetilde{\boldsymbol{X}}_{q}\widetilde{\boldsymbol{X}}_{q}' = \widetilde{\boldsymbol{X}}_{ti}\widetilde{\boldsymbol{X}}_{ti}' := \begin{bmatrix} M_1& M_2\\ M_3 & M_4\\\end{bmatrix}\in\mathbb{R}^{2d\times 2d}\\
        &=\begin{bmatrix}
    \sum_{k=1}^{N_t}\sum_{l=1}^{N_t} W_{t,ik}W_{t,il}\boldsymbol{X}_{tk}\boldsymbol{X}_{tl}'\cdot \boldsymbol{1}_{A_{tk}}(0)\boldsymbol{1}_{A_{tl}}(0) & \sum_{k=1}^{N_t}\sum_{l=1}^{N_t} W_{t,ik}W_{t,il}\boldsymbol{X}_{tk}\boldsymbol{X}_{tl}'\cdot \boldsymbol{1}_{A_{tk}}(0)\boldsymbol{1}_{A_{tl}}(1) \\
    \sum_{k=1}^{N_t}\sum_{l=1}^{N_t} W_{t,ik}W_{t,il}\boldsymbol{X}_{tk}\boldsymbol{X}_{tl}'\cdot \boldsymbol{1}_{A_{tk}}(1)\boldsymbol{1}_{A_{tl}}(0) & \sum_{k=1}^{N_t}\sum_{l=1}^{N_t} W_{t,ik}W_{t,il}\boldsymbol{X}_{tk}\boldsymbol{X}_{tl}'\cdot \boldsymbol{1}_{A_{tk}}(1)\boldsymbol{1}_{A_{tl}}(1) \\
\end{bmatrix},
    \end{aligned}
\end{equation}
where $\boldsymbol{1}_{A_{ti}}(a) = \boldsymbol{1}\{A_{ti} = a\}$. That is, $\widetilde{\boldsymbol{X}}_{q}\widetilde{\boldsymbol{X}}_{q}'$ can be decomposed as 4 $d$-by-$d$ block matrices, each differs in the action assignment of $(A_{tk},A_{tl})$. 

Take the upper left submatrix $M_1 := \sum_{k=1}^{N_t}\sum_{l=1}^{N_t} W_{t,ik}W_{t,il}\boldsymbol{X}_{tk}\boldsymbol{X}_{tl}'\cdot \boldsymbol{1}_{A_{tk}}(0)\boldsymbol{1}_{A_{tl}}(0)$ as an example. Since we assume $\boldsymbol{X}_{ti}\sim\mathcal{X}$ and $\boldsymbol{W}_t\sim \mathcal{W}$ are known to us, the main challenge of deriving the conditional variance lies in estimating the conditional expectation of $\mathbb{E}[\boldsymbol{1}_{A_{tk}}(0)\boldsymbol{1}_{A_{tl}}(0)|\boldsymbol{W}_t,\boldsymbol{X}_t,\mathcal{H}_{q-1}]$ for any $q\in\{1,\dots,\bar{N}_q\}$.

Recall that $\mathcal{H}_{q-1}$ is defined as the $\sigma$-algebra containing the information up to unit $q-1$. That is, 
$$
\mathcal{H}_{q-1} = \sigma (\boldsymbol{v}'\widetilde{\boldsymbol{X}}_{1}\epsilon_{1},\dots, \boldsymbol{v}'\widetilde{\boldsymbol{X}}_{q-1}\epsilon_{q-1}).
$$
For different indices $q$, there is a jump in information gathering for $\mathcal{H}_{q}$ whenever $q=Q(i=1,t)$ for some $t$. Since $\widetilde{\boldsymbol{X}}_{q}\in \mathcal{H}_{q}$, all of the action assignment information collected at round $t$, i.e. $\boldsymbol{A}_t$, is contained in $\mathcal{H}_{q}$ at the beginning of this round. Thanks to the property of $\epsilon_q$ that $\mathbb{E}[\epsilon_q|\boldsymbol{A}_t]=0$, the conditional variance $\mathbb{E}\big[(\boldsymbol{v}'\widetilde{\boldsymbol{X}}_{q})^2 \epsilon^2|\mathcal{H}_{q-1}\big] = \mathbb{E}\big[(\boldsymbol{v}'\widetilde{\boldsymbol{X}}_{q})^2\epsilon^2 \big] = \mathbb{E}\big[(\boldsymbol{v}'\widetilde{\boldsymbol{X}}_{q})^2\epsilon^2|\mathcal{H}_{t-1} \big]$ for any $q=Q(t,i)$. Still, we take the upper left $d\times d$ sub-matrix as an example to calculate the asymptotic variance.

\begin{equation}\label{eq:5}
\begin{aligned}
    \mathbb{E}[M_1]&=\mathbb{E}\bigg[\sum_{k=1}^{N_t}\sum_{l=1}^{N_t} W_{t,ik}W_{t,il}\boldsymbol{X}_{tk}\boldsymbol{X}_{tl}'\cdot \boldsymbol{1}_{A_{tk}}(0)\boldsymbol{1}_{A_{tl}}(0) \bigg]\\
    &=  \mathbb{E}\Bigg[\mathbb{E}\bigg[\sum_{k=1}^{N_t}\sum_{l=1}^{N_t} W_{t,ik}W_{t,il}\boldsymbol{X}_{tk}\boldsymbol{X}_{tl}'\cdot \boldsymbol{1}\{{A}_{tk} = 0\}\boldsymbol{1}\{{A}_{tl} = 0\} \big| \boldsymbol{W}_t, \boldsymbol{X}_t,\mathcal{H}_{t-1}\bigg]\Bigg]\\
    &=\mathbb{E}\Bigg[\sum_{k=1}^{N_t}\sum_{l=1}^{N_t} W_{t,ik}W_{t,il}\boldsymbol{X}_{tk}\boldsymbol{X}_{tl}'\cdot\mathbb{E}\big[ \boldsymbol{1}\{{A}_{tk} = 0\}\boldsymbol{1}\{{A}_{tl} = 0\} \big|\boldsymbol{W}_t, \boldsymbol{X}_t,\mathcal{H}_{t-1}\big]\Bigg].
\end{aligned}
\end{equation}
Notice that $A_{tk}\perp A_{tl}|\boldsymbol{W}_t, \boldsymbol{X}_t,\mathcal{H}_{t-1}$ for any $(k,l)\in\{1,\dots,N_t\}$ and $k\neq l$. This independence arises because the actions assigned to units $Q(k,t)$ and $Q(l,t)$ are determined by two factors: exploitation and exploration.
\begin{enumerate}
    \item \textbf{Exploitation}: The action $A_{tk}$ is partially determined by $\hat{\pi}_{t-1}(X_{tk})$, where $\hat{\pi}_{t-1}$ is a function of $\mathcal{H}_{t-1}$ and is obtained by fitting a model to data from the first $t-1$ rounds. Therefore, given $(\boldsymbol{W}_t,\boldsymbol{X}_t,\mathcal{H}_{t-1})$,  $\hat{\pi}_{t-1}(\boldsymbol{X}_{tk})$ and $\hat{\pi}_{t-1}(\boldsymbol{X}_{tl})|\mathcal{H}_{t-1}$ are both constants and thus independent from each other.
    \item \textbf{Exploration}: The action $A_{tk}$ is also influenced by a specific exploration method based on the ``optimal'' action identified during exploitation. In $\epsilon$-greedy, the level of exploration is determined by $\epsilon_t$, which is independently assigned to each unit. For UCB and TS, the exploration level for each unit is a function of $\mathcal{H}_{t-1}$, making them mutually independent given $\mathcal{H}_{t-1}$.
\end{enumerate}
Therefore, 
\begin{equation}\label{eq:14}
    \mathbb{E}\big[ \boldsymbol{1}\{{A}_{tk} = 0\}\boldsymbol{1}\{{A}_{tl} = 0\} \big| \boldsymbol{W}_t,\boldsymbol{X}_t,\mathcal{H}_{t-1}\big] = \mathbb{E}\big[ \boldsymbol{1}\{{A}_{tk} = 0\}\big| \boldsymbol{W}_t,\boldsymbol{X}_t,\mathcal{H}_{t-1}\big]\cdot\mathbb{E} \big[\boldsymbol{1}\{{A}_{tl} = 0\} \big| \boldsymbol{W}_t,\boldsymbol{X}_t,\mathcal{H}_{t-1}\big].
\end{equation}
For the simplicity of notation, we define $\nu_{ti}(\omega_{ti},\boldsymbol{X}_{ti},\mathcal{H}_{t-1}) = \mathbb{P}({A}_{ti} \neq  \pi^*(\boldsymbol{X}_{ti})|\boldsymbol{W}_t,\boldsymbol{X}_{ti},\mathcal{H}_{t-1}) = \mathbb{P}({A}_{ti} \neq  \pi^*(\boldsymbol{X}_{ti})|\omega_{ti},\boldsymbol{X}_{ti},\mathcal{H}_{t-1})$. Since 
$$
\boldsymbol{1}\{{A}_{ti} = a\} = \boldsymbol{1}\{{A}_{ti} = \pi^*(\boldsymbol{X}_{ti})\}\cdot \boldsymbol{1}\{\pi^*(\boldsymbol{X}_{ti}) = a\} + \boldsymbol{1}\{{A}_{ti} \neq  \pi^*(\boldsymbol{X}_{ti})\}\cdot \boldsymbol{1}\{\pi^*(\boldsymbol{X}_{ti}) \neq a\},
$$
we have
\begin{equation}\label{eq:4}
\begin{aligned}
    &\mathbb{E}\big[ \boldsymbol{1}\{{A}_{ti} = a\} \big| \boldsymbol{W}_t,\boldsymbol{X}_t,\mathcal{H}_{t-1}\big] \\
    &=\mathbb{E}\big[ \boldsymbol{1}\{{A}_{ti} = \pi^*(\boldsymbol{X}_{ti})\}\cdot \boldsymbol{1}\{\pi^*(\boldsymbol{X}_{ti}) = a\} + \boldsymbol{1}\{{A}_{ti} \neq  \pi^*(\boldsymbol{X}_{ti})\}\cdot \boldsymbol{1}\{\pi^*(\boldsymbol{X}_{ti}) \neq a\}\big| \boldsymbol{W}_t,\boldsymbol{X}_t,\mathcal{H}_{t-1}\big]\\
    & = \mathbb{P}({A}_{ti} = \pi^*(\boldsymbol{X}_{ti})|\boldsymbol{W}_t,\boldsymbol{X}_t,\mathcal{H}_{t-1})\boldsymbol{1}\{\pi^*(\boldsymbol{X}_{ti}) = a\} + \mathbb{P}({A}_{ti} \neq \pi^*(\boldsymbol{X}_{ti})|\boldsymbol{W}_t,\boldsymbol{X}_t,\mathcal{H}_{t-1})\boldsymbol{1}\{\pi^*(\boldsymbol{X}_{ti}) \neq a\}\\
    &= (1-\nu_{ti}(\omega_{ti},\boldsymbol{X}_{ti},\mathcal{H}_{t-1}))\boldsymbol{1}\{\omega_{ti}\boldsymbol{X}_{ti}\boldsymbol{\beta}_a\geq\omega_{ti}\boldsymbol{X}_{ti}\boldsymbol{\beta}_{1-a}\} + \nu_{ti}(\omega_{ti},\boldsymbol{X}_{ti},\mathcal{H}_{t-1})\boldsymbol{1}\{\omega_{ti}\boldsymbol{X}_{ti}\boldsymbol{\beta}_a<\omega_{ti}\boldsymbol{X}_{ti}\boldsymbol{\beta}_{1-a}\}.
\end{aligned}
\end{equation}
Plugging in the result of Equation \eqref{eq:14}, \eqref{eq:4} to Equation \eqref{eq:5}, one can obtain
\begin{equation*}
\begin{aligned}
    \mathbb{E}[M_1]&=\mathbb{E}_{\boldsymbol{W}_t,\boldsymbol{X}_t}\Bigg[\sum_{k=1}^{N_t}\sum_{l=1}^{N_t} W_{t,ik}W_{t,il}\boldsymbol{X}_{tk}\boldsymbol{X}_{tl}'\cdot\mathbb{E}\big[ \boldsymbol{1}\{{A}_{tk} = 0\}\boldsymbol{1}\{{A}_{tl} = 0\} \big|\boldsymbol{W}_t, \boldsymbol{X}_t,\mathcal{H}_{t-1}\big]\Bigg]\\
    & = \mathbb{E}_{\boldsymbol{W}_t,\boldsymbol{X}_t}\Bigg[\sum_{1\leq k,l\leq N_t}^{k\neq l} W_{t,ik}W_{t,il}\boldsymbol{X}_{tk}\boldsymbol{X}_{tl}'\\
    &\quad\quad\cdot\bigg\{(1-\nu_{tk}(\omega_{tk},\boldsymbol{X}_{tk},\mathcal{H}_{t-1}))\boldsymbol{1}\{\omega_{tk}\boldsymbol{X}_{tk}'(\boldsymbol{\beta}_0-\boldsymbol{\beta}_{1})\geq 0\} + \nu_{tk}(\omega_{tk},\boldsymbol{X}_{tk},\mathcal{H}_{t-1})\boldsymbol{1}\{\omega_{tk}\boldsymbol{X}_{tk}'(\boldsymbol{\beta}_0-\boldsymbol{\beta}_{1})<0\}\bigg\}\\
    &\quad\quad \cdot\bigg\{(1-\nu_{tl}(\omega_{tl},\boldsymbol{X}_{tl},\mathcal{H}_{t-1}))\boldsymbol{1}\{\omega_{tl}\boldsymbol{X}_{tl}'(\boldsymbol{\beta}_0-\boldsymbol{\beta}_{1})\geq 0\} + \nu_{tl}(\omega_{tl},\boldsymbol{X}_{tl},\mathcal{H}_{t-1})\boldsymbol{1}\{\omega_{tl}\boldsymbol{X}_{tl}'(\boldsymbol{\beta}_0-\boldsymbol{\beta}_{1})<0\}\bigg\}\\
    &\quad+ \sum_{1\leq k,l\leq N_t}^{k= l}W_{t,ik}^2\boldsymbol{X}_{tk}\boldsymbol{X}_{tk}' \\
    &\quad\quad \cdot\bigg\{(1-\nu_{tk}(\omega_{tk},\boldsymbol{X}_{tk},\mathcal{H}_{t-1}))\boldsymbol{1}\{\omega_{tk}\boldsymbol{X}_{tk}'(\boldsymbol{\beta}_0-\boldsymbol{\beta}_{1})\geq 0\} + \nu_{tk}(\omega_{tk},\boldsymbol{X}_{tk},\mathcal{H}_{t-1})\boldsymbol{1}\{\omega_{tk}\boldsymbol{X}_{tk}'(\boldsymbol{\beta}_0-\boldsymbol{\beta}_{1})<0\}\bigg\}\Bigg].
\end{aligned}
\end{equation*}

Define $\kappa_\infty (\omega,\boldsymbol{x}) = \lim_{q\rightarrow\infty}\mathbb{P}(A_{ti}\neq \pi^*(\boldsymbol{x}))$. Following similar procedure as page 19-20 and Lemma B.1 in \citet{ye2023doubly}, we can also derive that $\nu_{tk}(\omega,\boldsymbol{x},\mathcal{H}_{t-1})\xrightarrow{p}\kappa_\infty (\omega,\boldsymbol{x})$, where the limit is free of historical data $\mathcal{H}_{t-1}$.
Therefore, 
\begin{equation}\label{eq:M1}
\begin{aligned}
    \frac{1}{\bar{N}_q}\sum_{q=1}^{\bar{N}_q}\sigma^2\mathbb{E}[M_1]
    & \rightarrow \frac{1}{\bar{N}_q}\sum_{q=1}^{\bar{N}_q}\sigma^2\mathbb{E}_{\boldsymbol{W}_t,\boldsymbol{X}_t}\Bigg[\sum_{1\leq k,l\leq N_t}^{k\neq l} W_{t,ik}W_{t,il}\boldsymbol{X}_{tk}\boldsymbol{X}_{tl}'\mathcal{J}_{\infty}(\omega_{tk},\boldsymbol{X}_{tk},\boldsymbol{\beta})\mathcal{J}_{\infty}(\omega_{tl},\boldsymbol{X}_{tl},\boldsymbol{\beta})\\
    &\qquad\qquad\qquad\qquad\quad+ \sum_{1\leq k,l\leq N_t}^{k= l}W_{t,ik}^2\boldsymbol{X}_{tk}\boldsymbol{X}_{tk}' \mathcal{J}_{\infty}(\omega_{tk},\boldsymbol{X}_{tk},\boldsymbol{\beta})\Bigg],
\end{aligned}
\end{equation}
where
$
    \mathcal{J}_{\infty}(\omega,\boldsymbol{X},\boldsymbol{\beta})=  \Big\{(1-\kappa_{\infty}(\omega,\boldsymbol{X}))\boldsymbol{1}\{\omega\boldsymbol{X}'(\boldsymbol{\beta}_0-\boldsymbol{\beta}_{1})\geq 0\} + \kappa_{\infty}(\omega,\boldsymbol{X})\boldsymbol{1}\{\omega\boldsymbol{X}'(\boldsymbol{\beta}_0-\boldsymbol{\beta}_{1}\})<0\Big\}.
$

Similarly, we can derive the asymptotic variance for the rest three parts of $\frac{1}{\bar{N}_q} \sum_{q=1}^{\bar{N}_q}\sigma^2\mathbb{E}\big[\widetilde{\boldsymbol{X}}_{q}\widetilde{\boldsymbol{X}}_{q}' |\mathcal{H}_{t-1}\big]$ and obtain the final asymptotic variance. Specifically,
\begin{equation}\label{eq:8}
\begin{aligned}
    &\frac{1}{\bar{N}_q}\sum_{q=1}^{\bar{N}_q}\sigma^2\mathbb{E}[M_2]  \rightarrow   \frac{1}{\bar{N}_q}\sum_{q=1}^{\bar{N}_q}\sigma^2\mathbb{E}_{\boldsymbol{W}_t,\boldsymbol{X}_t}\bigg[\sum_{1\leq k,l\leq N_t}^{k\neq l} W_{t,ik}W_{t,il}\boldsymbol{X}_{tk}\boldsymbol{X}_{tl}'\mathcal{J}_{\infty}(\omega_{tk},\boldsymbol{X}_{tk},\boldsymbol{\beta})\widetilde{\mathcal{J}}_{\infty}(\omega_{tl},\boldsymbol{X}_{tl},\boldsymbol{\beta})\bigg],\\
    &\frac{1}{\bar{N}_q}\sum_{q=1}^{\bar{N}_q}\sigma^2\mathbb{E}[M_3]   \rightarrow   \frac{1}{\bar{N}_q}\sum_{q=1}^{\bar{N}_q}\sigma^2\mathbb{E}_{\boldsymbol{W}_t,\boldsymbol{X}_t}\bigg[\sum_{1\leq k,l\leq N_t}^{k\neq l} W_{t,ik}W_{t,il}\boldsymbol{X}_{tk}\boldsymbol{X}_{tl}'\widetilde{\mathcal{J}}_{\infty}(\omega_{tk},\boldsymbol{X}_{tk},\boldsymbol{\beta})\mathcal{J}_{\infty}(\omega_{tl},\boldsymbol{X}_{tl},\boldsymbol{\beta})\bigg],\\
    &\frac{1}{\bar{N}_q}\sum_{q=1}^{\bar{N}_q}\sigma^2\mathbb{E}[M_4]  \rightarrow  \frac{1}{\bar{N}_q}\sum_{q=1}^{\bar{N}_q}\sigma^2\mathbb{E}_{\boldsymbol{W}_t,\boldsymbol{X}_t}\bigg[\sum_{1\leq k,l\leq N_t}^{k\neq l} W_{t,ik}W_{t,il}\boldsymbol{X}_{tk}\boldsymbol{X}_{tl}'\widetilde{\mathcal{J}}_{\infty}(\omega_{tk},\boldsymbol{X}_{tk},\boldsymbol{\beta})\widetilde{\mathcal{J}}_{\infty}(\omega_{tl},\boldsymbol{X}_{tl},\boldsymbol{\beta})\\
    &\qquad\qquad\qquad\qquad\qquad\qquad\qquad\qquad+ \sum_{1\leq k,l\leq N_t}^{k= l}W_{t,ik}^2\boldsymbol{X}_{tk}\boldsymbol{X}_{tk}' \widetilde{\mathcal{J}}_{\infty}(\omega_{tk},\boldsymbol{X}_{tk},\boldsymbol{\beta})\bigg],
\end{aligned}
\end{equation}
where 
$
    \widetilde{\mathcal{J}}_{\infty}(\omega,\boldsymbol{X},\boldsymbol{\beta})=  \Big\{(1-\kappa_{\infty}(\omega,\boldsymbol{X}))\boldsymbol{1}\{\omega\boldsymbol{X}'(\boldsymbol{\beta}_1-\boldsymbol{\beta}_{0})\geq 0\} + \kappa_{\infty}(\omega,\boldsymbol{X})\boldsymbol{1}\{\omega\boldsymbol{X}'(\boldsymbol{\beta}_1-\boldsymbol{\beta}_{0}\})<0\Big\}.
$

Define $\boldsymbol{v} = (\boldsymbol{v}_1',\boldsymbol{v}_2')'$ where $\boldsymbol{v}_1$ and $\boldsymbol{v}_2$ are both $d$-dimensional vector. 

Then $\eta_1(\boldsymbol{v})=\frac{1}{\sqrt{\bar{N}_q}}\sum_{q=1}^{\bar{N}_q} \boldsymbol{v}'\widetilde{\boldsymbol{X}}_{q}\epsilon_q \xrightarrow{\mathcal{D}}\mathcal{N}(\boldsymbol{0}_{2d}, \boldsymbol{v}'G\boldsymbol{v})$ with

\begin{equation}\label{eq:9}
\begin{aligned}
    &G = \begin{bmatrix}\frac{1}{\bar{N}_q}\sum_{q=1}^{\bar{N}_q}\sigma^2\mathbb{E}[M_1]& \frac{1}{\bar{N}_q}\sum_{q=1}^{\bar{N}_q}\sigma^2\mathbb{E}[M_2]\\ \frac{1}{\bar{N}_q}\sum_{q=1}^{\bar{N}_q}\sigma^2\mathbb{E}[M_3] & \frac{1}{\bar{N}_q}\sum_{q=1}^{\bar{N}_q}\sigma^2\mathbb{E}[M_4]\\
    \end{bmatrix},
\end{aligned}
\end{equation}
where the detailed expression of each submatrix in $G$ is given in Equation \eqref{eq:M1}-\eqref{eq:8}.

\noindent\textbf{Step 2:} Show that $\eta_2= \left\{\frac{1}{\bar{N}_t}\sum_{s=1}^{t}\sum_{i=1}^{N_s}\widetilde{\boldsymbol{X}}_{si}\widetilde{\boldsymbol{X}}_{si}' \right\}^{-1}\xrightarrow{p}\sigma^2 G^{-1}$, where $\sigma^2 = \mathbb{E}[\epsilon_{ti}^2|\boldsymbol{A}_t]$.

Based on Lemma 6 of \citet{chen2021statisticala}, it suffice to find the limit of $\frac{1}{\bar{N}_q}\sum_{q=1}^{\bar{N}_q}\boldsymbol{v}'\widetilde{\boldsymbol{X}}_{q}\widetilde{\boldsymbol{X}}_{q}'\boldsymbol{v}$. According to Equation \eqref{eq:7}, we have 
\begin{equation*}
    \mathbb{P}(|\boldsymbol{v}'\widetilde{\boldsymbol{X}}_{q}\widetilde{\boldsymbol{X}}_{q}'\boldsymbol{v}|>h)\leq \mathbb{P}(2dK^2L_x^2\|\boldsymbol{v}\|_2^2>h).
\end{equation*}
Therefore, by Theorem 2.19 in \citet{hall2014martingale}, we have 
\begin{equation*}
    \frac{1}{\bar{N}_q}\sum_{q=1}^{\bar{N}_q}\left[\boldsymbol{v}'\widetilde{\boldsymbol{X}}_{q}\widetilde{\boldsymbol{X}}_{q}'\boldsymbol{v} - \mathbb{E}\big\{(\boldsymbol{v}'\widetilde{\boldsymbol{X}}_{q})^2 |\mathcal{H}_{q-1}\big\}\right]\xrightarrow{p} 0 \quad \text{as }q\rightarrow \infty.
\end{equation*}
Based on the results in Step 1, one can easily derive $\mathbb{E}\big\{(\boldsymbol{v}'\widetilde{\boldsymbol{X}}_{q})^2 |\mathcal{H}_{q-1}\big\} = \boldsymbol{v}'G\boldsymbol{v}/\sigma^2$. Combining the results above and by Continuous Mapping Theorem, we have
\begin{equation}
    \eta_2  =\left\{\frac{1}{\bar{N}_t}\sum_{s=1}^{t}\sum_{i=1}^{N_s}\widetilde{\boldsymbol{X}}_{si}\widetilde{\boldsymbol{X}}_{si}' \right\}^{-1}\xrightarrow{p}(G/\sigma^2)^{-1} = \sigma^2 G^{-1},
\end{equation}
which finishes the proof of this step.

\noindent\textbf{Step 3:} Summary.\\
According to the results of Step 1-2 and Slutsky's Theorem, we can conclude that
\begin{equation}
    \sqrt{\bar{N}_t}(\widehat{\boldsymbol{\beta}}_t -\boldsymbol{\beta}) = \eta_2\eta_1\xrightarrow{\mathcal{D}} \mathcal{N}(\boldsymbol{0}_{2d},\sigma^4G^{-1}), 
\end{equation}
where $G$ is specified in Equation \eqref{eq:9}.

In the special case when $N_t = 1$ for all $t$, i.e. there is no interference, the asymptotic variance would degenerate to
\begin{equation}
    G = \frac{1}{\bar{N}_q}\sum_{q=1}^{\bar{N}_q}\sigma^2\cdot \begin{bmatrix} \mathcal{K}_{\infty}(\boldsymbol{\beta})  & 0\\ 0  & \widetilde{\mathcal{K}}_{\infty}(\boldsymbol{\beta})\\\end{bmatrix}.
\end{equation}
with
\begin{equation*}
\begin{aligned}
    \mathcal{K}_{\infty}(\boldsymbol{\beta})&=\int_{\boldsymbol{x}}  \boldsymbol{x}\boldsymbol{x}'\cdot \Big\{(1-\kappa_{\infty}(\boldsymbol{x}))\boldsymbol{1}\{\boldsymbol{x}'(\boldsymbol{\beta}_0-\boldsymbol{\beta}_{1})\geq 0\} + \kappa_{\infty}(\boldsymbol{x})\boldsymbol{1}\{\boldsymbol{x}'(\boldsymbol{\beta}_0-\boldsymbol{\beta}_{1}\})<0\Big\}d\mathcal{P}_{\boldsymbol{x}}\\
    \widetilde{\mathcal{K}}_{\infty}(\boldsymbol{\beta})&=\int_{\boldsymbol{x}}  \boldsymbol{x}\boldsymbol{x}'\cdot \Big\{(1-\kappa_{\infty}(\boldsymbol{x}))\boldsymbol{1}\{\boldsymbol{x}'(\boldsymbol{\beta}_1-\boldsymbol{\beta}_{0})\geq 0\} + \kappa_{\infty}(\boldsymbol{x})\boldsymbol{1}\{\boldsymbol{x}'(\boldsymbol{\beta}_1-\boldsymbol{\beta}_{0}\})<0\Big\}d\mathcal{P}_{\boldsymbol{x}}.
\end{aligned}
\end{equation*}
which align perfectly with \citet{ye2023doubly} in the cases without interference.

The proof of this theorem is complete.

\section{Proof of Theorem \ref{thm:2}: the Asymptotic Normality of \texorpdfstring{$V^{\pi^*}$}{V}}
   
Recall that the DR optimal value function estimator we derived is given by
\begin{equation}
    \widehat{V}^{\text{DR}}_T= \frac{1}{\bar{N}_T}\sum_{t=1}^T \sum_{i=1}^{N_t} \left\{\frac{\boldsymbol{1}\{a_{ti} = \widehat{\pi}_{t-1}(\boldsymbol{X}_{ti})\}}{1-\hat{\kappa}_{t-1}(\boldsymbol{X}_{ti})} \cdot \Big(r_{si} -\widehat{\mu}_{t-1}^{(t,i)}(\boldsymbol{X}_t,\widehat{\pi}_{t-1}(\boldsymbol{X}_{t})) \Big) + \widehat{\mu}_{t-1}^{(t,i)}(\boldsymbol{X}_t,\widehat{\pi}_{t-1}(\boldsymbol{X}_{t}))\right\}.
\end{equation}

For the brevity of notation, we will omit the superscript in $\widehat{V}^{\text{DR}}_t$ in the following proof.

Now we defined two related value functions $\widetilde{V}_T$ and $\bar{V}_T$ as below:
\begin{equation*}
\begin{aligned}
  \widetilde{V}_T &= \frac{1}{\bar{N}_T}\sum_{t=1}^T \sum_{i=1}^{N_t} \left\{\frac{\boldsymbol{1}\{a_{ti} = \widehat{\pi}_{t-1}(\boldsymbol{X}_{ti})\}}{1-{\kappa}_{t-1}(\boldsymbol{X}_{ti})} \cdot \Big(r_{ti} -{\mu}^{(t,i)}(\boldsymbol{X}_t,\widehat{\pi}_{t-1}(\boldsymbol{X}_{t})) \Big) + {\mu}^{(t,i)}(\boldsymbol{X}_t,\widehat{\pi}_{t-1}(\boldsymbol{X}_{t}))\right\},\\  
  \bar{V}_T &= \frac{1}{\bar{N}_T}\sum_{t=1}^T \sum_{i=1}^{N_t} \left\{\frac{\boldsymbol{1}\{a_{ti} = {\pi}^*(\boldsymbol{X}_{ti})\}}{\mathbb{P}(a_{ti} = {\pi}^*(\boldsymbol{X}_{ti}))} \cdot \Big(r_{ti} -{\mu}^{(t,i)}(\boldsymbol{X}_t,{\pi}^*(\boldsymbol{X}_t)) \Big) + {\mu}^{(t,i)}(\boldsymbol{X}_t,{\pi}^*(\boldsymbol{X}_t))\right\}.
  \end{aligned}
\end{equation*}

The proof of this theorem can be decomposed into three steps. In step 1, we aim to prove $
    \widehat{V}_T = \widetilde{V}_T + o_p(\bar{N}_T^{-1/2})$.
In step 2, we show that $
    \widehat{V}_T = \widetilde{V}_T + o_p(\bar{N}_T^{-1/2})$.
In step 3, we show $
    \sqrt{\bar{N}_T}(\bar{V}_T -V^{\pi^*})\xrightarrow{\mathcal{D}} \mathcal{N}(0,\sigma^2_V)$, 
where the variance term is given by 
\begin{equation*}
    \sigma^2_V = \int \frac{\sigma^2}{1-\kappa_\infty (\boldsymbol{x})}d \mathcal{P}_{\boldsymbol{x}} +  \frac{\sum_{i,t}\omega_{ti}^2}{\bar{N}_T}\cdot \text{Var}\big\{{\pi}^*(\boldsymbol{x})\cdot \boldsymbol{x}'\boldsymbol{\beta}_1 +\{1-{\pi}^*(\boldsymbol{x})\}\cdot \boldsymbol{x}'\boldsymbol{\beta}_0\big\}.
\end{equation*}
Combining the above three steps, the proof of theorem \ref{thm:2} is thus complete.

Now, let's detail the proof of step 1-3.

\noindent\textbf{Step 1:} Prove that $\widehat{V}_T = \widetilde{V}_T + o_p(\bar{N}_T^{-1/2})$.\\

Notice that the different between $\widehat{V}_T$ and $\widetilde{V}_T$ lies in the estimation accuracy of (1) the propensity score function $\hat{\kappa}_{t-1}$ and (2) outcome estimation function $\widehat{\mu}_{t-1}^{(t,i)}$. To simplify this problem, we introduce another intermediate value function $\breve{V}_T$ as
\begin{equation*}
    \breve{V}_T = \frac{1}{\bar{N}_T}\sum_{t=1}^T \sum_{i=1}^{N_t} \left\{\frac{\boldsymbol{1}\{a_{ti} = \widehat{\pi}_{t-1}(\boldsymbol{X}_{ti})\}}{1-{\kappa}_{t-1}(\boldsymbol{X}_{ti})} \cdot \Big(r_{ti} -\widehat{\mu}_{t-1}^{(t,i)}(\boldsymbol{X}_t,\widehat{\pi}_{t-1}(\boldsymbol{X}_{ti})) \Big) + \widehat{\mu}_{t-1}^{(t,i)}(\boldsymbol{X}_t,\widehat{\pi}_{t-1}(\boldsymbol{X}_{ti}))\right\}.
\end{equation*}
Now the problem becomes proving (1) $\widehat{V}_T = \breve{V}_T + o_p(\bar{N}_T^{-1/2})$, and (2) $\breve{V}_T = \widetilde{V}_T + o_p(\bar{N}_T^{-1/2})$.

First, let's prove (1) $\widehat{V}_T = \breve{V}_T + o_p(\bar{N}_T^{-1/2})$. Notice that
\begin{equation*}
\begin{aligned}
     \widehat{V}_T - \breve{V}_T & = \frac{1}{\bar{N}_T}\sum_{t=1}^T \sum_{i=1}^{N_t} \left[\frac{\boldsymbol{1}\{a_{ti} = \widehat{\pi}_{t-1}(\boldsymbol{X}_{ti})\}}{1-\hat{\kappa}_{t-1}(\boldsymbol{X}_{ti})} - \frac{\boldsymbol{1}\{a_{ti} = \widehat{\pi}_{t-1}(\boldsymbol{X}_{ti})\}}{1-{\kappa}_{t-1}(\boldsymbol{X}_{ti})} \right]\cdot \Big\{r_{ti} -\widehat{\mu}_{t-1}^{(t,i)}(\boldsymbol{X}_t,\widehat{\pi}_{t-1}(\boldsymbol{X}_{t})) \Big\}\\
    & = \frac{1}{\bar{N}_T}\sum_{t=1}^T \sum_{i=1}^{N_t} \big\{\hat{\kappa}_{t-1}(\boldsymbol{X}_{ti}) - {\kappa}_{t-1}(\boldsymbol{X}_{ti})\big\} \cdot \left[\frac{\boldsymbol{1}\{a_{ti} = \widehat{\pi}_{t-1}(\boldsymbol{X}_{ti})\}\Big\{r_{ti} -\widehat{\mu}_{t-1}^{(t,i)}(\boldsymbol{X}_t,\widehat{\pi}_{t-1}(\boldsymbol{X}_{t})) \Big\}}{\{1-\hat{\kappa}_{t-1}(\boldsymbol{X}_{ti})\}\{1-{\kappa}_{t-1}(\boldsymbol{X}_{ti})\}}  \right]\\
    & = \underbrace{\frac{1}{\bar{N}_T}\sum_{t=1}^T \sum_{i=1}^{N_t} \big\{\hat{\kappa}_{t-1}(\boldsymbol{X}_{ti}) - {\kappa}_{t-1}(\boldsymbol{X}_{ti})\big\} \cdot \left[\frac{\boldsymbol{1}\{a_{ti} = \widehat{\pi}_{t-1}(\boldsymbol{X}_{ti})\}\Big\{r_{ti} -{\mu}^{(t,i)}(\boldsymbol{X}_t,\widehat{\pi}_{t-1}(\boldsymbol{X}_{t})) \Big\}}{\{1-\hat{\kappa}_{t-1}(\boldsymbol{X}_{ti})\}\{1-{\kappa}_{t-1}(\boldsymbol{X}_{ti})\}}  \right]}_{\Delta_1}\\
    & + \underbrace{\frac{1}{\bar{N}_T}\sum_{t=1}^T \sum_{i=1}^{N_t} \big\{\hat{\kappa}_{t-1}(\boldsymbol{X}_{ti}) - {\kappa}_{t-1}(\boldsymbol{X}_{ti})\big\} \cdot \left[\frac{\boldsymbol{1}\{a_{ti} = \widehat{\pi}_{t-1}(\boldsymbol{X}_{ti})\}\Big\{{\mu}^{(t,i)}(\boldsymbol{X}_t,\widehat{\pi}_{t-1}(\boldsymbol{X}_{t})) -\widehat{\mu}_{t-1}^{(t,i)}(\boldsymbol{X}_t,\widehat{\pi}_{t-1}(\boldsymbol{X}_{t})) \Big\}}{\{1-\hat{\kappa}_{t-1}(\boldsymbol{X}_{ti})\}\{1-{\kappa}_{t-1}(\boldsymbol{X}_{ti})\}}  \right]}_{\Delta_2}.
\end{aligned}
\end{equation*}
We first show that $\Delta_1$ is $o_p(T^{-1/2})$.
Similar to the proof of Theorem 3 in \citet{ye2023doubly}, we define a class of measurable functions 
\begin{equation*}
    \mathcal{F}(\boldsymbol{X}_t,a_{ti},r_{ti}) = \Bigg\{\big\{\hat{\kappa}_{t-1}(\boldsymbol{X}_{ti}) - {\kappa}_{t-1}(\boldsymbol{X}_{ti})\big\} \cdot \bigg[\frac{\boldsymbol{1}\{a_{ti} = \widehat{\pi}_{t-1}(\boldsymbol{X}_{ti})\}\Big\{r_{ti} -{\mu}^{(t,i)}(\boldsymbol{X}_t,\widehat{\pi}_{t-1}(\boldsymbol{X}_{t})) \Big\}}{\{1-\hat{\kappa}_{t-1}(\boldsymbol{X}_{ti})\}\{1-{\kappa}_{t-1}(\boldsymbol{X}_{ti})\}}\bigg]: \hat{\kappa}_{t}, {\kappa}_t\in \Lambda,\widehat{\pi}_{t}\in \Pi\Bigg\},
\end{equation*}
where $\Lambda$ and $\Pi$ are two classes of functions mapping context $\boldsymbol{X}_{ti}$ to a probability in $[0,1]$. Denote the empirical measure $\mathbb{G}_n = \sqrt{n}\mathbb{P}_n(f-\mathbb{P}f)$. Here, $n = Q(t,i)$ is the sample index, which is determined by reordering the units $i\in\{1,\dots,N_t\}$ according to time $t$. Denote $\|z\|_{\mathcal{F}} = \sup_{f\in\mathcal{F}}|z(f)|$. Therefore,
\begin{equation}\label{eq:Gn_f}
    \|\mathbb{G}_n  \|_{\mathcal{F}} := \sup_{\pi\in\Pi} \bigg|\frac{1}{\sqrt{\bar{N}_T}}\sum_{q\in Q(t,i)} \left[\mathcal{F}(\boldsymbol{X}_t,a_{ti},r_{ti}) -\mathbb{E}\big\{\mathcal{F}(\boldsymbol{X}_t,a_{ti},r_{ti}) \mid \mathcal{H}_{t-1}\big\}\right] \bigg|.
\end{equation}
Since $\mu^{(t,i)}$ in $\mathcal{F}$ is correctly specified, we always have
\begin{equation*}
\begin{aligned}
    &\mathbb{E}\big\{\mathcal{F}(\boldsymbol{X}_t,a_{ti},r_{ti}) \mid \mathcal{H}_{t-1}\big\} \\
    &= \mathbb{E} \Bigg[\big\{\hat{\kappa}_{t-1}(\boldsymbol{X}_{ti}) - {\kappa}_{t-1}(\boldsymbol{X}_{ti})\big\} \cdot \Bigg\{\frac{\boldsymbol{1}\{a_{ti} = \widehat{\pi}_{t-1}(\boldsymbol{X}_{ti})\}\big\{r_{ti} -{\mu}^{(t,i)}(\boldsymbol{X}_t,\widehat{\pi}_{t-1}(\boldsymbol{X}_{t})) \big\}}{\{1-\hat{\kappa}_{t-1}(\boldsymbol{X}_{ti})\}\{1-{\kappa}_{t-1}(\boldsymbol{X}_{ti})\}}\Bigg\}\text{ }\Big|\text{ }\mathcal{H}_{t-1}\Bigg]\\
    & =\mathbb{E} \Bigg[\big\{\hat{\kappa}_{t-1}(\boldsymbol{X}_{ti}) - {\kappa}_{t-1}(\boldsymbol{X}_{ti})\big\} \cdot \Bigg\{\frac{\boldsymbol{1}\{a_{ti} = \widehat{\pi}_{t-1}(\boldsymbol{X}_{ti})\}\cdot e_{ti}}{\{1-\hat{\kappa}_{t-1}(\boldsymbol{X}_{ti})\}\{1-{\kappa}_{t-1}(\boldsymbol{X}_{ti})\}}\Bigg\}\text{ }\Big|\text{ }\mathcal{H}_{t-1}\Bigg]
\end{aligned}.
\end{equation*}
According to the iteration of expectation, the equality above can be further derived as
\begin{equation*}
\begin{aligned}
    &\mathbb{E}\big\{\mathcal{F}(\boldsymbol{X}_t,a_{ti},r_{ti}) \mid \mathcal{H}_{t-1}\big\} \\
    &=\mathbb{E} \Bigg[\big\{\hat{\kappa}_{t-1}(\boldsymbol{X}_{ti}) - {\kappa}_{t-1}(\boldsymbol{X}_{ti})\big\} \cdot \Bigg\{\frac{\mathbb{E}[\boldsymbol{1}\{a_{ti} = \widehat{\pi}_{t-1}(\boldsymbol{X}_{ti})\}|\boldsymbol{X}_{ti}]}{\{1-\hat{\kappa}_{t-1}(\boldsymbol{X}_{ti})\}\{1-{\kappa}_{t-1}(\boldsymbol{X}_{ti})\}}\cdot \mathbb{E}[e_{ti}|\boldsymbol{X}_{ti},\boldsymbol{A}_t]\Bigg\}\text{ }\Big|\text{ }\mathcal{H}_{t-1}\Bigg]=0,
\end{aligned}.
\end{equation*}
where the last equality holds  by $\mathbb{E}[\epsilon_{ti}|\boldsymbol{X}_{ti},\boldsymbol{A}_t] = 0$ according to the definition of noise $\epsilon_{ti}$.

Therefore, Equation \eqref{eq:Gn_f} can be simplified as
\begin{equation*}
    \|\mathbb{G}_n  \|_{\mathcal{F}} = \sup_{\pi\in\Pi} \bigg|\frac{1}{\sqrt{\bar{N}_T}}\sum_{q\in Q(t,i)}\mathcal{F}(\boldsymbol{X}_t,a_{ti},r_{ti})  \bigg|.
\end{equation*}
Following Section 4.2 of \citet{dedecker2002maximal}, we define 
\begin{equation*}
\begin{aligned}
    d_1(f) := \left\|\mathbb{E}\big\{|f(\boldsymbol{X}_{1},a_{11},r_{11})|\big|\mathcal{H}_0\big\}\right\|_{\infty},  \quad 
    d_2(f) := \left\|\mathbb{E}\big\{(f(\boldsymbol{X}_{1},a_{11},r_{11}))^2\big|\mathcal{H}_0\big\}\right\|_{\infty}^{1/2}.
\end{aligned}
\end{equation*}
First, we show that both $d_1(f)$ and $d_2(f)$ are finite numbers. For the brevity of content, we will take $d_2(f)$ as an example and $d_1(f)<\infty$ can be proved similarly.

In a valid bandits algorithm, the probability of exploration ${\kappa}_t(\boldsymbol{X}_{ti})$ is bounded away from 1. That is, there exists a constant $C_1<1$, such that ${\kappa}_t(\boldsymbol{X}_{ti}) = \mathbb{P}(a_{ti}\neq \widehat{\pi}_{t-1}(\boldsymbol{X}_{ti}))\leq C_1<1$, and $\hat{\kappa}_t(\boldsymbol{X}_{ti})\leq C_1<1$. Therefore, for any $t\in\{1,\dots,T\}$, 
\begin{equation*}
\begin{aligned}
    & \mathbb{E}\big\{(f(\boldsymbol{X}_{t},a_{ti},r_{ti}))^2\big|\mathcal{H}_{t-1}\big\} \\
    &= \mathbb{E} \Bigg[\Bigg\{\big\{\hat{\kappa}_{t-1}(\boldsymbol{X}_{ti}) - {\kappa}_{t-1}(\boldsymbol{X}_{ti})\big\} \cdot \frac{\boldsymbol{1}\{a_{ti} = \widehat{\pi}_{t-1}(\boldsymbol{X}_{ti})\}\big\{r_{ti} -{\mu}^{(t,i)}(\boldsymbol{X}_t,\widehat{\pi}_{t-1}(\boldsymbol{X}_{t})) \big\}}{\{1-\hat{\kappa}_{t-1}(\boldsymbol{X}_{ti})\}\{1-{\kappa}_{t-1}(\boldsymbol{X}_{ti})\}}\Bigg\}^2\text{ }\Big|\text{ }\mathcal{H}_{t-1}\Bigg]\\
    & =\mathbb{E} \Bigg[\Bigg\{\frac{\hat{\kappa}_{t-1}(\boldsymbol{X}_{ti}) - {\kappa}_{t-1}(\boldsymbol{X}_{ti})}{\{1-\hat{\kappa}_{t-1}(\boldsymbol{X}_{ti})\}\{1-{\kappa}_{t-1}(\boldsymbol{X}_{ti})\}} \Bigg\}^2\cdot \boldsymbol{1}\{a_{ti} = \widehat{\pi}_{t-1}(\boldsymbol{X}_{ti})\}\cdot \big\{r_{ti} -{\mu}^{(t,i)}(\boldsymbol{X}_t,\widehat{\pi}_{t-1}(\boldsymbol{X}_{t})) \big\}^2\text{ }\Big|\text{ }\mathcal{H}_{t-1}\Bigg]\\
    & =\mathbb{E} \Bigg[\Bigg\{\frac{\hat{\kappa}_{t-1}(\boldsymbol{X}_{ti}) - {\kappa}_{t-1}(\boldsymbol{X}_{ti})}{\{1-\hat{\kappa}_{t-1}(\boldsymbol{X}_{ti})\}\{1-{\kappa}_{t-1}(\boldsymbol{X}_{ti})\}} \Bigg\}^2\cdot \boldsymbol{1}\{a_{ti} = \widehat{\pi}_{t-1}(\boldsymbol{X}_{ti})\}\cdot \mathbb{E}[\epsilon_{ti}^2|\boldsymbol{X_t},\boldsymbol{A}_t]\text{ }\Big|\text{ }\mathcal{H}_{t-1}\Bigg]\\
    &\leq \bigg(\frac{2}{1-C_1}\bigg)^2 \cdot 1\cdot \sigma^2 <\infty.
\end{aligned}
\end{equation*}
Therefore, by Rosenthal’s inequality derived for Martingale [see \citet{dedecker2002maximal} for details], we have
\begin{equation}\label{eq:EGn_f}
    \mathbb{E}\left[\|\mathbb{G}_n\|_{\mathcal{F}}\right]\leq K\left(d_2(f) + \frac{1}{\sqrt{\bar{N}_T}}\bigg\|\max_{q\in Q(t,i)} \big|\mathcal{F}(\boldsymbol{X}_t,a_{ti},r_{ti}) -\mathbb{E}\big\{\mathcal{F}(\boldsymbol{X}_t,a_{ti},r_{ti}) \mid \mathcal{H}_{t-1}\big\} \big|\bigg\|_1\right)
\end{equation}
Since the right hand side is $O_p(T^{-1/2})$, we have
\begin{equation}\label{eq:delta_1}
    \Delta_1 =\frac{1}{\sqrt{\bar{N}_T}}\sum_q \mathcal{F}(\boldsymbol{X}_t,a_{ti},r_{ti}) \leq  \frac{1}{\sqrt{\bar{N}_T}}\mathbb{E}\left[\|\mathbb{G}_n\|_{\mathcal{F}}\right] = O_p(T^{-1}) = o_p(T^{-1/2}).
\end{equation}

Now let's derive the order for $\Delta_2$.
\begin{equation}\label{eq:delta_2}
\begin{aligned}
    \Delta_2 &= \frac{1}{\bar{N}_T}\sum_{t=1}^T \sum_{i=1}^{N_t} \big\{\hat{\kappa}_{t-1}(\boldsymbol{X}_{ti}) - {\kappa}_{t-1}(\boldsymbol{X}_{ti})\big\} \cdot \left[\frac{\boldsymbol{1}\{a_{ti} = \widehat{\pi}_{t-1}(\boldsymbol{X}_{ti})\}\Big\{{\mu}^{(t,i)}(\boldsymbol{X}_t,\widehat{\pi}_{t-1}(\boldsymbol{X}_{t})) -\widehat{\mu}_{t-1}^{(t,i)}(\boldsymbol{X}_t,\widehat{\pi}_{t-1}(\boldsymbol{X}_{t})) \Big\}}{\{1-\hat{\kappa}_{t-1}(\boldsymbol{X}_{ti})\}\{1-{\kappa}_{t-1}(\boldsymbol{X}_{ti})\}}  \right]\\
    & \leq \frac{1}{\bar{N}_T}\sum_{t=1}^T \sum_{i=1}^{N_t} C_2 \big|\hat{\kappa}_{t-1}(\boldsymbol{X}_{ti}) - {\kappa}_{t-1}(\boldsymbol{X}_{ti})\big|\cdot \big|\widehat{\mu}_{t-1}^{(t,i)}(\boldsymbol{X}_t,\widehat{\pi}_{t-1}(\boldsymbol{X}_{t})) - {\mu}^{(t,i)}(\boldsymbol{X}_t,\widehat{\pi}_{t-1}(\boldsymbol{X}_{t}))\big| \\
    &\leq C_2\sqrt{\frac{1}{\bar{N}_T}\sum_{t=1}^T \sum_{i=1}^{N_t} \big|\hat{\kappa}_{t-1}(\boldsymbol{X}_{ti}) - {\kappa}_{t-1}(\boldsymbol{X}_{ti})\big|^2 \cdot\frac{1}{\bar{N}_T}\sum_{t=1}^T \sum_{i=1}^{N_t} \big|\widehat{\mu}_{t-1}^{(t,i)}(\boldsymbol{X}_t,\widehat{\pi}_{t-1}(\boldsymbol{X}_{t})) - {\mu}^{(t,i)}(\boldsymbol{X}_t,\widehat{\pi}_{t-1}(\boldsymbol{X}_{t}))\big|^2 } \\
    &= o_p(\bar{N}_T^{-1/2}),
\end{aligned}
\end{equation}
where the last line holds by Cauchy-Schwartz inequality, and the last line holds by Assumption \ref{assump:4}.

Combining the results of Equation \eqref{eq:delta_1}
 and \eqref{eq:delta_2}, we have 
 \begin{equation}
     \widehat{V}_T - \breve{V}_T  = \Delta_1 + \Delta_2 =o_p(\bar{N}_T^{-1/2}) + o_p(\bar{N}_T^{-1/2}) = o_p(\bar{N}_T^{-1/2}).
 \end{equation}

Now the question becomes proving (2) $\breve{V}_T = \widetilde{V}_T + o_p(\bar{N}_T^{-1/2})$.

\begin{equation*}
\begin{aligned}
     \breve{V}_T - \widetilde{V}_T & = \frac{1}{\bar{N}_T}\sum_{t=1}^T \sum_{i=1}^{N_t} \left[1-\frac{\boldsymbol{1}\{a_{ti} = \widehat{\pi}_{t-1}(\boldsymbol{X}_{ti})\}}{1-{\kappa}_{t-1}(\boldsymbol{X}_{ti})}  \right]\cdot \Big\{\widehat{\mu}_{t-1}^{(t,i)}(\boldsymbol{X}_t,\widehat{\pi}_{t-1}(\boldsymbol{X}_{t}))-{\mu}^{(t,i)}(\boldsymbol{X}_t,\widehat{\pi}_{t-1}(\boldsymbol{X}_{t})) \Big\}.
\end{aligned}
\end{equation*}
Following similar structure as we prove $\Delta_1 = o_p({\bar{N}_T}^{-1/2})$, one can define a new class of functions 
\begin{equation*}
    \mathcal{F}'(\boldsymbol{X}_t,a_{ti},r_{ti}) = \Bigg\{\left[1-\frac{\boldsymbol{1}\{a_{ti} = \widehat{\pi}_{t-1}(\boldsymbol{X}_{ti})\}}{1-{\kappa}_{t-1}(\boldsymbol{X}_{ti})}  \right]\cdot \Big\{\widehat{\mu}_{t-1}^{(t,i)}(\boldsymbol{X}_t,\widehat{\pi}_{t-1}(\boldsymbol{X}_{t}))-{\mu}^{(t,i)}(\boldsymbol{X}_t,\widehat{\pi}_{t-1}(\boldsymbol{X}_{t})) \Big\}: \widehat{\mu}_{t-1}^{(t,i)}, {\mu}\in \Lambda,\widehat{\pi}_{t}\in \Pi\Bigg\},
\end{equation*}
and using Rosenthal’s inequality for Martingale to prove that $ \breve{V}_T - \widetilde{V}_T = o_p(\bar{N}_T^{-1/2})$.

\noindent\textbf{Step 2:} Prove that $\widetilde{V}_T = \bar{V}_T + o_p(\bar{N}_T^{-1/2})$.\\

By definition of $\widetilde{V}_T$ and $\bar{V}_T$, we have
\begin{equation}
\begin{aligned}
    \sqrt{\bar{N}_T}(\widetilde{V}_T - \bar{V}_T) &= \underbrace{\frac{1}{\sqrt{\bar{N}_T}}\sum_{t=1}^T \sum_{i=1}^{N_t} \left[\frac{\boldsymbol{1}\{a_{ti} = \widehat{\pi}_{t-1}(\boldsymbol{X}_{ti})\}}{1-{\kappa}_{t-1}(\boldsymbol{X}_{ti})}  -1\right]\cdot \Big\{{\mu}^{(t,i)}(\boldsymbol{X}_t,{\pi}^*(\boldsymbol{X}_{t}))-{\mu}^{(t,i)}(\boldsymbol{X}_t,\widehat{\pi}_{t-1}(\boldsymbol{X}_{t})) \Big\}}_{\Delta_3}\\
    & + \underbrace{\frac{1}{\sqrt{\bar{N}_T}}\sum_{t=1}^T \sum_{i=1}^{N_t} \left[\frac{\boldsymbol{1}\{a_{ti} = \widehat{\pi}_{t-1}(\boldsymbol{X}_{ti})\}}{1-{\kappa}_{t-1}(\boldsymbol{X}_{ti})}  -\frac{\boldsymbol{1}\{a_{ti} = {\pi}^*(\boldsymbol{X}_{ti})\}}{\mathbb{P}(a_{ti} = \pi^*(\boldsymbol{X}_{ti}))}\right]\cdot \Big\{r_{ti}-{\mu}^{(t,i)}(\boldsymbol{X}_t,{\pi}^*(\boldsymbol{X}_{t})) \Big\}}_{\Delta_4}.
\end{aligned}
\end{equation}
\noindent\textbf{Step 2.1:} We start from proving $\Delta_3 = o_p(1)$. Since ${\kappa}_t(\boldsymbol{X}_{ti})\leq C_1<1$, $\left|\frac{\boldsymbol{1}\{a_{ti} = \widehat{\pi}_{t-1}(\boldsymbol{X}_{ti})\}}{1-{\kappa}_{t-1}(\boldsymbol{X}_{ti})}  -1\right| $ is upper bounded by a constant. Therefore, to prove $\Delta_3= o_p(1)$, it suffice to show that 
\begin{equation}\label{eq:10}
    \begin{aligned}
        \frac{1}{\sqrt{\bar{N}_T}}\sum_{t=1}^T \sum_{i=1}^{N_t} \left[{\mu}^{(t,i)}(\boldsymbol{X}_t,{\pi}^*(\boldsymbol{X}_{t}))-{\mu}^{(t,i)}(\boldsymbol{X}_t,\widehat{\pi}_{t-1}(\boldsymbol{X}_{t})) \right] =o_p(1).
    \end{aligned}
\end{equation}
Before proceeding, let's break down this term to do some transformation. Notice that
\begin{equation}\label{eq:mu_it}
    \begin{aligned}
        \sum_{i=1}^{N_t} {\mu}^{(t,i)}(\boldsymbol{X}_t,\widehat{\pi}_{t-1}(\boldsymbol{X}_{t})) &=\sum_{i=1}^{N_t} \sum_{j=1}^{N_t} W_{t,ij}\big(\boldsymbol{X}_{tj}'\boldsymbol{\beta}_0 \cdot \boldsymbol{1}\{\widehat{\pi}_{t-1}(\boldsymbol{X}_{tj})=0\}+ \boldsymbol{X}_{tj}'\boldsymbol{\beta}_1\cdot \boldsymbol{1}\{\widehat{\pi}_{t-1}(\boldsymbol{X}_{tj})=1\} \big)\\
        &=\sum_{j=1}^{N_t} \sum_{i=1}^{N_t} W_{jit}\big(\boldsymbol{X}_{ti}'\boldsymbol{\beta}_0 \cdot \boldsymbol{1}\{\widehat{\pi}_{t-1}(\boldsymbol{X}_{ti})=0\}+ \boldsymbol{X}_{ti}'\boldsymbol{\beta}_1\cdot \boldsymbol{1}\{\widehat{\pi}_{t-1}(\boldsymbol{X}_{ti})=1\} \big)\\
        & = \sum_{i=1}^{N_t}  \Big\{\sum_{j=1}^{N_t}  W_{jit}\Big\}\cdot \big(\boldsymbol{X}_{ti}'\boldsymbol{\beta}_0 \cdot \boldsymbol{1}\{\widehat{\pi}_{t-1}(\boldsymbol{X}_{ti})=0\}+ \boldsymbol{X}_{ti}'\boldsymbol{\beta}_1\cdot \boldsymbol{1}\{\widehat{\pi}_{t-1}(\boldsymbol{X}_{ti})=1\} \big)\\
        & = \sum_{i=1}^{N_t}  \left[\boldsymbol{1}\big\{\omega_{ti}\boldsymbol{X}_{ti}'(\widehat{\boldsymbol{\beta}}_{t-1,1}-\widehat{\boldsymbol{\beta}}_{t-1,0}) > 0\big\}\cdot \omega_{ti}\boldsymbol{X}_{ti}'(\boldsymbol{\beta}_1-\boldsymbol{\beta}_0) + \omega_{ti}\boldsymbol{X}_{ti}'\boldsymbol{\beta}_0\right].
    \end{aligned}
\end{equation}
where the second equality holds by switching the index of $(i,j)$ to $(j,i)$, and the third equality holds by Fubini's theorem.

Going back to the previous equation, we have
\begin{equation*}
    \begin{aligned}
        &\frac{1}{\sqrt{\bar{N}_T}}\sum_{t=1}^T \sum_{i=1}^{N_t} \left[{\mu}^{(t,i)}(\boldsymbol{X}_t,{\pi}^*(\boldsymbol{X}_{t}))-{\mu}^{(t,i)}(\boldsymbol{X}_t,\widehat{\pi}_{t-1}(\boldsymbol{X}_{t})) \right] \\
        &= \frac{1}{\sqrt{\bar{N}_T}}\sum_{t=1}^T \sum_{i=1}^{N_t} \left[ \boldsymbol{1}\big\{\omega_{ti}\boldsymbol{X}_{ti}'(\widehat{\boldsymbol{\beta}}_{t-1,1}-\widehat{\boldsymbol{\beta}}_{t-1,0}) > 0\big\}-\boldsymbol{1}\big\{\omega_{ti}\boldsymbol{X}_{ti}'({\boldsymbol{\beta}}_{1}-{\boldsymbol{\beta}}_{0}) > 0\big\}\right]\cdot \omega_{ti}\boldsymbol{X}_{ti}'(\boldsymbol{\beta}_1-\boldsymbol{\beta}_0)\\
        & \leq \frac{1}{\sqrt{\bar{N}_T}}\sum_{t=1}^T \sum_{i=1}^{N_t} \bigg|\left[ \boldsymbol{1}\big\{\omega_{ti}\boldsymbol{X}_{ti}'(\widehat{\boldsymbol{\beta}}_{t-1,1}-\widehat{\boldsymbol{\beta}}_{t-1,0}) > 0\big\}-\boldsymbol{1}\big\{\omega_{ti}\boldsymbol{X}_{ti}'({\boldsymbol{\beta}}_{1}-{\boldsymbol{\beta}}_{0}) > 0\big\}\right]\cdot \omega_{ti}\boldsymbol{X}_{ti}'(\boldsymbol{\beta}_1-\boldsymbol{\beta}_0)\bigg|
    \end{aligned}
\end{equation*}
Again, for the brevity of notation, we denote $\widehat{\zeta}_{ti} = \omega_{ti}\boldsymbol{X}_{ti}'(\widehat{\boldsymbol{\beta}}_{t-1,1}-\widehat{\boldsymbol{\beta}}_{t-1,0})$, and ${\zeta}_{ti} = \omega_{ti}\boldsymbol{X}_{ti}'({\boldsymbol{\beta}}_{1}-{\boldsymbol{\beta}}_{0})$.
The RHS of the above equation is thus equivalent to
\begin{equation*}
    \frac{1}{\sqrt{\bar{N}_T}}\sum_{t=1}^T \sum_{i=1}^{N_t} \bigg|\left(\boldsymbol{1}\big\{\widehat{\zeta}_{ti} > 0\big\}-\boldsymbol{1}\big\{{\zeta}_{ti} > 0\big\}\right)\cdot {\zeta}_{ti}\bigg|.
\end{equation*}

Let's first consider the case where ${\zeta}_{ti} >0$. The opposite scenario can be derived in a similar manner. When ${\zeta}_{ti} >0$,
\begin{equation*}
    \begin{aligned}
        &{\mu}^{(t,i)}(\boldsymbol{X}_t,{\pi}^*(\boldsymbol{X}_{t}))-{\mu}^{(t,i)}(\boldsymbol{X}_t,\widehat{\pi}_{t-1}(\boldsymbol{X}_{t}))  = - \boldsymbol{1}\big\{\widehat{\zeta}_{ti} \leq 0\big\}\cdot {\zeta}_{ti} \leq 0.
    \end{aligned}
\end{equation*}
Since $\boldsymbol{1}\big\{\widehat{\zeta}_{ti} \leq 0\big\}\cdot \widehat{\zeta}_{ti}\leq 0$, we have
\begin{equation*}
    \begin{aligned}
        &\big|{\mu}^{(t,i)}(\boldsymbol{X}_t,{\pi}^*(\boldsymbol{X}_{t}))-{\mu}^{(t,i)}(\boldsymbol{X}_t,\widehat{\pi}_{t-1}(\boldsymbol{X}_{t}))\big|  =  \boldsymbol{1}\big\{\widehat{\zeta}_{ti} \leq 0\big\}\cdot {\zeta}_{ti} \leq  \boldsymbol{1}\big\{\widehat{\zeta}_{ti} \leq 0\big\}\cdot \big({\zeta}_{ti} -\widehat{\zeta}_{ti} \big).
    \end{aligned}
\end{equation*}

To show that ${\bar{N}_T}^{-1/2}\sum_{t=1}^T \sum_{i=1}^{N_t} \big|{\mu}^{(t,i)}(\boldsymbol{X}_t,{\pi}^*(\boldsymbol{X}_{t}))-{\mu}^{(t,i)}(\boldsymbol{X}_t,\widehat{\pi}_{t-1}(\boldsymbol{X}_{t}))\big| = o_p(1)$, it suffice to prove 
\begin{equation*}
    \boldsymbol{\zeta} := {\bar{N}_T}^{-1/2}\sum_{t=1}^T \sum_{i=1}^{N_t}\boldsymbol{1}\big\{\widehat{\zeta}_{ti} \leq 0\big\}\cdot \big({\zeta}_{ti} -\widehat{\zeta}_{ti} \big)= o_p(1).
\end{equation*}
For any $\alpha\in(0,{1}/{2})$, we can further decompose
\begin{equation*}
\begin{aligned}
    \boldsymbol{\zeta} &= \underbrace{\mathbb{P}(0<\zeta_{ti}<\bar{N}_T^{-\alpha})\cdot \frac{1}{\sqrt{\bar{N}_T}}\sum_{t=1}^T \sum_{i=1}^{N_t}\boldsymbol{1}\big\{\widehat{\zeta}_{ti} \leq 0\big\}\cdot \big({\zeta}_{ti} -\widehat{\zeta}_{ti} \big)\cdot \boldsymbol{1}\{0<\zeta_{ti}<\bar{N}_T^{-\alpha}\}}_{\boldsymbol{\zeta}_1} \\
    &+ \underbrace{\mathbb{P}(\zeta_{ti}\geq \bar{N}_T^{-\alpha})\cdot\frac{1}{\sqrt{\bar{N}_T}}\sum_{t=1}^T \sum_{i=1}^{N_t}\boldsymbol{1}\big\{\widehat{\zeta}_{ti} \leq 0\big\}\cdot \big({\zeta}_{ti} -\widehat{\zeta}_{ti} \big)\cdot \boldsymbol{1}\{\zeta_{ti}\geq \bar{N}_T^{-\alpha}\}}_{\boldsymbol{\zeta}_2}.
\end{aligned}
\end{equation*}
First, we show $\boldsymbol{\zeta}_1 = o_p(1)$.
According to Theorem \ref{thm:1}, $\widehat{\zeta}_{ti}-{\zeta}_{ti} = O_p(\bar{N}_t^{-1/2}) = o_p(\bar{N}_t^{-(1/2-\alpha\gamma)})$ for any $\alpha\gamma>0$.
Therefore, 
\begin{equation*}
\begin{aligned}
    \frac{1}{\sqrt{\bar{N}_T}}&\sum_{t=1}^T \sum_{i=1}^{N_t} \left|\boldsymbol{1}\big\{\widehat{\zeta}_{ti} \leq 0\big\}\cdot \big({\zeta}_{ti} -\widehat{\zeta}_{ti} \big)\cdot \boldsymbol{1}\{0<\zeta_{ti}<\bar{N}_T^{-\alpha}\}\right|\leq \frac{1}{\sqrt{\bar{N}_T}}\sum_{t=1}^T \sum_{i=1}^{N_t}\left| {\zeta}_{ti} -\widehat{\zeta}_{ti} \right|\\
    &\leq \sqrt{\bar{N}_T}\cdot \frac{1}{{\bar{N}_T}}\sum_{t=1}^T \sum_{i=1}^{N_t}\left| {\zeta}_{ti} -\widehat{\zeta}_{ti} \right| = \sqrt{\bar{N}_T}\cdot o_p(\bar{N}_T^{-(1/2-\alpha\gamma)}) =o_p(\bar{N}_T^{\alpha\gamma}),
\end{aligned}
\end{equation*}
where the second last equality holds by Lemma 6 in \citet{luedtke2016statistical}. 

Since $|\omega|\geq 1$, by setting $\epsilon = \bar{N}_T^{-\alpha}$ in Assumption \ref{assump:5t}, we have $\mathbb{P}\left(0<|\omega f(\boldsymbol{X},1) - \omega f(\boldsymbol{X},0)| <\bar{N}_T^{-\alpha}\right) \leq \mathbb{P}\left(0<|f(\boldsymbol{X},1) - f(\boldsymbol{X},0)| <\bar{N}_T^{-\alpha}\right) = O(\bar{N}_T^{-\alpha\gamma})$. Therefore, 
\begin{equation}\label{eq:boldzeta1}
\begin{aligned}
    \boldsymbol{\zeta}_1 &= \mathbb{P}(0<\zeta_{ti}<\bar{N}_T^{-\alpha})\cdot \frac{1}{\sqrt{\bar{N}_T}}\sum_{t=1}^T \sum_{i=1}^{N_t}\boldsymbol{1}\big\{\widehat{\zeta}_{ti} \leq 0\big\}\cdot \big({\zeta}_{ti} -\widehat{\zeta}_{ti} \big)\cdot \boldsymbol{1}\{0<\zeta_{ti}<\bar{N}_T^{-\alpha}\} \\
    &\leq \left|\mathbb{P}(0<\zeta_{ti}<\bar{N}_T^{-\alpha}) \right| \cdot \frac{1}{\sqrt{\bar{N}_T}}\sum_{t=1}^T \sum_{i=1}^{N_t} \left|\boldsymbol{1}\big\{\widehat{\zeta}_{ti} \leq 0\big\}\cdot \big({\zeta}_{ti} -\widehat{\zeta}_{ti} \big)\cdot \boldsymbol{1}\{0<\zeta_{ti}<\bar{N}_T^{-\alpha}\}\right|\\
    &\leq O(\bar{N}_T^{-\alpha\gamma}) \cdot o_p(\bar{N}_T^{\alpha\gamma})= o_p(1).
\end{aligned}
\end{equation}
Next, we show $\boldsymbol{\zeta}_2 = o_p(1)$.

Since $\boldsymbol{1}\{\widehat{\zeta}_{ti}\leq 0\} = \boldsymbol{1}\{\widehat{\zeta}_{ti} - {\zeta}_{ti}\leq -{\zeta}_{ti}\} = \boldsymbol{1}\{|\widehat{\zeta}_{ti}-{\zeta}_{ti}|>{\zeta}_{ti}\} $, we have
\begin{equation}\label{eq:zeta1}
    \left|\boldsymbol{1}\{\widehat{\zeta}_{ti}\leq 0\}(\widehat{\zeta}_{ti}-\zeta_{ti})\right|  = \boldsymbol{1}\big\{|\widehat{\zeta}_{ti}-{\zeta}_{ti}|>{\zeta}_{ti}\big\}\cdot \big|\widehat{\zeta}_{ti}-\zeta_{ti}| \leq \frac{|\widehat{\zeta}_{ti}-\zeta_{ti}|}{\zeta_{ti}} \cdot\big|\widehat{\zeta}_{ti}-\zeta_{ti}\big| =  \frac{|\widehat{\zeta}_{ti}-\zeta_{ti}|^2}{\zeta_{ti}} .
\end{equation}

Since we assumed that $\zeta_{ti}>0$, based on the result of Equation \eqref{eq:zeta1}, we further have 
\begin{equation*}
    \boldsymbol{1}\{\widehat{\zeta}_{ti}\leq 0\}(\zeta_{ti}-\widehat{\zeta}_{ti}) \leq \frac{|\widehat{\zeta}_{ti}-\zeta_{ti}|^2}{\zeta_{ti}} 
\end{equation*}
as $\zeta_{ti}-\widehat{\zeta}_{ti}\geq 0$ always holds. Additionally, notice that $\boldsymbol{1}\{\zeta_{ti}\geq \bar{N}_T^{-\alpha}\}\leq \zeta_{ti}\bar{N}_T^{\alpha}$. Therefore,
\begin{equation*}
\begin{aligned}
    \boldsymbol{\zeta}_2 &= \mathbb{P}(\zeta_{ti}\geq \bar{N}_T^{-\alpha})\cdot\frac{1}{\sqrt{\bar{N}_T}}\sum_{t=1}^T \sum_{i=1}^{N_t}\boldsymbol{1}\big\{\widehat{\zeta}_{ti} \leq 0\big\}\cdot \big({\zeta}_{ti} -\widehat{\zeta}_{ti} \big)\cdot \boldsymbol{1}\{\zeta_{ti}\geq \bar{N}_T^{-\alpha}\} \\
    &\leq \frac{1}{\sqrt{\bar{N}_T}}\sum_{t=1}^T \sum_{i=1}^{N_t}\frac{|\widehat{\zeta}_{ti}-\zeta_{ti}|^2}{\zeta_{ti}} \cdot  \zeta_{ti}\bar{N}_T^{\alpha} = \bar{N}_T^{1/2+\alpha}\cdot  \frac{1}{{\bar{N}_T}}\sum_{t=1}^T \sum_{i=1}^{N_t}|\widehat{\zeta}_{ti}-\zeta_{ti}|^2.
\end{aligned}
\end{equation*}
By Theorem \ref{thm:1}, $|\widehat{\zeta}_{ti}-\zeta_{ti}| = O_p(\bar{N}_t^{-1/2})$, which implies $|\widehat{\zeta}_{ti}-\zeta_{ti}|^2 = O_p(\bar{N}_t^{-1})$. According to Lemma 6 of \citet{luedtke2016statistical}, ${\bar{N}_T}^{-1}\sum_{t=1}^T \sum_{i=1}^{N_t}|\widehat{\zeta}_{ti}-\zeta_{ti}|^2 = O_p(\bar{N}_T^{-1})$. Therefore,
\begin{equation}\label{eq:boldzeta2}
\begin{aligned}
    \boldsymbol{\zeta}_2 &\leq \bar{N}_T^{1/2+\alpha}\cdot  \frac{1}{{\bar{N}_T}}\sum_{t=1}^T \sum_{i=1}^{N_t}|\widehat{\zeta}_{ti}-\zeta_{ti}|^2\leq \bar{N}_T^{1/2+\alpha}\cdot O_p(\bar{N}_T^{-1}) = o_p(1)
\end{aligned}
\end{equation}
for any $\alpha<1/2$.

Combining the result of Equation \eqref{eq:boldzeta1} and Equation \eqref{eq:boldzeta2}, we have 
\begin{equation}
    \boldsymbol{\zeta} = \boldsymbol{\zeta}_1 + \boldsymbol{\zeta}_2 = o_p(1) + o_p(1) = o_p(1).
\end{equation}

Therefore, ${\bar{N}_T}^{-1/2}\sum_{t=1}^T \sum_{i=1}^{N_t} \big|{\mu}^{(t,i)}(\boldsymbol{X}_t,{\pi}^*(\boldsymbol{X}_{t}))-{\mu}^{(t,i)}(\boldsymbol{X}_t,\widehat{\pi}_{t-1}(\boldsymbol{X}_{t}))\big| = o_p(1)$, and thus $\Delta_3 = o_p(1)$. The proof of first part is done.

\noindent\textbf{Step 2.2:} Next, we show that $\Delta_4 = o_p(1)$ as well. 

Recall that
\begin{equation}
\begin{aligned}
    \Delta_4 &= \frac{1}{\sqrt{\bar{N}_T}}\sum_{t=1}^T \sum_{i=1}^{N_t} \left[\frac{\boldsymbol{1}\{a_{ti} = \widehat{\pi}_{t-1}(\boldsymbol{X}_{ti})\}}{1-{\kappa}_{t-1}(\boldsymbol{X}_{ti})}  -\frac{\boldsymbol{1}\{a_{ti} = {\pi}^*(\boldsymbol{X}_{ti})\}}{\mathbb{P}(a_{ti} = \pi^*(\boldsymbol{X}_{ti}))}\right]\cdot \Big\{r_{ti}-{\mu}^{(t,i)}(\boldsymbol{X}_t,{\pi}^*(\boldsymbol{X}_{t})) \Big\}\\
    & = \frac{1}{\sqrt{\bar{N}_T}}\sum_{t=1}^T \sum_{i=1}^{N_t} \left[\frac{\boldsymbol{1}\{a_{ti} = \widehat{\pi}_{t-1}(\boldsymbol{X}_{ti})\}}{1-{\kappa}_{t-1}(\boldsymbol{X}_{ti})}  -\frac{\boldsymbol{1}\{a_{ti} = {\pi}^*(\boldsymbol{X}_{ti})\}}{\mathbb{P}(a_{ti} = \pi^*(\boldsymbol{X}_{ti}))}\right]\cdot \Big\{r_{ti}-{\mu}^{(t,i)}(\boldsymbol{X}_t,{\pi}^*(\boldsymbol{X}_{t})) \Big\}
\boldsymbol{1}\{a_{ti} = {\pi}^*(\boldsymbol{X}_{ti})\}\\
& + \frac{1}{\sqrt{\bar{N}_T}}\sum_{t=1}^T \sum_{i=1}^{N_t} \left[\frac{\boldsymbol{1}\{a_{ti} = \widehat{\pi}_{t-1}(\boldsymbol{X}_{ti})\}}{1-{\kappa}_{t-1}(\boldsymbol{X}_{ti})} \right]\cdot \Big\{r_{ti}-{\mu}^{(t,i)}(\boldsymbol{X}_t,{\pi}^*(\boldsymbol{X}_{t})) \Big\}
\cdot \boldsymbol{1}\{a_{ti} \neq {\pi}^*(\boldsymbol{X}_{ti})\}\\
& = \underbrace{\frac{1}{\sqrt{\bar{N}_T}}\sum_{t=1}^T \sum_{i=1}^{N_t} \left[\frac{\boldsymbol{1}\{a_{ti} = \widehat{\pi}_{t-1}(\boldsymbol{X}_{ti})\}}{1-{\kappa}_{t-1}(\boldsymbol{X}_{ti})}  -\frac{\boldsymbol{1}\{a_{ti} = {\pi}^*(\boldsymbol{X}_{ti})\}}{\mathbb{P}(a_{ti} = \pi^*(\boldsymbol{X}_{ti}))}\right]\cdot \epsilon_{ti}\cdot 
\boldsymbol{1}\{a_{ti} = {\pi}^*(\boldsymbol{X}_{ti})\}}_{\boldsymbol{\zeta}_3}\\
& + \underbrace{\frac{1}{\sqrt{\bar{N}_T}}\sum_{t=1}^T \sum_{i=1}^{N_t} \left[\frac{\boldsymbol{1}\{a_{ti} = \widehat{\pi}_{t-1}(\boldsymbol{X}_{ti})\}}{1-{\kappa}_{t-1}(\boldsymbol{X}_{ti})} \right]\cdot \Big\{{\mu}^{(t,i)}(\boldsymbol{X}_t,\widehat{\pi}_{t-1}(\boldsymbol{X}_{t}))-{\mu}^{(t,i)}(\boldsymbol{X}_t,{\pi}^*(\boldsymbol{X}_{t})) \Big\}
\cdot \boldsymbol{1}\{a_{ti} \neq {\pi}^*(\boldsymbol{X}_{ti})\}}_{\boldsymbol{\zeta}_4}\\
& +\underbrace{\frac{1}{\sqrt{\bar{N}_T}}\sum_{t=1}^T \sum_{i=1}^{N_t} \left[\frac{\boldsymbol{1}\{a_{ti} = \widehat{\pi}_{t-1}(\boldsymbol{X}_{ti})\}}{1-{\kappa}_{t-1}(\boldsymbol{X}_{ti})} \right]\cdot \epsilon_{ti}
\cdot \boldsymbol{1}\{a_{ti} \neq {\pi}^*(\boldsymbol{X}_{ti})\}}_{\boldsymbol{\zeta}_5}.
\end{aligned}
\end{equation}
We only need to show $\boldsymbol{\zeta}_3$, $\boldsymbol{\zeta}_4$ and $\boldsymbol{\zeta}_5$ are all $o_p(1)$. The proof for $\boldsymbol{\zeta}_5$ is similar to that for $\boldsymbol{\zeta}_3$ using Rosenthal's inequality for Martingales. Therefore, we will focus on proving $\boldsymbol{\zeta}_3$ and $\boldsymbol{\zeta}_4$, and omit the details for $\boldsymbol{\zeta}_5$ for brevity.

To prove $\boldsymbol{\zeta}_3 =o_p(1)$, we define a function class
\begin{equation*}
    \mathcal{F}(\boldsymbol{X}_t,a_{ti},\epsilon_{ti}) = \Bigg\{\left[\frac{\boldsymbol{1}\{a_{ti} = \widehat{\pi}_{t-1}(\boldsymbol{X}_{ti})\}}{1-{\kappa}_{t-1}(\boldsymbol{X}_{ti})}  -\frac{\boldsymbol{1}\{a_{ti} = {\pi}^*(\boldsymbol{X}_{ti})\}}{\mathbb{P}(a_{ti} = \pi^*(\boldsymbol{X}_{ti}))}\right]\cdot \epsilon_{ti}\cdot 
\boldsymbol{1}\{a_{ti} = {\pi}^*(\boldsymbol{X}_{ti})\}: {\kappa}_t\in \Lambda,\widehat{\pi}_{t}\in \Pi\Bigg\},
\end{equation*}
where $\Lambda$ and $\Pi$ are two classes of functions mapping context $\boldsymbol{X}_{ti}$ to a probability in $[0,1]$. Define the supremum of the empirical process indexed by $\mathcal{F}$ as
\begin{equation}\label{eq:Gn_f2}
    \|\mathbb{G}_n  \|_{\mathcal{F}} := \sup_{\pi\in\Pi} \bigg|\frac{1}{\sqrt{\bar{N}_T}}\sum_{q\in Q(t,i)} \left[\mathcal{F}(\boldsymbol{X}_t,a_{ti},\epsilon_{ti}) -\mathbb{E}\big\{\mathcal{F}(\boldsymbol{X}_t,a_{ti},\epsilon_{ti}) \mid \mathcal{H}_{t-1}\big\}\right] \bigg|.
\end{equation}
Since $\mathbb{E}[\epsilon_{ti}|\boldsymbol{X}_{ti},\boldsymbol{A}_t] = 0$, according to the iteration of expectation, the second term in the above equation can be derived as
\begin{equation*}
\begin{aligned}
    &\mathbb{E}\big\{\mathcal{F}(\boldsymbol{X}_t,a_{ti},\epsilon_{ti}) \mid \mathcal{H}_{t-1}\big\} \\
    &=\mathbb{E} \Bigg[\left[\frac{\boldsymbol{1}\{a_{ti} = \widehat{\pi}_{t-1}(\boldsymbol{X}_{ti})\}}{1-{\kappa}_{t-1}(\boldsymbol{X}_{ti})}  -\frac{\boldsymbol{1}\{a_{ti} = {\pi}^*(\boldsymbol{X}_{ti})\}}{\mathbb{P}(a_{ti} = \pi^*(\boldsymbol{X}_{ti}))}\right]\cdot 
\boldsymbol{1}\{a_{ti} = {\pi}^*(\boldsymbol{X}_{ti})\}\cdot \mathbb{E}[\epsilon_{ti}|\boldsymbol{X}_{ti},\boldsymbol{A}_{t}]\text{ }\Big|\text{ }\mathcal{H}_{t-1}\Bigg]=0,
\end{aligned}.
\end{equation*}

Therefore, Equation \eqref{eq:Gn_f2} can be simplified as
\begin{equation*}
    \|\mathbb{G}_n  \|_{\mathcal{F}} = \sup_{\pi\in\Pi} \bigg|\frac{1}{\sqrt{\bar{N}_T}}\sum_{q\in Q(t,i)}\mathcal{F}(\boldsymbol{X}_t,a_{ti},\epsilon_{ti})  \bigg|.
\end{equation*}
Following a similar derivation structure as that used between Equation \eqref{eq:Gn_f} and Equation \eqref{eq:EGn_f} in Step 1, we have 
\begin{equation}\label{eq:EGn_f2}
    \boldsymbol{\zeta}_3 \leq \mathbb{E}\left[\|\mathbb{G}_n\|_{\mathcal{F}}\right]\leq K\left(d_2(f) + \frac{1}{\sqrt{\bar{N}_T}}\bigg\|\max_{q\in Q(t,i)} \big|\mathcal{F}(\boldsymbol{X}_t,a_{ti},\epsilon_{ti}) -\mathbb{E}\big\{\mathcal{F}(\boldsymbol{X}_t,a_{ti},\epsilon_{ti}) \mid \mathcal{H}_{t-1}\big\} \big|\bigg\|_1\right) = o_p(1).
\end{equation}

Next, let's prove $\boldsymbol{\zeta}_4 = o_p(1)$. Since both $\frac{\boldsymbol{1}\{a_{ti} = \widehat{\pi}_{t-1}(\boldsymbol{X}_{ti})\}}{1-{\kappa}_{t-1}(\boldsymbol{X}_{ti})}$ and $\boldsymbol{1}\{a_{ti} \neq {\pi}^*(\boldsymbol{X}_{ti})\}$ can be upper bounded, it suffice to prove that
\begin{equation}
    \frac{1}{\sqrt{\bar{N}_T}}\sum_{t=1}^T \sum_{i=1}^{N_t}  \Big\{{\mu}^{(t,i)}(\boldsymbol{X}_t,\widehat{\pi}_{t-1}(\boldsymbol{X}_{t}))-{\mu}^{(t,i)}(\boldsymbol{X}_t,{\pi}^*(\boldsymbol{X}_{t})) \Big\} = o_p(1),
\end{equation}
which has already been established in Equation \eqref{eq:10} in Step 2.1. Therefore, $\boldsymbol{\zeta}_4 = o_p(1)$.

Combining the results above, we have 
\begin{equation}
    \Delta_4 = \boldsymbol{\zeta}_3+\boldsymbol{\zeta}_4+\boldsymbol{\zeta}_5 = o_p(1).
\end{equation}
The proof of Step 2 is thus complete.

\noindent\textbf{Step 3:} Prove that $
    \sqrt{\bar{N}_T}(\bar{V}_T -V^{\pi^*})\xrightarrow{\mathcal{D}} \mathcal{N}(0,\sigma^2_V)$ and derive the asymptotic variance $\sigma^2_V$.\\
    
Recall that 
\begin{equation*}
\begin{aligned}
    \bar{V}_T &= \frac{1}{\bar{N}_T}\sum_{t=1}^T \sum_{i=1}^{N_t} \left\{\frac{\boldsymbol{1}\{a_{ti} = {\pi}^*(\boldsymbol{X}_{ti})\}}{\mathbb{P}(a_{ti} = {\pi}^*(\boldsymbol{X}_{ti}))} \cdot \Big(r_{ti} -{\mu}^{(t,i)}(\boldsymbol{X}_t,{\pi}^*(\boldsymbol{X}_t)) \Big) + {\mu}^{(t,i)}(\boldsymbol{X}_t,{\pi}^*(\boldsymbol{X}_t))\right\}\\
    & = \frac{1}{\bar{N}_T}\sum_{t=1}^T \sum_{i=1}^{N_t} \left\{\frac{\boldsymbol{1}\{a_{ti} = {\pi}^*(\boldsymbol{X}_{ti})\}}{\mathbb{P}(a_{ti} = {\pi}^*(\boldsymbol{X}_{ti}))} \cdot \epsilon_{ti} + {\mu}^{(t,i)}(\boldsymbol{X}_t,{\pi}^*(\boldsymbol{X}_t))\right\}.
\end{aligned}
\end{equation*}
Given the derivation of Equation \eqref{eq:mu_it}, we have
\begin{equation}
    \begin{aligned}
        \sum_{i=1}^{N_t} {\mu}^{(t,i)}(\boldsymbol{X}_t,{\pi}^*(\boldsymbol{X}_{t})) &
        = \sum_{i=1}^{N_t}  \left[{\pi}^*(\boldsymbol{X}_{ti})\cdot \omega_{ti}\boldsymbol{X}_{ti}'\boldsymbol{\beta}_1 +\{1-{\pi}^*(\boldsymbol{X}_{ti})\}\cdot \omega_{ti}\boldsymbol{X}_{ti}'\boldsymbol{\beta}_0\right].
    \end{aligned}
\end{equation}
Combining the above term with the expression of $\bar{V}_T$, we have 
\begin{equation*}
\begin{aligned}
    \bar{V}_T =  \frac{1}{\bar{N}_T}\sum_{t=1}^T \sum_{i=1}^{N_t} \left\{\frac{\boldsymbol{1}\{a_{ti} = {\pi}^*(\boldsymbol{X}_{ti})\}}{\mathbb{P}(a_{ti} = {\pi}^*(\boldsymbol{X}_{ti}))} \cdot \epsilon_{ti} + \Big[{\pi}^*(\boldsymbol{X}_{ti})\cdot \omega_{ti}\boldsymbol{X}_{ti}'\boldsymbol{\beta}_1 +\{1-{\pi}^*(\boldsymbol{X}_{ti})\}\cdot \omega_{ti}\boldsymbol{X}_{ti}'\boldsymbol{\beta}_0\Big]\right\}.
\end{aligned}
\end{equation*}

To decompose, we define 
\begin{equation}
    \begin{aligned}
        \xi_{q} := \underbrace{\frac{\boldsymbol{1}\{a_{q} = {\pi}^*(\boldsymbol{X}_{q})\}}{\mathbb{P}(a_{q} = {\pi}^*(\boldsymbol{X}_{q}))} \cdot \epsilon_{q}}_{\xi_{1q}} + \underbrace{\Big[{\pi}^*(\boldsymbol{X}_{ti})\cdot \omega_{ti}\boldsymbol{X}_{ti}'\boldsymbol{\beta}_1 +\{1-{\pi}^*(\boldsymbol{X}_{ti})\}\cdot \omega_{ti}\boldsymbol{X}_{ti}'\boldsymbol{\beta}_0- V^{\pi^*}\Big]}_{\xi_{2q}},
    \end{aligned}
\end{equation}
where $q$ denotes an unit in a flattened unit queue $Q(t,i) = \sum_{s=1}^{t-1}N_s + i$. Similar to the the proof of Theorem \ref{thm:1}, we define $\mathcal{H}_{q}$ as the $\sigma-$algebra containing the information up to unit $q$ where $
\mathcal{H}_{q_0} = \sigma (\boldsymbol{v}'\widetilde{\boldsymbol{X}}_{1}\epsilon_{1},\dots, \boldsymbol{v}'\widetilde{\boldsymbol{X}}_{q_0}\epsilon_{q_0})$. 

Since
\begin{equation*}
    \mathbb{E}[\xi_{2q}] = \mathbb{E}[{\pi}^*(\boldsymbol{X}_{ti})\cdot \omega_{ti}\boldsymbol{X}_{ti}'\boldsymbol{\beta}_1 +\{1-{\pi}^*(\boldsymbol{X}_{ti})\}\cdot \omega_{ti}\boldsymbol{X}_{ti}'\boldsymbol{\beta}_0] - V^{\pi^*} = 0,
\end{equation*}
it holds that $\mathbb{E}[\xi_{2q}|\mathcal{H}_{q-1}] = 0$. Additionally, notice that 
\begin{equation*}
    \mathbb{E}[\xi_{1q}|\mathcal{H}_{q-1}] = \mathbb{E}\left[\frac{\boldsymbol{1}\{a_{q} = {\pi}^*(\boldsymbol{X}_{q})\}}{\mathbb{P}(a_{q} = {\pi}^*(\boldsymbol{X}_{q}))} \cdot \epsilon_{q}\big|\mathcal{H}_{q-1}\right] = \mathbb{E}\left[\frac{\boldsymbol{1}\{a_{q} = {\pi}^*(\boldsymbol{X}_{q})\}}{\mathbb{P}(a_{q} = {\pi}^*(\boldsymbol{X}_{q}))} \cdot \mathbb{E}[\epsilon_{q}|\mathcal{H}_{q-1},\boldsymbol{A}_t]\big|\mathcal{H}_{q-1}\right] = 0.
\end{equation*}
Thus, $\mathbb{E}[\xi_q|\mathcal{H}_{q-1}] = 0$, and $\{\xi_q\}_{1\leq q\leq \bar{N}_T}$ is a Martingale difference sequence. To show that $
    \sqrt{\bar{N}_T}(\bar{V}_T -V^{\pi^*})\xrightarrow{\mathcal{D}} \mathcal{N}(0,\sigma^2_V)$ and derive the asymptotic variance $\sigma^2_V$, it suffice to check the Lindeberg condition and use Martingale CLT to establish asymptotic normality.

\noindent \textbf{(1) First, let's 
check the Lindeberg condition.}
\begin{equation*}
    \sum_{q=1}^{\bar{N}_T} \mathbb{E}\left[ \frac{1}{\bar{N}_T}\xi_q^2 \cdot \boldsymbol{1}\Big\{\big|\frac{1}{\sqrt{\bar{N}}_T}\xi_q\big|\geq \delta\Big\} \Big| \mathcal{H}_{q-1}\right] =  \frac{1}{\bar{N}_T} \sum_{q=1}^{\bar{N}_T} \mathbb{E}\Big[\xi_q^2\boldsymbol{1}\big\{\xi_q^2\geq \bar{N}_T\delta^2\big\} \big| \mathcal{H}_{q-1}\Big].
\end{equation*}
Notice that $\xi_q^2\boldsymbol{1}\big\{\xi_q^2\geq \bar{N}_T\delta^2\big\}$ converges to $0$ as $\bar{N}_T$ goes to infinity and is bounded by $\xi_q^2$ given $\mathcal{H}_{q-1}$. Therefore, we only need to check the integrability of $\xi_q^2$ given $\mathcal{H}_{q-1}$, then by Dominated Convergence Theorem (DCT), the Lindeberg condition is checked.

Since the derivation of $\mathbb{E}[\xi_q^2|\mathcal{H}_{q-1}]$ is exactly the asymptotic variance $\sigma^2_V$, we will leave the details to part (2).

\noindent \textbf{(2) Next, we derive the limit of conditional variance $\sigma^2_V = \frac{1}{\bar{N}_T}\sum_{q=1}^{\bar{N}_T}\mathbb{E}[\xi^2_{q}|\mathcal{H}_{q-1}]$}.

\begin{equation*}
\begin{aligned}
    \mathbb{E}[\xi^2_{1q}|\mathcal{H}_{q-1}] &= \mathbb{E}\left[\frac{\boldsymbol{1}\{a_{q} = {\pi}^*(\boldsymbol{X}_{q})\}}{[\mathbb{P}\{a_{q} = {\pi}^*(\boldsymbol{X}_{q})\}]^2} \cdot \epsilon_{q}^2\big|\mathcal{H}_{q-1}\right] = \mathbb{E}\left[\frac{\boldsymbol{1}\{a_{q} = {\pi}^*(\boldsymbol{X}_{q})\}}{[\mathbb{P}\{a_{q} = {\pi}^*(\boldsymbol{X}_{q})\}]^2} \cdot \mathbb{E}[\epsilon_{q}^2|\mathcal{H}_{q-1},\boldsymbol{A}_t]\big|\mathcal{H}_{q-1}\right]\\
    & =\mathbb{E}\left[\frac{\boldsymbol{1}\{a_{q} = {\pi}^*(\boldsymbol{X}_{q})\}}{[\mathbb{P}\{a_{q} = {\pi}^*(\boldsymbol{X}_{q})\}]^2} \cdot \sigma^2\big|\mathcal{H}_{q-1}\right].
\end{aligned}
\end{equation*}
Therefore, 
\begin{equation*}
    \frac{1}{\bar{N}_T}\sum_{q=1}^{\bar{N}_T}\mathbb{E}[\xi^2_{1q}|\mathcal{H}_{q-1}] = \sigma^2\cdot  \mathbb{E}\left[\frac{1}{\bar{N}_T}\sum_{q=1}^{\bar{N}_T}\frac{1-\nu_{q}(\boldsymbol{X}_{q},\mathcal{H}_{t-1})}{[\mathbb{P}\{a_{q} = {\pi}^*(\boldsymbol{X}_{q})\}]^2}\right],
\end{equation*}
where $\nu_{q}(\boldsymbol{X}_{q},\mathcal{H}_{t-1}) = \nu_{ti}(\boldsymbol{X}_{ti},\mathcal{H}_{t-1}) = \mathbb{P}({A}_{ti} \neq  \pi^*(\boldsymbol{X}_{ti})|\boldsymbol{X}_{ti},\mathcal{H}_{t-1})$.

Following similar proof structure of \citet{ye2023doubly} in Appendix page 34-35, we are able to establish that 
\begin{equation*}
    \frac{1}{\bar{N}_T}\sum_{q=1}^{\bar{N}_T}\mathbb{E}[\xi^2_{1q}|\mathcal{H}_{q-1}] = \sigma^2\cdot  \mathbb{E}\left[\frac{1}{\bar{N}_T}\sum_{q=1}^{\bar{N}_T}\frac{1-\nu_{q}(\boldsymbol{X}_{q},\mathcal{H}_{t-1})}{[\mathbb{P}\{a_{q} = {\pi}^*(\boldsymbol{X}_{q})\}]^2}\right]\rightarrow \int \frac{\sigma^2}{1-\kappa_\infty (\boldsymbol{x})}d \mathcal{P}_{\boldsymbol{x}},
\end{equation*}
 where $\kappa_\infty (\boldsymbol{x}) = \lim_{q\rightarrow\infty}\mathbb{P}(A_{ti}\neq \pi^*(\boldsymbol{x}))$.

Since the randomness in $\xi_{2q}$ only comes from $\boldsymbol{X}_q$ and $\mathbb{E}[\xi_{2q}] =0$, we have
\begin{equation*}
\begin{aligned}
    \mathbb{E}[\xi_{2q}^2|\mathcal{H}_{q-1}] = \text{Var}(\xi_{2q}) &= \text{Var}\big\{{\pi}^*(\boldsymbol{X}_{ti})\cdot \omega_{ti}\boldsymbol{X}_{ti}'\boldsymbol{\beta}_1 +\{1-{\pi}^*(\boldsymbol{X}_{ti})\}\cdot \omega_{ti}\boldsymbol{X}_{ti}'\boldsymbol{\beta}_0\big\}.
\end{aligned}
\end{equation*}
Furthermore, 
\begin{equation*}
    \mathbb{E}[\xi_{1q}\xi_{2q}|\mathcal{H}_{q-1}] = \mathbb{E}\left[\xi_{2q}\cdot \frac{\boldsymbol{1}\{a_{q} = {\pi}^*(\boldsymbol{X}_{q})\}}{\mathbb{P}(a_{q} = {\pi}^*(\boldsymbol{X}_{q}))} \cdot \epsilon_{q}\big|\mathcal{H}_{q-1}\right] = \mathbb{E}\left[\xi_{2q}\cdot \frac{\boldsymbol{1}\{a_{q} = {\pi}^*(\boldsymbol{X}_{q})\}}{\mathbb{P}(a_{q} = {\pi}^*(\boldsymbol{X}_{q}))} \cdot \mathbb{E}[\epsilon_{q}|\mathcal{H}_{q-1},\boldsymbol{A}_t]\big|\mathcal{H}_{q-1}\right] = 0.
\end{equation*}
Thus, 
\begin{equation}
\begin{aligned}
    \frac{1}{\bar{N}_T}& \sum_{q=1}^{\bar{N}_T}\mathbb{E}[\xi_q^2|\mathcal{H}_{q-1}] =  \frac{1}{\bar{N}_T}\sum_{q=1}^{\bar{N}_T}\mathbb{E}[(\xi_{1q}+\xi_{2q})^2|\mathcal{H}_{q-1}] \\
    & = \frac{1}{\bar{N}_T}\sum_{q=1}^{\bar{N}_T}\mathbb{E}[\xi_{1q}^2|\mathcal{H}_{q-1}] + \frac{1}{\bar{N}_T}\sum_{q=1}^{\bar{N}_T}\mathbb{E}[\xi_{2q}^2|\mathcal{H}_{q-1}] + \frac{2}{\bar{N}_T}\sum_{q=1}^{\bar{N}_T}\mathbb{E}[\xi_{1q}\xi_{2q}|\mathcal{H}_{q-1}]\\
    & \rightarrow \int \frac{\sigma^2}{1-\kappa_\infty (\boldsymbol{x})}d \mathcal{P}_{\boldsymbol{x}} +  \text{Var}\big\{{\pi}^*(\boldsymbol{X}_{ti})\cdot \omega_{ti}\boldsymbol{X}_{ti}'\boldsymbol{\beta}_1 +\{1-{\pi}^*(\boldsymbol{X}_{ti})\}\cdot \omega_{ti}\boldsymbol{X}_{ti}'\boldsymbol{\beta}_0\big\}.
\end{aligned}
\end{equation}
Therefore,
\begin{equation}
    \sigma^2_V  = \int \frac{\sigma^2}{1-\kappa_\infty (\boldsymbol{x})}d \mathcal{P}_{\boldsymbol{x}} +  \text{Var}\big\{{\pi}^*(\boldsymbol{x})\cdot\omega \boldsymbol{x}'\boldsymbol{\beta}_1 +\{1-{\pi}^*(\boldsymbol{x})\}\cdot \omega\boldsymbol{x}'\boldsymbol{\beta}_0\big\}.
\end{equation}

\noindent\textbf{Finally}, by combining the results of Step 1-3, we are able to show that $\sqrt{\bar{N}_T}(\widehat{V}^{\text{DR}}_T -V^{\pi^*})\xrightarrow{\mathcal{D}} \mathcal{N}(0,\sigma^2_V)$, which concludes the proof of this theorem.

\section{Proof of Regret Bound}\label{appendix:regret_bound}
\noindent \textbf{Step 1:} Decompose $R_T =R^{(1)}_T + R^{(2)}_T$, which accounts for the regret of exploitation and exploration. 

Recall that the regret $R_T$ is defined as 
\begin{equation*}
\begin{aligned}
    R_T &=\sum_{t=1}^T\sum_{i=1}^{N_t}\mathbb{E}\big[\mu^{(t,i)}(\boldsymbol{X}_{t},\pi^*(\boldsymbol{X}_{t})) - \mu^{(t,i)}(\boldsymbol{X}_{t},\boldsymbol{A}_{t})\big] \\
    &= \sum_{t=1}^T\sum_{i=1}^{N_t}\mathbb{E}\big[\mu^{(t,i)}(\boldsymbol{X}_{t},\pi^*(\boldsymbol{X}_{t})) - \mu^{(t,i)}(\boldsymbol{X}_{t},\widehat{\pi}_{t-1}(\boldsymbol{X}_{t})) \big] + \sum_{t=1}^T\sum_{i=1}^{N_t}\mathbb{E}\big[ \mu^{(t,i)}(\boldsymbol{X}_{t},\widehat{\pi}_{t-1}(\boldsymbol{X}_{t})) - \mu^{(t,i)}(\boldsymbol{X}_{t},\boldsymbol{A}_{t})\big]\\
    & = \underbrace{\sum_{t=1}^T\sum_{i=1}^{N_t}\mathbb{E}\Big[\mu^{(t,i)}(\boldsymbol{X}_{t},\pi^*(\boldsymbol{X}_{t})) - \mu^{(t,i)}(\boldsymbol{X}_{t},\widehat{\pi}_{t-1}(\boldsymbol{X}_{t})) \Big]}_{R^{(1)}_T} \\
    & + \underbrace{\sum_{t=1}^T\sum_{i=1}^{N_t}\mathbb{E}\Big[\big\{\mu^{(t,i)}(\boldsymbol{X}_{t},\widehat{\pi}_{t-1}(\boldsymbol{X}_{t})) -\mu^{(t,i)}(\boldsymbol{X}_{t},\boldsymbol{A}_{t}) \big\}\cdot \boldsymbol{1}\{ A_{ti} \neq\widehat{\pi}_{t-1}(\boldsymbol{X}_{ti})\}\Big] }_{R^{(2)}_T}.
\end{aligned}
\end{equation*}
By definition, $R^{(1)}_T$ is nonzero only when $\pi^*(\boldsymbol{X}_{ti})\neq \widehat{\pi}_{t-1}(\boldsymbol{X}_{ti})$, which accounts for the regret caused by estimation accuracy, i.e. exploitation. $R^{(2)}_T$ is nonzero only $ A_{ti}\neq\widehat{\pi}_{t-1}(\boldsymbol{X}_{ti})$, which accounts for the regret caused by exploration. In Step 2-3, we will derive the regret bound of $R^{(1)}_T$ and $R^{(2)}_T$ separately to prove the sublinearity of $R_T$.

\noindent \textbf{Step 2:} Prove that $R^{(1)}_T = o(\bar{N}_T^{1/2})$. 

Notice that in the proof of Theorem \ref{thm:2}, step 2.1, we've proved in Equation \eqref{eq:10} that
\begin{equation*}
    \begin{aligned}
        \frac{1}{\sqrt{\bar{N}_T}}\sum_{t=1}^T \sum_{i=1}^{N_t} \left[{\mu}^{(t,i)}(\boldsymbol{X}_t,{\pi}^*(\boldsymbol{X}_{t}))-{\mu}^{(t,i)}(\boldsymbol{X}_t,\widehat{\pi}_{t-1}(\boldsymbol{X}_{t})) \right] =o_p(1).
    \end{aligned}
\end{equation*}
Therefore, 
\begin{equation*}
    R^{(1)}_T = \sum_{t=1}^T\sum_{i=1}^{N_t}\mathbb{E}\Big[\mu^{(t,i)}(\boldsymbol{X}_{t},\pi^*(\boldsymbol{X}_{t})) - \mu^{(t,i)}(\boldsymbol{X}_{t},\widehat{\pi}_{t-1}(\boldsymbol{X}_{t})) \Big] =o(\bar{N}_T^{1/2}) .
\end{equation*}

\noindent \textbf{Step 3:} Prove that $R^{(2)}_T= O(\bar{N}_T^{1/2}\log \bar{N}_T)$. 

According to the upper bound derived in Theorem \ref{thm:kappa_bound}, 
\begin{equation*}
\begin{aligned}
    R^{(2)}_T \leq &\sum_{t=1}^T\sum_{i=1}^{N_t}\mathbb{E}\Big[\big|\mu^{(t,i)}(\boldsymbol{X}_{t},\widehat{\pi}_{t-1}(\boldsymbol{X}_{t})) -\mu^{(t,i)}(\boldsymbol{X}_{t},\boldsymbol{A}_{t}) \big|\cdot \boldsymbol{1}\{ A_{ti} \neq\widehat{\pi}_{t-1}(\boldsymbol{X}_{ti})\}\Big] \\
    & \leq 2U \cdot \sum_{t=1}^T\sum_{i=1}^{N_t}\mathbb{E}\big[\boldsymbol{1}\{ A_{ti} \neq\widehat{\pi}_{t-1}(\boldsymbol{X}_{ti})\}\big] = 2U \cdot \sum_{t=1}^T\sum_{i=1}^{N_t}\kappa_{ti}(\omega_{ti},\boldsymbol{X}_{ti}).
\end{aligned}
\end{equation*}
 Now let's decompose according to different exploration algorithms. For simplicity of notations, we continue with the flattened unit queue $q = Q(t,i)\sum_{s=1}^{t-1}N_s + t$ as shown in the proof of Theorem \ref{thm:UB}. As such, we can extend the definition of $p_t$ to $p_q$ by simply setting $p_q=p_t$ for any unit $q$ in round $t$. As such, $p_q$ is still a non-increasing sequence w.r.t. $q$. By Theorem \ref{thm:kappa_bound}, we have
 
 \begin{enumerate}
\item In UCB, $\kappa_{ti}(\omega_{ti},\boldsymbol{X}_{ti})$ is upper bounded by
$O(L_w^{\gamma}\cdot (\bar{N}_{q-1} p_{q-1})^{-\gamma/2})$. Let $p_q = \bar{N}_{q}^{u/\gamma - 1}$ with $u>0$. For $\gamma >u$, $p_q$ is decreasing. Then 
\begin{equation*}
\begin{aligned}
    R^{(2)}_T \lesssim \sum_{t=1}^T\sum_{i=1}^{N_t}\kappa_{ti}(\omega_{ti},\boldsymbol{X}_{ti}) \lesssim L_w^{\gamma}\cdot \sum_{q=1}^{\bar{N}_T} q^{-u/2} = O(L_w^{\gamma}\cdot \bar{N}_T^{1-u/2}).
\end{aligned}
\end{equation*}
Taking $u = 1$ gives us $R^{(2)}_T = O(L_w^\gamma \cdot \bar{N}_T^{1/2})$. note that when the interference constraint $L_w$ is large, the regret bound $R^{(2)}_T$ tends to be larger.
\item In TS, $\kappa_{ti}(\omega_{ti},\boldsymbol{X}_{ti})$ is upper bounded by $O(\exp\{-\bar{N}_{q-1} p_{q-1}^2/L_w^4\})$. Let $p_q= \sqrt{\alpha\log q/\bar{N}_{q}}$. Then $\kappa_{ti}(\omega_{ti},\boldsymbol{X}_{ti}) \leq O(\exp\{-\alpha\log(q-1)/L_w^4\}) = O (q^{-L_w^{-4}\alpha})$. Thus, 
\begin{equation*}
\begin{aligned}
    R^{(2)}_T \lesssim \sum_{t=1}^T\sum_{i=1}^{N_t}\kappa_{ti}(\omega_{ti},\boldsymbol{X}_{ti}) \lesssim \sum_{q=1}^{\bar{N}_T} q^{-L_w^{-4}\alpha} = O(\bar{N}_T^{1-L_w^{-4}}\alpha).
\end{aligned}
\end{equation*}
When the interference constraint $L_w$ is large, the regret bound $R^{(2)}_T$ tends to be larger. By taking $\alpha = L_w^{4}/2$, we have $R_T^{(2)} = O(\bar{N}_T^{1/2})$.
\item In EG, $\kappa_{ti}(\omega_{ti},\boldsymbol{X}_{ti}) = \epsilon_q/2$. If we set $\epsilon_q = O(q^{-m})$ with any $m<1/2$, then
\begin{equation*}
\begin{aligned}
    R^{(2)}_T \lesssim \sum_{t=1}^T\sum_{i=1}^{N_t}\kappa_{ti}(\omega_{ti},\boldsymbol{X}_{ti}) \lesssim \sum_{q=1}^{\bar{N}_T} q^{-m} = O(\bar{N}_T^{1-m}).
\end{aligned}
\end{equation*}
By setting $\epsilon_q = O(\log q /\sqrt{q})$, we have $R_T^{(2)} = O(\bar{N}_T^{1/2}\log \bar{N}_T)$.
 \end{enumerate}

Thus, in UCB, TS, and EG, the regret caused by exploration can be controlled by  $R_T^{(2)} = O(\bar{N}_T^{1/2}\log \bar{N}_T)$.

Therefore, by combining the results of Step 1-3, we are able to show that 
$$R_T = \sum_{t=1}^T\sum_{i=1}^{N_t}\mathbb{E}[R^*_{ti} - R_{ti}] 
 = O(\bar{N}_T^{1/2}\log \bar{N}_T),$$ 
 which is sublinear in $\bar{N}_T$.

\end{document}